\documentclass[manuscript]{acmart}

\usepackage{amssymb}
\usepackage{amsmath}
\usepackage{graphicx} 
\usepackage{multicol}
\usepackage{float}
\usepackage{blindtext}
\makeatletter
\let\over\@@over
\makeatother
\usepackage{subcaption}
\usepackage{natbib}

\usepackage{enumitem}
\usepackage{hyperref} 
\usepackage{tikz}
\usetikzlibrary{calc}
\usepackage{subfiles}

\hypersetup{
    colorlinks=true,
    linkcolor=blue,
    filecolor=magenta,      
    urlcolor=cyan,
    citecolor=blue, 
    }

\title{\textbf{Evaluating the Reasoning  Abilities of LLMs on Underrepresented Mathematics Competition Problems}}
\author{Samuel Golladay}
\email{sgwtd@umkc.edu}
\orcid{0009-0002-1724-7846}
\author{Majid Bani Yaghoub}
\email{baniyaghoubm@umkc.edu}
\affiliation{%
    \institution{University of Missouri: Kansas City}
    \city{Kansas City}
    \state{Missouri}
    \country{USA}
}

\begin{document}


\begin{abstract}
Understanding the limitations of Large Language Models (LLMs) in mathematical reasoning has been  the focus of several recent studies; however, the majority of these studies use the same datasets for benchmarking, which limits the generalizability of their findings and may not fully capture the diverse challenges present in mathematical tasks. The purpose of the present study is to analyze the performance of LLMs on underrepresented mathematics competition problems. We prompted three leading LLMs (GPT-4o-mini, Gemini-2.0-Flash, and DeepSeek-V3) with the Missouri Collegiate Mathematics Competition problems in the areas of Calculus, Analytic Geometry, and Discrete Mathematics. The LLM's responses were then compared to the known correct solutions in order to determine the accuracy of the LLM for each problem domain. We also analyzed the LLMs’ reasoning to explore patterns in errors across problem types and models. DeepSeek-V3 has the best performance in all three categories of Calculus (80\%), Analytic Geometry (80\%), and Discrete Mathematics (90\%), both in reasoning and correct final answers. All three LLMs exhibited notably weak performance in Geometry. The majority of errors made by DeepSeek-V3 were attributed to computational and logical mistakes, whereas GPT-4o-mini frequently exhibited logical and approach-related errors. Gemini, on the other hand, tended to struggle with incomplete reasoning and drawing rushed conclusions. In conclusion, evaluating LLMs on underrepresented mathematics competition datasets can provide deeper insights into their distinct error patterns and highlight ongoing challenges in structured reasoning, particularly within the domain of Geometry.
        
\end{abstract}

\settopmatter{printacmref=false}
\setcopyright{none}
\renewcommand\footnotetextcopyrightpermission[1]{}
\pagestyle{plain}

\maketitle

\section{Introduction}
Large Language Models (LLMs) have increasingly become popular in various fields, from research to everyday questions. They have often demonstrated good performance in some fields of science and technology \cite{zhang2024careful, huang2023chatgpt, wu2023bloomberggpt}. Nonetheless, their ability to solve mathematical problems has become a question of considerable importance. Problems in Calculus, Analytic Geometry, and Discrete Mathematics require a blend of abstract and logical reasoning, visualization, numerical performance, and knowledge of problem-solving techniques \cite{ahn2024largelanguagemodelsmathematical, hendrycks2021measuringmathematicalproblemsolving}. Problems in these fields form a metric for analyzing LLM performance in logic-oriented problems \cite{herman2018analysis, huang2022towards, kenderov2006competitions}. \\

Several studies have tested LLMs' abilities to solve math problems \cite{doi:10.1073/pnas.2318124121, ahn2024largelanguagemodelsmathematical, herman2018analysis,  kenderov2006competitions}; however, they have often limitations as follows: (i) Most studies use the same datasets for benchmarking, such as the MATH and GSM8K datasets. This can lead to dataset "contamination," where LLMs are trained on the same data used for evaluation. Such contamination compromises the fairness and validity of the assessment and artificially inflates performance metrics, making it difficult to accurately measure a model’s true capabilities;  (ii) some studies focus exclusively on evaluating flagship language models. While this is useful for understanding the current limits of LLM capabilities, it does not accurately reflect the models that are accessible to the general public;  (iii) other studies assess LLM performance solely based on the final answer, paying little attention to the underlying reasoning process.\\

To address these issues, we evaluate the performance of freely accessible and currently prominent LLMs on underrepresented mathematics competition datasets, focusing not only on accuracy but also on the quality and structure of their reasoning. Moreover, this study extends beyond overall accuracy metrics to provide a detailed analysis of the reasoning errors each model exhibits. By identifying and characterizing these reasoning failures, we aim to uncover the specific challenges LLMs face when solving complex, underrepresented mathematical problems and to better understand their current limitations. This investigation addresses two central questions: (1) To what extent can LLMs reliably produce correct solutions to complex, multi-step mathematical problems? and (2) Are the reasoning processes underpinning these solutions mathematically valid? The findings offer a more nuanced understanding of LLMs' mathematical capabilities to solve underrepresented mathematical problems, contributing valuable insights into their potential use in education, scientific research, and automated problem-solving systems \cite{chang2024survey, hadi2023survey, kasneci2023chatgpt}.\\

\section{Materials and Methods}


In order to assess the performance of the LLMs in the three mathematical disciplines, we prompted the models with the problem statements, as well as a command to solve the problem analytically. 

\subsection{Datasets} The Missouri Collegiate Mathematics Competition (MCMC) datasets are relatively unknown and, to the best of our knowledge, have not been previously used to evaluate LLMs capabilities \cite{missouri_contest_archives}. This significantly reduces the likelihood of dataset contamination and allows for a more reliable assessment of model performance. The MCMC draws its problems from a variety of reputable sources, including but not limited to the William Lowell Putnam Mathematical Competition, the MAA (Mathematical Association of America) Problem Directory, the Mathematical Olympiads of the USSR, and the Iowa Collegiate Mathematics Competition. The MCMC problems were categorized into three mathematical disciplines: Calculus, Analytic Geometry, and Discrete Mathematics. Ten problems were chosen from each category, resulting in a total of thirty problems. Additionally, problems were chosen in a way that they could be used for assessing the LLMs' accuracy based on a final answer as well as the reasoning process. Hence, problems requiring mathematical proof were not included in this study. The sets of selected problems with the specific year and correct answers are available in Section 1 of the supplementary document.


\subsection{LLMs Evaluated}  The study assessed the performance of three prominent Large Language Models (LLMs): GPT-4o-mini, Gemini-2.0-flash, and DeepSeek-V3. GPT-4o-mini is a lightweight variant of OpenAI's GPT-4o, optimized for efficiency and speed \cite{siddiky2025optimizing, 4oMiniOpenAI}. Gemini-2.0-flash, developed by Google, excels in natural language understanding and is also designed for efficiency \cite{imran2024google, GeminiGoogle}. DeepSeek is designed for technical tasks, particularly in mathematics, and is known for its ability to model logical progression and identify mathematical patterns, as well as its enormously reduced training cost \cite{liu2024deepseek, DeepSeekV3DeepSeek}. All three models were tested using the same set of problems from the Missouri Collegiate Mathematics Competition, with each model prompted to provide both reasoning and final answers.

\subsection{Evaluation Criteria} The models' responses were categorized into four possible outcomes: (1) correct final answer with correct reasoning, (2) incorrect final answer with correct reasoning (e.g., minor computational mistakes), (3) incorrect final answer with incorrect reasoning, and (4) incorrect final answer with correct reasoning. We conducted a detailed analysis of reasoning errors based on expert evaluation, categorizing responses with incorrect reasoning into 11 distinct types. These reasoning error categories are summarized in Table~\ref{tab:error-categories}.



\begin{table}[H]
\centering
\caption{Categories of Reasoning Errors in Underrepresented Mathematical Problem Solving}
\vspace{-0.3cm}
\label{tab:error-categories}
\begin{tabular}{p{3.5cm} p{11cm}}
\hline
\textbf{Error Type} & \textbf{Description} \\
\hline
Ambiguous Step & Steps that are unclear, imprecise, or open to multiple interpretations. \\
Debatable Theories & Use of theorems, principles, or methods with significant ambiguity or lacking consensus. \\
Incomplete Reasoning & Omission of necessary steps or failure to fully develop the solution. \\
Irrelevant Step & Inclusion of steps that do not contribute to solving the problem. \\
Logical Error & Reasoning that follows an illogical or invalid progression. \\
Computational Error & Miscalculation while evaluating expressions or performing operations. \\
Methodological Error & Incorrect application or selection of a mathematical method or technique. \\
Fact Error & Misuse or misstatement of mathematical facts, theorems, or definitions. \\
Approach Error & Adoption of an incorrect or inefficient strategy to solve the problem. \\
Redundancy Error & Inclusion of repetitive or unnecessary steps without no progress. \\
Rushed to Conclusion & Prematurely providing a final answer without sufficient  reasoning. \\
\hline
\end{tabular}
\end{table}








\section{Results}

The performance of the evaluated LLMs in solving mathematics competition problems was assessed based on two main criteria: the accuracy of the final answer and the correctness of the reasoning process. The output solutions provided by GPT-4o-mini, Gemini-2.0-flash, and DeepSeek-V3 are available in sections 2-4 of the supplementary document. 

\subsection{Evaluation of Model Performances}

\begin{figure}[h]
    \centering
    \vspace{-0.5cm}
    \includegraphics[width=0.42\textwidth]{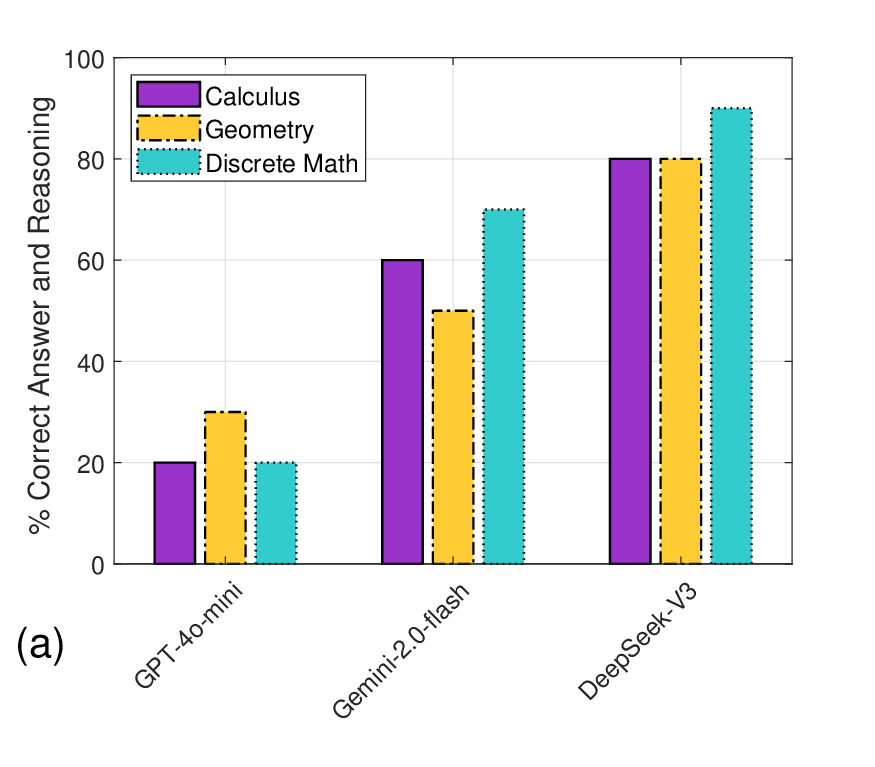}
    \vspace{-0.5cm}
    \includegraphics[width=0.42\textwidth]{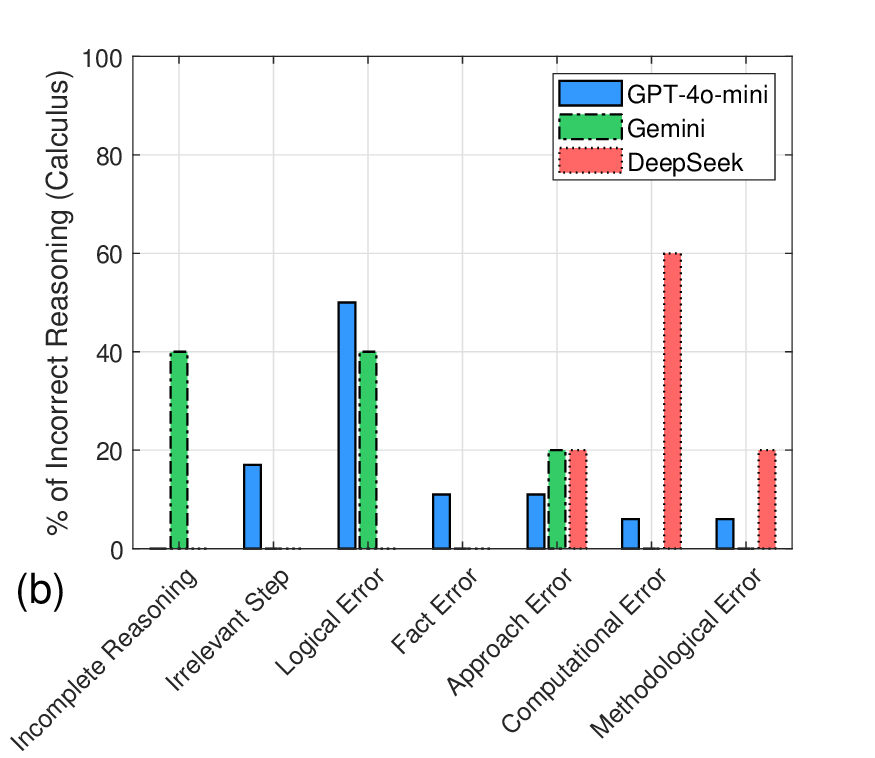}
   \vspace{0.5cm}
    
    \includegraphics[width=0.42\textwidth]{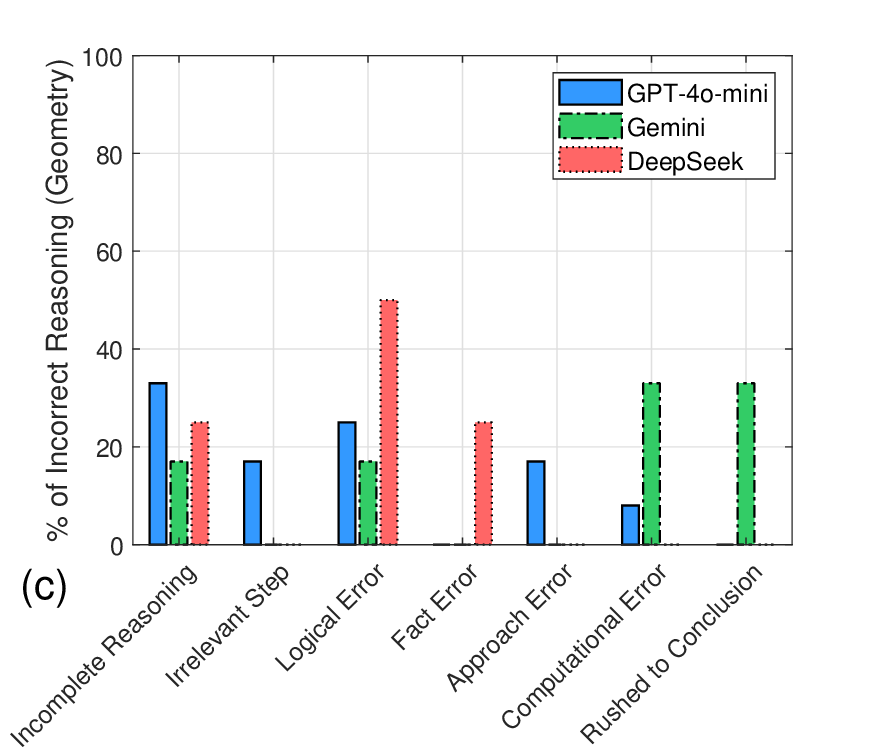}
        \vspace{-0.5cm}
    \includegraphics[width=0.42\textwidth]{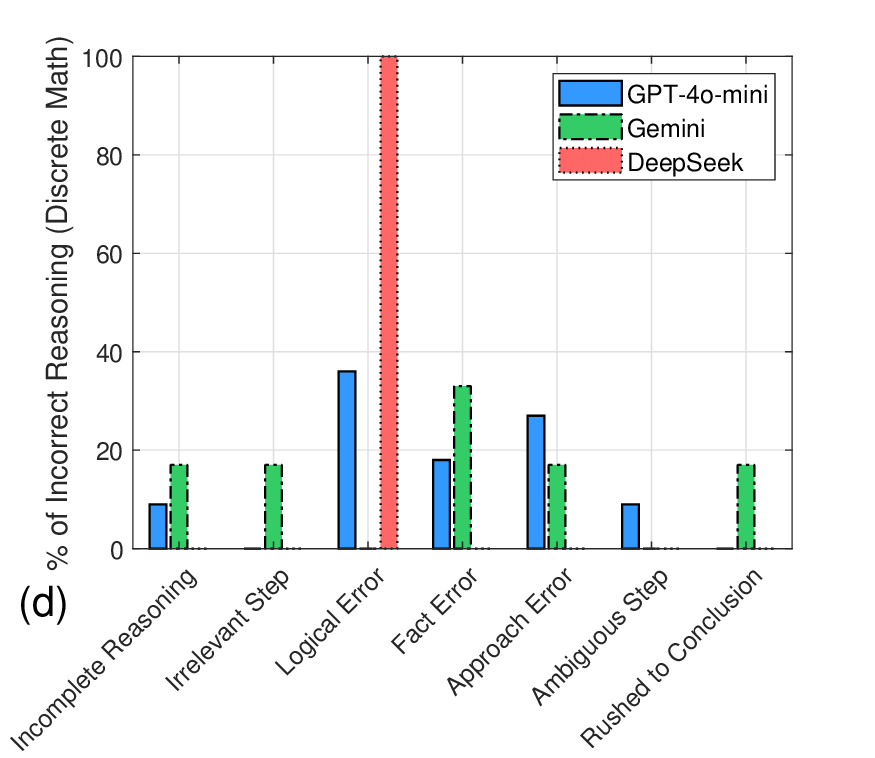}\\
    \vspace{0.5cm}
    \caption{Model evaluation on mathematical reasoning tasks. 
    (a) Percentages of correct answers with correct reasoning.
    (b)–(d) Analysis of incorrect reasoning types in Calculus, Geometry, and Discrete Math problems, respectively.}
    \label{fig:model-evaluation}
\end{figure}

Figure~\ref{fig:model-evaluation} provides a comparison of the three language models—GPT-4o-mini, Gemini 2.0 Flash, and DeepSeek V3—across the three mathematical domains. Panel (a) summarizes the overall performance of each model in terms of the percentage of correct answers with correct reasoning, incorrect answers with correct reasoning, and incorrect answers with incorrect reasoning. Among the models, DeepSeek-V3 demonstrates the highest accuracy with the vast majority of its answers classified as correct.
Panels (b), (c), and (d) break down the types of reasoning errors observed in the calculus, geometry, and discrete mathematics problems, respectively. Each panel shows only the error types that occurred with non-zero frequency. The most prominent reasoning error across all three domains is the presence of logical errors. In Calculus (Panel b), both GPT-4o-mini and Gemini exhibit a majority of logical mistakes, while DeepSeek mostly suffers from computational errors. In Geometry (Panel c), logical and computational errors are common among all three models. For Discrete Math (Panel d), all of DeepSeek-V3's errors are due to logical reasoning mistakes, whereas Gemini displays mostly logical errors and approach errors.


\subsection{Assessment of Model Accuracies}

Table 2 shows the classification of the models' responses for each mathematical category. These outcomes reflect the models' ability to arrive at the correct answer and the logical soundness of their reasoning processes. There was no case of an incorrect final answer with correct reasoning, possibly due to low sample size. 

\begin{table}[H]
\centering
\caption{Classification of the models based on final answer accuracy and reasoning correctness.}\vspace{-0.3cm}
\begin{tabular}{p{6.5cm}cccc}
\hline  \hline
\textbf{Model} & \textbf{Calculus} & \textbf{Geometry} & \textbf{Discrete Math} & \textbf{Average} \\ \hline
\textbf{GPT-4o-mini} & 20\% 0\% 80\% & 30\% 0\% 70\% & 20\% 10\% 70\% & 23\% 3\% 73\% \\
\textbf{Gemini-2.0-flash} & 60\% 0\% 40\% & 50\% 0\% 50\% & 70\% 0\% 30\% & 60\% 0\% 40\% \\
\textbf{DeepSeek-V3} & 80\% 0\% 20\% & 80\% 10\% 10\% & 90\% 0\% 10\% & 83\% 3\% 13\% \\
\hline
\end{tabular}
\textit{First percentage = Correct final answer with correct reasoning, Second percentage = Incorrect final answer with correct reasoning, Third percentage = Incorrect final answer with incorrect reasoning}
\end{table}

\subsection{Analysis of Incorrect Reasoning}

Table 3 provides a deeper look into the types of reasoning errors made by the models when their final answers were incorrect. The table highlights the percentage of errors in each of the 11 reasoning categories: ambiguous steps, debatable theories, incomplete reasoning, irrelevant steps, logical errors, computational errors, methodological errors, fact errors, approach errors, redundancy errors, and rushed conclusions. 

\begin{table}[H]
\centering
\caption{Analysis of Incorrect Reasoning for each mathematical category. The values correspond to percentages for GPT-4o-mini, Gemini, and DeepSeek models, respectively.} \vspace{-0.3cm}
\begin{tabular}{p{6.0cm}cccc}
\hline
\textbf{Error Type} & \textbf{Calculus} & \textbf{Geometry} & \textbf{Discrete Math} & \textbf{Average}  \\ \hline  \hline
Ambiguous Step & 00\% 00\% 00\% & 00\% 00\% 00\% & 9\% 00\% 00\% & 2\% 00\% 00\%  \\
Debatable Theories & 00\% 00\% 00\% & 00\% 00\% 00\% & 00\% 00\% 00\% & 00\% 00\% 00\%  \\
Incomplete Reasoning & 00\% \textbf{40}\% 00\% & 33\% 17\% 25\% & 09\% 17\% 00\% & 12\% 24\% 10\%  \\
Irrelevant Step & 17\% 00\% 00\% & 17\% 00\% 00\% & 00\% 17\% 00\% & 12\% 06\% 00\%  \\
Logical Error & 50\% 40\% 00\% & 25\% 17\% 50\% & 36\% 00\% 100\% & 39\% 18\% 30\%  \\
Computational Error & 06\% 00\% 60\% & 08\% 33\% 00\% & 00\% 00\% 00\% & 05\% 11\% 30\%  \\
Methodological Error & 06\% 00\% 20\% & 00\% 00\% 00\% & 00\% 00\% 00\% & 02\% 00\% 10\%  \\
Fact Error & 11\% 00\% 00\% & 00\% 00\% 25\% & 18\% 33\% 00\% & 10\% 11\% 10\%  \\
Approach Error & 11\% 20\% 20\% & 17\% 00\% 00\% & 27\% 17\% 00\% & 17\% 11\% 10\%  \\
Redundancy Error & 00\% 00\% 00\% & 00\% 00\% 00\% & 00\% 00\% 00\% & 00\% 00\% 00\%  \\
Rushed to Conclusion & 00\% 00\% 00\% & 00\% 33\% 00\% & 00\% 17\% 00\% & 00\% 18\% 00\%  \\
\hline
\end{tabular}
\textit{Bolded entry signifies that Gemini-2.0-flash made 40\% of its Calculus errors due to incomplete reasoning.}
\end{table}

GPT-4o-mini primarily struggles with logical and approach errors across all mathematical domains, indicating a tendency to follow flawed reasoning paths or select suboptimal problem-solving strategies. In Geometry, it also shows a notable rate of incomplete reasoning, suggesting difficulties in carrying solutions through to completion. Gemini exhibits inconsistent reasoning patterns, with significant incomplete reasoning and rushed conclusions in Geometry, pointing to a tendency to jump to answers without fully developed steps. DeepSeek shows the highest frequency of logical and computational errors, especially in Calculus and Discrete Math, reflecting issues with both reasoning accuracy and numerical operations. Overall, each model reveals distinct weaknesses in how they process and execute mathematical reasoning.

\section{Discussion and Conclusion}

The results presented in this study highlight the strengths and weaknesses of the three large language models (LLMs)—GPT-4o-mini, Gemini-2.0-flash, and DeepSeek-V3—in solving underrepresented mathematical problems from the following categories: Calculus, Analytic Geometry, and Discrete Mathematics. \\

Overall, the models performance varied significantly in providing correct final answers. DeepSeek-V3 exhibited the highest accuracy in both final answers and reasoning, with particularly impressive results in discrete mathematics, where it achieved the highest percentage of correct answers with correct reasoning (90\%). This indicates that DeepSeek-V3 may have a better understanding of the mathematical principles involved in discrete mathematics problems compared to the other models. In contrast, GPT-4o-mini performed poorly across all three problem domains, receiving a maximum percentage of correct answers with correct reasoning of 30\%. While previous studies have shown that GPT-4o-mini can successfully solve simple math problems, GPT-4o-mini's performance in this study demonstrates its inability to solve more complex problems \cite{chang2024survey, ahn2024largelanguagemodelsmathematical, hadi2023survey}. Gemini 2.0-flash demonstrated mediocre performance, with it solving 70\% of discrete math problems, but only 50\% of geometry problems. This highlights the limitations of Gemini 2.0-flash in visualizing abstract geometric figures.\\

When analyzing the incorrect reasoning, the most frequent errors vary by model and mathematical domain. However, in general the most common errors were logical errors and errors due to incomplete reasoning. Additionally, many of the incomplete reasoning errors involved models failing to test all cases. That is, a model would check one or two cases and then assume that a condition would hold for all possible cases. Additionally, logical errors were typically related to steps being performed incorrectly. The prevalence of these errors demonstrates that large language models currently struggle to understand individual steps in the grand scheme of the entire problem, and rather begin a step, or a case, and then upon completion determine that the problem is finished. However, DeepSeek-V3 had the fewest errors in any problem domain, likely due to the fact that it double-checks its solution before finishing. This gave it an edge on problems requiring multiple cases to be checked.\\

The findings underscore the need for continued improvements in LLMs' reasoning capabilities in mathematics. Although sample size is a limitation, we are 95\% confident that the true accuracy of DeepSeek-V3, Gemini-2.0-flash, and GPT-4o-mini lie within (69\%, 97\%), (42\%, 78\%), and (8\%, 38\%) respectively. As such, it is clear that these models have room for improvement. Gemini-2.0-flash demonstrated unreliable performance, showing that its responses to complex, multi-step mathematical problems should require verification. Finally, GPT-4o-mini's poor performance across all three mathematical domains showcases its inability to solve complex underrepresented mathematical problems and thus perhaps worse overall performance in logic-oriented tasks.\\

Other prominent LLMs to be tested include Meta's LLaMa 3 family as well as more advanced variants of Gemini and GPT 4o. The models tested in this study were chosen specifically due to their free-to-use nature and current prominence. Additionally, research into other novel methods of improving LLM performance, such as proxies that offload computation to improve computational accuracy, is ongoing \cite{lu2024mathcoder2bettermathreasoning}. Another crucial element of evaluating LLM performance is using less common datasets for benchmarking \cite{doi:10.1073/pnas.2318124121}. Using underrepresented datasets prevents contamination and prevents LLMs from appearing overly accurate when performance testing \cite{sainz2023nlpevaluationtroubleneed}.\\

In conclusion, while LLMs have made significant strides in solving mathematical problems, there are still notable areas, such as Geometry, where their reasoning needs improvement. Models such as DeepSeek-V3 show promise in handling underrepresented, multi-step mathematical tasks, but other prevalent models (e.g., GPT-4o-mini and Gemini-2.0-flash) still exhibit limitations in terms of logical consistency and computational accuracy. As these models continue to evolve, it will be important to address these challenges to improve their overall reliability in mathematical problem-solving. Future research should focus on refining reasoning processes, reducing errors, and expanding the models' understanding of the process of solving multi-step problems with greater consistency and accuracy.
\bibliographystyle{plain}
\bibliography{References}

@misc{missouri_contest_archives,
  author = {{Missouri Collegiate Mathematics Competition}},
  title = {Missouri Collegiate Mathematics Contest Archives},
  howpublished = {\url{http://sections.maa.org/missouri/contest_archives.html}},
}

@misc{ahn2024largelanguagemodelsmathematical,
      title={Large Language Models for Mathematical Reasoning: Progresses and Challenges}, 
      author={Janice Ahn and Rishu Verma and Renze Lou and Di Liu and Rui Zhang and Wenpeng Yin},
      eprint={2402.00157},
      archivePrefix={arXiv},
      primaryClass={cs.CL},
      url={https://arxiv.org/abs/2402.00157}, 
}

@misc{hendrycks2021measuringmathematicalproblemsolving,
      title={Measuring Mathematical Problem Solving With the MATH Dataset}, 
      author={Dan Hendrycks and Collin Burns and Saurav Kadavath and Akul Arora and Steven Basart and Eric Tang and Dawn Song and Jacob Steinhardt},
      eprint={2103.03874},
      archivePrefix={arXiv},
      primaryClass={cs.LG},
      url={https://arxiv.org/abs/2103.03874}, 
}

@inproceedings{herman2018analysis,
  title={Analysis of students’ mathematical reasoning},
  author={Herman, T and others},
  booktitle={Journal of Physics: Conference Series},
  volume={948},
  number={1},
  pages={012036},
  organization={IOP Publishing}
}

@article{zhang2024careful,
  title={A careful examination of large language model performance on grade school arithmetic},
  author={Zhang, Hugh and Da, Jeff and Lee, Dean and Robinson, Vaughn and Wu, Catherine and Song, William and Zhao, Tiffany and Raja, Pranav and Zhuang, Charlotte and Slack, Dylan and others},
  journal={Advances in Neural Information Processing Systems},
  volume={37},
  pages={46819--46836},
}

@article{huang2023chatgpt,
  title={ChatGPT for shaping the future of dentistry: the potential of multi-modal large language model},
  author={Huang, Hanyao and Zheng, Ou and Wang, Dongdong and Yin, Jiayi and Wang, Zijin and Ding, Shengxuan and Yin, Heng and Xu, Chuan and Yang, Renjie and Zheng, Qian and others},
  journal={International Journal of Oral Science},
  volume={15},
  number={1},
  pages={29},
  publisher={Nature Publishing Group UK London}
}

@article{wu2023bloomberggpt,
  title={Bloomberggpt: A large language model for finance},
  author={Wu, Shijie and Irsoy, Ozan and Lu, Steven and Dabravolski, Vadim and Dredze, Mark and Gehrmann, Sebastian and Kambadur, Prabhanjan and Rosenberg, David and Mann, Gideon},
  journal={arXiv preprint arXiv:2303.17564},
}

@article{imran2024google,
  title={Google Gemini as a next generation AI educational tool: a review of emerging educational technology},
  author={Imran, Muhammad and Almusharraf, Norah},
  journal={Smart Learning Environments},
  volume={11},
  number={1},
  pages={22},
  publisher={Springer}
}

@article{liu2024deepseek,
  title={Deepseek-v3 technical report},
  author={Liu, Aixin and Feng, Bei and Xue, Bing and Wang, Bingxuan and Wu, Bochao and Lu, Chengda and Zhao, Chenggang and Deng, Chengqi and Zhang, Chenyu and Ruan, Chong and others},
  journal={arXiv preprint arXiv:2412.19437},
}

@article{siddiky2025optimizing,
  title={Optimizing ai language models: A study of chatgpt-4 vs. chatgpt-4o},
  author={Siddiky, Md Nurul Absar and Rahman, Muhammad Enayetur and Hossen, MD and Rahman, Muhammad Rezaur and Jaman, Md Shahadat},
  journal={Preprints. org},
}

@inproceedings{kenderov2006competitions,
  title={Competitions and mathematics education},
  author={Kenderov, Petar S},
  booktitle={Proceedings of the international congress of mathematicians},
  volume={3},
  pages={1583--1598},
  organization={IMU. Madrid}
}

@article{chang2024survey,
  title={A survey on evaluation of large language models},
  author={Chang, Yupeng and Wang, Xu and Wang, Jindong and Wu, Yuan and Yang, Linyi and Zhu, Kaijie and Chen, Hao and Yi, Xiaoyuan and Wang, Cunxiang and Wang, Yidong and others},
  journal={ACM transactions on intelligent systems and technology},
  volume={15},
  number={3},
  pages={1--45},
  publisher={ACM New York, NY}
}

@article{hadi2023survey,
  title={A survey on large language models: Applications, challenges, limitations, and practical usage},
  author={Hadi, Muhammad Usman and Qureshi, Rizwan and Shah, Abbas and Irfan, Muhammad and Zafar, Anas and Shaikh, Muhammad Bilal and Akhtar, Naveed and Wu, Jia and Mirjalili, Seyedali and others},
  journal={Authorea Preprints},
  volume={3},
}

@article{huang2022towards,
  title={Towards reasoning in large language models: A survey},
  author={Huang, Jie and Chang, Kevin Chen-Chuan},
  journal={arXiv preprint arXiv:2212.10403},
}

@article{kasneci2023chatgpt,
  title={ChatGPT for good? On opportunities and challenges of large language models for education},
  author={Kasneci, Enkelejda and Se{\ss}ler, Kathrin and K{\"u}chemann, Stefan and Bannert, Maria and Dementieva, Daryna and Fischer, Frank and Gasser, Urs and Groh, Georg and G{\"u}nnemann, Stephan and H{\"u}llermeier, Eyke and others},
  journal={Learning and individual differences},
  volume={103},
  pages={102274},
  publisher={Elsevier}
}

@misc{GeminiGoogle,
  title = {Introducing Gemini 2.0: our new AI model for the agentic era},
author={Google},
howpublished = {\url{https://web.archive.org/web/20250407005349/https://blog.google/technology/google-deepmind/google-gemini-ai-update-december-2024/#project-mariner}},
  note = {Accessed: 2025-04-06}
}

@misc{4oMiniOpenAI,
    title={GPT-4o mini: advancing cost-efficient intelligence},
    author={OpenAI},
    howpublished={\url{https://web.archive.org/web/20250407005631/https://openai.com/index/gpt-4o-mini-advancing-cost-efficient-intelligence/}},
    note = {Accessed: 2025-04-06}
}

@misc{DeepSeekV3DeepSeek,
    title={DeepSeek v3: Advanced AI Language Model},
    author={DeepSeek},
    howpublished=
    {\url{https://web.archive.org/web/20250401133236/https://api-docs.deepseek.com/news/news1226}},
    note={Accessed: 2025-04-06}
}

@misc{lu2024mathcoder2bettermathreasoning,
      title={MathCoder2: Better Math Reasoning from Continued Pretraining on Model-translated Mathematical Code}, 
      author={Zimu Lu and Aojun Zhou and Ke Wang and Houxing Ren and Weikang Shi and Junting Pan and Mingjie Zhan and Hongsheng Li},
      year={2024},
      eprint={2410.08196},
      archivePrefix={arXiv},
      primaryClass={cs.CL},
      url={https://arxiv.org/abs/2410.08196}, 
}

@article{
doi:10.1073/pnas.2318124121,
author = {Katherine M. Collins  and Albert Q. Jiang  and Simon Frieder  and Lionel Wong  and Miri Zilka  and Umang Bhatt  and Thomas Lukasiewicz  and Yuhuai Wu  and Joshua B. Tenenbaum  and William Hart  and Timothy Gowers  and Wenda Li  and Adrian Weller  and Mateja Jamnik },
title = {Evaluating language models for mathematics through interactions},
journal = {Proceedings of the National Academy of Sciences},
volume = {121},
number = {24},
pages = {e2318124121},
year = {2024},
doi = {10.1073/pnas.2318124121},
URL = {https://www.pnas.org/doi/abs/10.1073/pnas.2318124121},
eprint = {https://www.pnas.org/doi/pdf/10.1073/pnas.2318124121}}

@misc{sainz2023nlpevaluationtroubleneed,
      title={NLP Evaluation in trouble: On the Need to Measure LLM Data Contamination for each Benchmark}, 
      author={Oscar Sainz and Jon Ander Campos and Iker García-Ferrero and Julen Etxaniz and Oier Lopez de Lacalle and Eneko Agirre},
      year={2023},
      eprint={2310.18018},
      archivePrefix={arXiv},
      primaryClass={cs.CL},
      url={https://arxiv.org/abs/2310.18018}, 
}

\pagebreak
\appendix

\section{Competition Problems with Correct Solutions}
\subsection{Calculus Problems with Correct Solutions}

\subsubsection{1996 Session 2: 1}

Evaluate the definite integrals
\smallskip
(a) $$\int_1^3 {dx \over \sqrt {(x-1)(3-x)}} ,$$
\smallskip
(b) $$\int_1^\infty {dx \over e^{x+1} + e^{3-x}} .$$
\bigskip
Solution.
\smallskip
(a)  Since the integrand is not defined at either bound of integration, 
one should write
\begin{align*}
\int_1^3 {\frac{dx}{\sqrt {(x-1)(3-x)}}} &= \lim_{\substack{\epsilon \to 0^+ \\ \delta \to 0^+}} \int_{1+\epsilon}^{3-\delta} {\frac{dx}{\sqrt {(x-1)(3-x)}}} \\
&= \lim_{\substack{\epsilon \to 0^+ \\ \delta \to 0^+}} \int_{1+\epsilon}^{3-\delta} {\frac{dx}{\sqrt {1- (x-2)^2}}} \\
&= \lim_{\substack{\epsilon \to 0^+ \\ \delta \to 0^+}} \arcsin (x-2) \bigg\vert _{1+\epsilon}^{3-\delta} \\
&= \lim_{\substack{\epsilon \to 0^+ \\ \delta \to 0^+}} [ \arcsin (1-\delta) - \arcsin (\epsilon - 1)] \\
&= {\pi \over 2} - (-\frac{\pi}{2}) \\
&= \pi.
\end{align*}
\smallskip
(b)  The difficulty here is with the infinite interval of integration.  
Let $y=x-1$; then
\begin{align*}
\int_1^\infty \frac{dx}{e^{x+1} + e^{3-x}} &= \lim_{N \to \infty} \int_1^N \frac{dx}{e^{x+1} + e^{3-x}} \\
&= \lim_{N \to \infty} \frac{1}{e^2} [\arctan e^{N-1} - \arctan e^0 ] \\
&= \frac{1}{e^2} \left( \frac{\pi}{2} - \frac{\pi}{4} \right) \\
&= \frac{\pi}{4e^2}.
\end{align*}

\subsubsection{1997 Session 2: 4}

Define a family of curves by
$$S_n = \{ (x,y) : y = {1 \over n} \sin (n^2 x),\ 0 \le x \le \pi \},$$
where $n$ is a positive integer.  What is the limit of the length of 
$S_n$ as $n \to \infty$? 
\bigskip
Solution.
\smallskip
Denote the length of $S_n$ by $L(S_n)$.  Then
$$L(S_n) = \int_0^\pi \sqrt {1 + n^2 \cos ^2 (n^2 x)} dx .$$
Clearly,
$$\sqrt {1 + n^2 \cos ^2 (n^2 x)} > \sqrt {n^2 \cos ^2 (n^2 x)} = n \vert 
\cos (n^2 x) \vert ,$$
so
$$L(S_n) > \int_0^\pi n \vert \cos (n^2 x) \vert dx .$$
The period of $\cos (n^2 x)$ is $2 \pi / n^2$, and $\cos (n^2 x) \ge 0$ 
for $0 \le x \le \pi / 2n^2$, so
\begin{align*}
L(S_n) &> 2n^3 \int_0^{\pi / 2n^2} \cos (n^2 x) dx \\
&= 2n^3 \left[ \frac{\sin (n^2 x)}{n^2} \right]_0^{\pi / 2n^2} \\
&= 2n^3 \left( \frac{\sin (n^2 \cdot \frac{\pi}{2n^2})}{n^2} - \frac{\sin (n^2 \cdot 0)}{n^2} \right) \\
&= 2n^3 \left( \frac{\sin (\frac{\pi}{2})}{n^2} - 0 \right) \\
&= 2n^3 \left( \frac{1}{n^2} \right) \\
&= 2n.
\end{align*}
Therefore,
$$\lim_{n \to \infty} L(S_n) = \infty .$$

\subsubsection{1998 Session 1: 3}

If $m$ and $n$ are positive integers and $a < b$, find a formula for
$$\int_a^b {(b-x)^m \over m!} {(x-a)^n \over n!} dx$$
and use your formula to evaluate
$$\int_0^1 (1-x^2)^n dx .$$
\bigskip
Solution.
\smallskip
Integration by parts once yields
$$\int_a^b {(b-x)^m \over m!} {(x-a)^n \over n!} dx = \int_a^b 
{(b-x)^{m-1} \over (m-1)!} {(x-a)^{n+1} \over (n+1)!} dx $$
so continuing, we get
$$\int_a^b {(b-x)^m \over m!} {(x-a)^n \over n!} dx = \int_a^b 
{(x-a)^{n+m} \over (n+m)!} dx = {(b-a)^{n+m+1} \over (n+m+1)!} .$$
Now,
$$\int_0^1 (1-x^2)^n dx = {1 \over 2} \int_{-1}^1 (1-x)^n (x-(-1))^n dx = 
{1 \over 2} (n!)^2 {2^{2n+1} \over (2n+1)!} = {2 \cdot 4 \cdot 6 \cdots 
2n \over 1 \cdot 3 \cdot 5 \cdots (2n+1)} .$$

\subsubsection{1999 Session 2: 2}

Give a precise characterization of those points in the plane which do 
$\underline{\text{not}}$ lie on a tangent line to the curve $y = x^4 - 
6x^2$. 
\bigskip
Solution.
\smallskip
Given a point $( t , t^4 - 6t^2 )$ on the curve, the equation of the   
tangent line at that point is $y = (4 t^3 - 12 t ) x - (3 t^4 - 6 t^2 
)$.  For a fixed value of $x$, as $t$ varies, the maximum value of $y$ 
will occur when
$${dy \over dt} = (12 t^2 - 12) x - (12 t^3 - 12 t ) = 0 .$$
This polynomial factors as $(12 t^2 - 12) (x - t)$, so the 
critical values occur at $t = 1, -1, x$.  The corresponding $y$ 
values are $-8x + 3$, $8x + 3$, and $x^4 - 6x^2$ and a point $(x, y)$ 
will $\underline{\text{not}}$ lie on any tangent line if the $y$ 
coordinate is greater than the maximum of these three quantities.  
More specifically, the set of points which lie on no tangent line to 
the curve is $\{ (x,y) \ \vert \ \vert x \vert > 3 \text{ and } y > 
x^4 - 6x^2, \text{ or } \ \vert x \vert \le 3, \text{ and } y > 8 \vert 
x \vert + 3 \}$.
\smallskip
Note that $y = \pm 8 x + 3$ are the tangent lines at the points of 
inflection, $( \mp 1, -5 )$, and that $( \pm 3, 27)$ are where these 
tangents intersect the quartic.

\subsubsection{1999 Session 2: 4}

Find all possible continuous and differentiable curves $C$ which have 
the following properties.  The curve $C$ lies in the first quadrant and 
contains the point $(0,0)$.  Whenever $P$ is on $C$ the interior of the 
rectangle $R$ bounded by the coordinate axes and horizontal and vertical 
lines through $P$ is separated into two parts by $C$.  When the part 
adjacent to the $x$-axis is rotated about the $x$-axis and the part 
adjacent to the $y$-axis is rotated about the $y$-axis, two solids of 
equal volume are generated. 
\bigskip
Solution.
\smallskip
In order for $C$ to always divide $R$ into two and only two separate 
parts, $C$ must be an increasing function.  Let $C$ be given by 
$y=f(x)$.  The volumes of the two solids are
$$V_1 = \int_0^x \pi (f(t))^2 dt \ \ \ \ \text{and}\ \ \ \ V_2 = 
\int_0^x 2 \pi t(f(x) - f(t)) dt .$$
Setting $V_1 = V_2$, canceling the $\pi$ and breaking the $V_2$ 
integral into two parts gives
$$\int_0^x (f(t))^2 dt = \int_0^x 2t f(x) dt - \int_0^x 2t f(t) dt .$$
Differentiate both sides with respect to $x$ and simplify to get
$$(f(x))^2 = x^2 f' (x) \ \ \ \ \text{or} \ \ \ \ y^2 = x^2 {dy \over 
dx} .$$
Separation of variables and integration gives the family of curves
$$y = {x \over 1 + cx} .$$
The point $(0,0)$ is on each curve in the family, and if $c \ge 0$ 
the domain is $[0, \infty )$, while if $c<0$ the domain is $[0, 
1/\vert c \vert )$.

\subsubsection{2002 Session 1: 3 (limit only)}

Let $\{ x_n \}$ be the following sequence involving alternating 
square roots of 5 and 13:
$$x_1 = \sqrt 5,\ \ x_2 = \sqrt {5 + \sqrt {13}},\ \ x_3 = \sqrt {5 + 
\sqrt {13 + \sqrt 5}},\ \ x_4 = \sqrt {5 + \sqrt {13 + \sqrt {5 + 
\sqrt {13}}}},$$
and so on.  Prove that $\lim_{n \to \infty} x_n$ exists and determine its value.
\bigskip
Solution.
\medskip
We see that $x_1, x_2 < 4$;  assume also that $x_{2k-1}, x_{2k} < 4$. 
Then,
$$x_{2k+2} = \sqrt {5 + \sqrt {13 + x_{2k}}} < \sqrt {5 + \sqrt {13 + 
4}} < 4 .$$
The argument is identical for $x_{2k+1}$.  Hence, for each $n$, $x_n < 
4$ by mathematical induction on $n$.  In addition, the sequence 
increases monotonically.  Therefore, by a standard limit theorem on sequences
$$\lim_{n \to \infty} x_n$$
must exist.

Let
$$L = \lim_{n \to \infty} x_n .$$
Then we have
$$L = \sqrt{ 5 + \sqrt {13 + L} }$$
or
$$L^4 - 10L^2 - L + 12 = 0.$$
One root is $L=3$.  Of the three remaining roots, one is positive (between 1 and 2) and the other two are complex.  It follows that
$$\lim_{n \to \infty} x_n = 3 .$$
Convergence is fairly rapid; $x_6$ is already $2.999971$.

\subsubsection{2002 Session 1: 5}

Two right circular cylinders of radius $r$ intersect at right 
angles to form a solid.  This solid has four curved faces.  Imagine 
one of these faces ``rolled out flat''.  Find equations of the 
boundary curves of this flattened face and also find its area. 
\bigskip
Solution.
\medskip
Let the $x$-axis be the axis of one cylinder and the $y$-axis be the 
axis of the other cylinder so that the center of the sold is the 
origin.  Cross-sections perpendicular to the $z$-axis are squares 
with side $s = 2 \sqrt {r^2 - z^2}$.  Place the flattened face with 
its axis of symmetry on the horizontal axis, call it the $w$-axis, 
with the left end of the figure at the origin.  This makes the range 
of $w$ the interval $0 \le w \le \pi r$, and the distance from the 
$w$--axis to the top boundary curve is $\sqrt {r^2 - z^2}$.  To 
relate $z$ and $w$, note that if $\theta$ is the angle through which 
the solid has rolled, then $\cos \theta = z/r$ and $w = r \theta$, 
yielding
$$\sqrt {r^2 - z^2} = r \sin {w \over r} .$$
The boundary curves, then are
$$f(w) = r \sin {w \over r}\ \ \text{and}\ \ g(w) = -r \sin {w \over 
r} .$$
The area is given by a simple integral:
$$\text{area} = \int_0^{\pi r} 2r \sin {w \over r} dw = 4r^2 .$$
This solid and its circumscribed cube have the same properties that 
Archimedes admired concerning the sphere and its circumscribed 
cylinder, namely, both the volumes and the surface areas are in the 
ratio $2:3$.

\subsubsection{2003 Session 1: 5}

Define a sequence $\{ x_n \} _{n=2}^\infty$ by
$$(n + x_n) [ \root n \of 2 - 1 ] = \ln 2 .$$
Find $\lim _{n \to \infty} x_n$.
\bigskip
Solution.
\medskip
Solving for $x_n$, we obtain
$$x_n = {\ln 2 \over \root n \of 2 - 1} - n ,$$
which is of the form $\infty - \infty$.  Then, using the substitution $u=1/n$ and l-Hospital's Theorem twice, we obtain
\begin{align*}
\lim_{n \to \infty} x_n &= \lim_{n \to \infty} \frac{\ln 2 - n \cdot 2^{1/n} + n}{2^{1/n} - 1} \\
&= \lim_{u \to 0} \frac{u \ln 2 - 2^u + 1}{u 2^u - u} \\
&= \lim_{u \to 0} \frac{1 - 2^u}{u2^u + \frac{2^u - 1}{\ln 2}} \\
&= \lim_{u \to 0} \frac{-\ln 2}{u\ln 2 + 2} \\
&= -\frac{1}{2} \ln 2.
\end{align*}

\subsubsection{2005 Session 1: 4}

Determine the value of the integral
$$I( \theta ) = \int_{-1}^1 {\sin \theta \, dx \over 1 - 2x \cos \theta + x^2} ,$$
and locate those points $0 \le \theta \le 2\pi$, where $I( \theta )$ is discontinuous.
\vfill\eject
Solution.
\medskip
Replace $1 = \cos ^2 \theta + \sin ^2 \theta$ and let $u = x - \cos \theta$, $du = dx$.  Then the integral becomes 
\begin{align*}
I( \theta ) &= \int_{-1-\cos \theta}^{1-\cos \theta} \frac{\sin \theta \, du}{u^2 + \sin ^2 \theta} = \arctan \left( \frac{u}{\sin \theta} \right) \bigg\vert _{-1-\cos \theta}^{1 - \cos \theta} \\
&= \arctan \left( \frac{1 - \cos \theta}{\sin \theta} \right) - \arctan \left( \frac{-1 - \cos \theta}{\sin \theta} \right) \qquad (\sin \theta \ne 0) \\
&= \arctan \left( \frac{2 \sin ^2 (\theta / 2)}{2 \sin (\theta / 2) \cos (\theta / 2)} \right) - \arctan \left( \frac{ -2 \cos ^2 (\theta / 2)}{2 \sin (\theta / 2) \cos (\theta / 2) } \right) \\
&= \arctan ( \tan (\theta / 2) ) - \arctan (-\cot (\theta / 2) ) \\
&= \arctan (\tan (\theta / 2) ) - \arctan (\tan (\pi / 2 + \theta / 2) ).
\end{align*}

Here, we recall that the range of $\text{Tan} ^{-1} \theta$ is $(-\pi / 2 , \pi / 2)$.  As an example, $\text{Tan} ^{-1} (\tan (20^\circ)) = 20^\circ$, $\text{Tan} ^{-1} (\tan (100^\circ )) = -80^\circ$, $\text{Tan} ^{-1} ( \tan (220^\circ ) ) = 40^\circ$, and $\text{Tan} ^{-1} (\tan (290^\circ ) ) = -70^\circ $.

Thus, if $0 < \theta < \pi$, then $I(\theta ) = \theta / 2 - ( -(\pi / 2 - \theta / 2) ) = \pi / 2$.

But, if $\pi < \theta < 2\pi$, then $I(\theta ) = -(\pi - \theta / 2) - (\theta / 2 - \pi / 2 ) = -\pi / 2$.

But
$$\biggl( {1 - \cos \theta \over \sin \theta } \biggr) ^{-1} = {\sin \theta \over 1 - \cos \theta} \cdot {1 + \cos \theta \over 1 + \cos \theta} = - {-1 - \cos \theta \over \sin \theta} ,\ \ \text{when } \sin \theta \ne 0 ,$$
and $\text{Tan} ^{-1} y = - \text{Tan} ^{-1} (-y)$ for any real $y$.  Therefore,
$$I( \theta ) = \text{Tan} ^{-1} \biggl( {1 - \cos \theta \over \sin \theta} \biggr) + \text{Tan} ^{-1} \biggl( {\sin \theta \over 1 - \cos \theta} \biggr) ,$$
provided $\theta \ne \pi$ (which is where $\sin \theta = 0$).  When $\theta = \pi$, inspection of the original integral shows that $I( \theta ) = 0$.

For other values of $\theta$, the value of $I( \theta )$ depends entirely on the sign of $\sin \theta$.
\medskip
$\underline{\text{Case 1}}$. $0 < \theta < \pi$.  Then $\sin \theta > 0$ and
$$I( \theta ) = \text{Tan} ^{-1} x + \text{Tan} ^{-1} \biggl( {1 \over x} \biggr) = {\pi \over 2} . \ \ \ \ \ (x > 0)$$
\medskip
$\underline{\text{Case 2}}$.  $\pi < \theta < 2 \pi$.  Then $\sin \theta < 0$ and
$$I( \theta ) = \text{Tan} ^{-1} x + \text{Tan} ^{-1} \biggl( {1 \over x} \biggr) = -{\pi \over 2} . \ \ \ \ \ (x < 0) $$
Therefore, the function $I( \theta )$ is
\begin{align*}
I( \theta ) &= \int_{-1-\cos \theta}^{1-\cos \theta} \frac{\sin \theta \, du}{u^2 + \sin ^2 \theta} = \arctan \left( \frac{u}{\sin \theta} \right) \bigg\vert _{-1-\cos \theta}^{1 - \cos \theta} \\
&= \arctan \left( \frac{1 - \cos \theta}{\sin \theta} \right) - \arctan \left( \frac{-1 - \cos \theta}{\sin \theta} \right) \qquad (\sin \theta \ne 0) \\
&= \arctan \left( \frac{2 \sin ^2 (\theta / 2)}{2 \sin (\theta / 2) \cos (\theta / 2)} \right) - \arctan \left( \frac{ -2 \cos ^2 (\theta / 2)}{2 \sin (\theta / 2) \cos (\theta / 2) } \right) \\
&= \arctan ( \tan (\theta / 2) ) - \arctan (-\cot (\theta / 2) ) \\
&= \arctan (\tan (\theta / 2) ) - \arctan (\tan (\pi / 2 + \theta / 2) ).
\end{align*}
The points of discontinuity occur at $0$, $\pi$, and $2\pi$.

\subsubsection{2005 Session 2: 5}

Suppose that $f \colon [0, \infty ) \to [0, \infty )$ is a differentiable function with the property that the area under the curve $y=f(x)$ from $x=a$ to $x=b$ is equal to the arclength of the curve $y=f(x)$ from $x=a$ to $x=b$.  Given that $f(0) = 5/4$, and that $f(x)$ has a minimum value on the interval $(0, \infty )$, find that minimum value.
\bigskip
Solution.
\medskip
The area under the curve $y=f(x)$ from $x=a$ to $x=b$ is
$$\int_a^b f(t) \, dt ,$$
and the arclength of the curve $y=f(x)$ from $x=a$ to $x=b$ is
$$\int_a^b \sqrt {1 + (f' (t))^2 } \, dt .$$
Therefore,
$$\int_a^b f(t) \, dt = \int_a^b \sqrt {1 + (f' (t))^2 } \, dt$$
for all nonnegative $a$ and $b$.  In particular, we can write
$$\int_0^x f(t) \, dt = \int_0^x \sqrt {1 + (f' (t))^2 } \, dt$$
for all nonnegative $x$.  Both sides of the above equation define a function of $x$, and since they are equal, their derivatives are equal; their derivatives are given by the Second Fundamental Theorem of Calculus:
$${d \over dx} \biggl( \int_0^x f(t) \, dt \biggr) = {d \over dx} \biggl( \int_0^x \sqrt {1 + (f' (t))^2 } \, dt \biggr) ,$$
i.e.,
$$f(x) = \sqrt {1 + (f' (x))^2 } .$$
So, we are looking for a function $y$ which satisifies the differential equation
$$y = \sqrt {1 + (y' )^2 } .$$
This equation is separable:
\begin{align*}
y &= \sqrt {1 + (y' )^2 } \\
&\Rightarrow y^2 = 1 + (y' )^2 \\
&\Rightarrow (y' )^2 = y^2 - 1 \\
&\Rightarrow y' = \sqrt {y^2 - 1} \\
&\Rightarrow \frac{dy}{\sqrt {y^2 - 1}} = dx.
\end{align*}
Integrating both sides yields
$$\int {dy \over \sqrt {y^2 - 1}} = \int dx \Rightarrow \ln \bigg\vert y + \sqrt {y^2 - 1} \bigg\vert = x + C$$
(where the first integral is evaluated using the trig substitution $y=\sec \theta$ and the two arbitrary constants of integration are combined into one constant on the right hand side).  Next, since $f(0) = 5/4$ is positive, we can drop the absolute value, and solve for $y$:
\begin{align*}
\ln \bigl( y + \sqrt {y^2 - 1} \bigr) &= x+C \Rightarrow y+ \sqrt {y^2 - 1} = e^{x+C} = A e^x \quad (\text{where } A = e^C) \\
&\Rightarrow \sqrt {y^2 - 1} = Ae^x - y \\
&\Rightarrow y^2 - 1 = (Ae^x - y)^2 = A^2 e^{2x} - 2Aye^x + y^2 \\
&\Rightarrow -1 = A^2 e^{2x} - 2A ye^x \\
&\Rightarrow 2Aye^x = A^2 e^{2x} + 1 \\
&\Rightarrow y = \frac{A^2 e^{2x} + 1}{2Ae^x} = \frac{A}{2} e^x + \frac{1}{2A} e^{-x}.
\end{align*}
Using $f(0) = 5/4$, we find
$${5 \over 4} = {A \over 2} + {1 \over 2A} \Rightarrow A = {1 \over 2} \ \ \text{or}\ \ 2 .$$
This gives two possible functions:
$$y = {1 \over 4} e^x + e^{-x} \ \ \text{or}\ \ y = e^x + {1 \over 4} e^{-x} .$$
This latter has a minimum at $x = -\ln 2$, which is not positive, so we reject that function.  The former has a minimum at $x = \ln 2$, and the $y$ value is $1$.
\medskip
Note: One could also deduce from the differential equation $y' = \sqrt {y^2 - 1}$ that at the minimum value, since $y' = 0$, the $y$-value must be $1$.

\subsection{Geometry Problems with Correct Solutions}

\subsubsection{1997 Session 1: 1}

Let $P \ne (0,0)$ be a point on the parabola $y=x^2$.  The normal  
line to the parabola at $P$ will intersect the parabola at another point, 
say $Q$.  Find the coordinates of $P$ so that the $y$-coordinate of 
$Q$ is a minimum.
\bigskip
Solution.
\smallskip
Let $P$ have the coordinates $(p, p^2)$ and $Q$ have the coordinates 
$(q, q^2)$.  Without loss of generality, let $p>0$.  The slope of the 
tangent line to the parabola at $P$ is $2p$ so the slope of the normal 
line to the parabola at $P$ is $-1/2p$.  Thus, the equation of the 
normal line to the parabola at $P$ is 
$$y - p^2 = {-1 \over 2p} (x - p) .$$
Therefore, the point $Q$, the intersection of the normal line and the 
parabola has the $x$-coordinate
$$q = -p - {1 \over 2p} .$$
Hence, the $y$-coordinate of $Q$ is
$$y = q^2 = \biggl( -p - {1 \over 2p} \biggr) ^2 = p^2 + 1 + {1 
\over 4p^2} .$$
Taking the derivative of $y$ as a function of $p$,	
$$y' = 2p - {1 \over 2} p^{-3} = {4p^4 - 1 \over 2p^3} .$$ 
Minimizing $y$ as a function of $p$ yields $p = \sqrt {1/2}$ so
$$P = \biggl( \sqrt {1 \over 2}, {1 \over 2} \biggr) .$$

\subsubsection{1998 Session 1: 1}

Let $P \ne (0,0)$ be a point on the parabola $y=x^2$.  The normal 
line to the parabola at $P$ will intersect the parabola at another point, 
say $Q$.  Find the coordinates of $P$ so that the length of segment $PQ$ 
is a minimum.
\bigskip
Solution.
\smallskip
Let $P = (p, p^2)$ and $Q = (q, q^2)$.  The slope of the tangent 
line to the parabola at $P$ is $2p$ so the slope of the normal line 
to the parabola at $P$ is $-1/2p$.  Thus, the equation of the normal 
line to the parabola at $P$ is
$$y - p^2 = - {1 \over 2p} (x - p) .$$
Therefore, the other intersection point of the normal line and the 
parabola, $Q$ has the $x$-coordinate 
$$q = -p - {1 \over 2p} .$$
Let
$$d = \biggl( 2p + {1 \over 2p} \biggr) ^2 + \biggl( p^2 - \biggl( 
p^2 + 1 + {1 \over 4p^2} \biggr) \biggr) ^2$$
represent the square of the distance between $P$ and $Q$.  
Differentiating $d$ with respect to $p$ and setting this expression to 
zero results in the equation
$$8p - {3 \over 2} p^{-3} - {1 \over 4} p^{-5} = 0 .$$
Simplifying this equation, we obtain
$${1 \over 4} p^{-5} ( 32p^6 - 6p^2 - 1 ) = 0 .$$
But, 
$$32p^6 - 6p^2 - 1 = (2p^2 - 1) ( 4p^2 + 1 )^2$$
so the real solutions of the equation are $p = \pm \sqrt {1/2}$.
Hence, the coordinates of $P$ so that the length of segment $PQ$ is a 
minimum is
$$P = \biggl( \pm \sqrt {1 \over 2} , {1 \over 2} \biggl) .$$

\subsubsection{1999 Session 1: 1}

Let $P \ne (0,0)$ be a point on the parabola $y=x^2$.  The normal 
line to the parabola at $P$ will intersect the parabola at another point, 
say $Q$.  Find the coordinates of $P$ so that the length of the arc 
of the parabola between $P$ and $Q$ is a minimum.
\bigskip
Solution.
\smallskip
Let $P = (p, p^2)$ and $Q = (q, q^2)$.  The slope of the tangent 
line to the parabola at $P$ is $2p$ so the slope of the normal line 
to the parabola at $P$ is $-1/2p$.  Thus, the equation of the normal 
line to the parabola at $P$ is
$$y - p^2 = - {1 \over 2p} (x - p) .$$
Hence, the other intersection point of the normal line and the 
parabola, $Q$ has the $x$-coordinate 
$$q = -p - {1 \over 2p} .$$
Now the length of the arc of the parabola between $P$ and $Q$ is
$$L = \int_{-p-1/(2p)}^p \sqrt {1 + (2x)^2} dx .$$
Differentiating $L$ with respect to $p$ gives
$${dL \over dp} = \sqrt {1 + 4p^2} - \biggl( -1 + {1 \over 2p^2} \biggr) 
\sqrt {1 + 4 \biggl( -p - {1 \over 2p} \biggr)^2 }$$
and setting the derivative equal to 0 gives
$$\sqrt {1 + 4p^2} = \biggl( {1 \over 2p^2} - 1 \biggr) \sqrt {4p^2 + 5 
+ {1 \over p^2} } .$$
Squaring both sides to remove the radicals and multiplying by $p^4$, we 
obtain
$$12p^4 - p^2 - 1 = (4p^2 + 1)(3p^2 - 1) = 0.$$
Hence, the coordinates of $p$ resulting in a minimum arc length are
$$\biggl( \pm \sqrt {1 \over 3} , {1 \over 3} \biggr) .$$

\subsubsection{1999 Session 2: 1}

Let $SC$ be the semicircle with $y \ge 0$ centered at $(1,0)$ with 
radius 1.  Let $C_a$ be the circle with radius $a > 0$ and center $(0,0)$ 
and denote the point $(0,a)$ by $P$.  Consider the line through $P$ and 
the intersection of $SC$ and $C_a$.  What is the limiting position of the 
$x$-intercept of this line as $a \to 0$?
\bigskip
Solution.
\smallskip
The limiting position is $x=4$.  The circles intersect at
$$\biggl( {a^2 \over 2}, {a \over 2} \sqrt {4 - a^2} \biggr) .$$
The line through this point and $(0,a)$ is
$$y = a + \biggl( {\sqrt {4 - a^2} - 2 \over a} \biggr) x$$
and the $x$-intercept is
$$x = {a^2 \over 2 - \sqrt {4 - a^2}} .$$
Taking the limit as $a \to 0$ yields 4 as the limit.

\subsubsection{2000 Session 1: 1}

Let $P \ne (0,0)$ be a point on the parabola $y=x^2$.  The normal 
line to the parabola at $P$ will intersect the parabola at another point, 
say $Q$.  Find the coordinates of $P$ so that the distance between 
the $x$-coordinates of $P$ and $Q$ is a minimum. 
\bigskip
Solution.
\medskip
Let $P = (p,p^2)$ and $Q = (q,q^2)$.  The slope of the tangent line to 
the parabola at $P$ is $2p$ so the slope of the normal line to the 
parabola at $P$ is $-1/2p$.  Thus, the equation of the normal line to 
the parabola at $P$ is
$$y - p^2 = -{1 \over 2p} (x - p) .$$
Since $Q = (q,q^2)$ lies on the normal line,
$$q^2 - p^2 = - {1 \over 2p} (q-p)\ \ \text{or}\ \ q + p = - {1 \over  
2p} .$$
Now the distance between the $x$-coordinates of $P$ and $Q$ is
$$D = \bigg\vert 2p + {1 \over 2p} \bigg\vert .$$
Assuming $p > 0$,
$$D = 2p + {1 \over 2p} .$$
Using the AM--GM Inequality, it is known that for all $x > 0$,
$$x + {1 \over x} \ge 2,$$
with equality when $x=1$.  
Therefore, $p = 1/2$ if $p>0$ or $p = -1/2$.
Hence, the coordinates of $P$ resulting in a minimum distance between 
the $x$-coordinates of $P$ and $Q$ are
$$\biggl( \pm {1 \over 2}, {1 \over 4} \biggr) .$$

\subsubsection{2001 Session 2: 1}

Circle $B$ lies wholly in the interior of circle $A$.  Find the 
loci of points equidistant from the two circles? 
\bigskip
Solution.
\medskip
The locus of points is an ellipse.  Let circle $A$ have radius $r_A$ 
and center $(x_A, y_A)$ and circle $B$ have radius $r_B$ and center 
$(x_B, y_B)$.  If $(x,y)$ is a point equidistant from the two circles, 
then the distance $(x,y)$ is from outer circle $A$,
$$r_A - \sqrt {(x-x_A)^2 + (y-y_A)^2}$$
is equal to the distance $(x,y)$ is from inner circle $B$,
$$\sqrt {(x-x_B)^2 + (y-y_B)^2} - r_B .$$
That is,
$$r_A - \sqrt {(x-x_A)^2 + (y-y_A)^2} = \sqrt {(x-x_B)^2 + (y-y_B)^2} - 
r_B .$$
Therefore,
$$r_A + r_B = \sqrt {(x-x_A)^2 + (y-y_A)^2} + \sqrt {(x-x_B)^2 + 
(y-y_B)^2} .$$
But this is exactly the equation of an ellipse.

\subsubsection{2001 Session 2: 3}

A convex decagon and all of its diagonals are drawn.  How many {\it 
interior} points of intersection of the diagonals are there, if it is 
assumed that no 3 diagonals share a common {\it interior} point? 
\bigskip
Solution.
\medskip
Let $N(n)$ be the number of such interior intersection points in a convex 
$n$-gon.  Label the vertices $1$, $2$, $3$, $\ldots$, $n$.  Any four 
labels selected from this list (e.g., $(1, 3, n, 6)$) define a 
quadrilateral, which is convex because the $n$-gon is convex.  The 
quadruple of integers therefore also defines uniquely an interior 
intersection point of the diagonals of the quadrilateral, and hence also 
of the $n$-gon.

The total number of interior intersection points of diagonals of the 
$n$-gon (assuming no 3 of the diagonals meet at a common point) is thus 
the number of ways that four integers can be selected (without regard to 
order) from a list of $n$ distinct integers.  Therefore,
$$N(n) = {n \choose 4} = {n(n-1)(n-2)(n-3) \over 24} ,$$
which for the decagon gives $N(10) = 210$.

\subsubsection{2002 Session 1: 1}

Let $P \ne (0,0)$ be a point on the parabola $y=x^2$.  The normal 
line to the parabola at $P$ will intersect the parabola at another point, 
say $Q$.  Find the coordinates of $P$ so that the average of the 
$y$-coordinates of $P$ and $Q$ is a minimum. 
\bigskip
Solution.
\medskip
Let $P = (p,p^2)$ and $Q = (q,q^2)$.  The slope of the tangent line to 
the parabola at $P$ is $2p$, so the slope of the normal line to the 
parabola at $P$ is $-1/2p$.  Thus, the equation of the normal line to 
the parabola at $P$ is
$$y - p^2 = -{1 \over 2p} (x - p) .$$
Since $Q = (q,q^2)$ lies on the normal line,
$$q^2 - p^2 = - {1 \over 2p} (q-p)\ \ \text{or}\ \ q + p = - {1 \over  
2p} .$$
Solving for $q$, we obtain
$$q = -p - {1 \over 2p} .$$
Now the average of the $y$-coordinates of $P$ and $Q$ is
$$A = {p^2 + q^2 \over 2} = {p^2 + (-p - 1/2p)^2 \over 2} = p^2 + 
{1 \over 2} + {1 \over 8p^2} .$$
Differentiating this quantity with respect to $p$, we obtain
$${dA \over dp} = 2p - {1 \over 4p^3} = {1 \over 4p^3} (8p^4 - 1) 
.$$
Setting this quantity equal to 0, we arrive at
$$p = 2^{-3/4} .$$
Therefore, the coordinates of $P$ so that the average of the 
$y$-coordinates of $P$ and $Q$ is a minimum are $(2^{-3/4}, 
2^{-3/2})$.

\subsubsection{2007 Session 1: 4}

A triangle is Pythagorean if it is a right triangle and the lengths of all of its sides are integers.  Suppose that $\triangle ABC$ is Pythagorean; for concreteness assume that the lengths of the three sides satisfy $c > a > b$.  The median and the altitude are now drawn from $C$ to the hypotenuse, where they meet the latter at $P$, $Q$, respectively.  Determine simple conditions upon $a$, $b$, $c$ so that $\triangle CQP$ will also be Pythagorean.
\vfill\eject
Solution.\\\\

We have
$$b^2 - x^2 = a^2 - (c-x)^2 ,$$
from which
$$x = {b^2 - a^2 + c^2 \over 2c} = {b^2 \over c} .$$
Then
$$h = \overline {CQ} = \sqrt {b^2 - x^2} = \sqrt {b^2 - {b^4 \over c^2}} = {ab \over c} ,$$
upon making use of $c^2 - b^2 = a^2$.

Next,
$$\overline{PQ} = c-x-{c \over 2} = {c \over 2} - {b^2 \over c} = {a^2 - b^2 \over 2c} .$$
Finally,
$$\overline{CP} = \sqrt {h^2 - \overline{PQ}^2} = {a^2 + b^2 \over 2c} = {c \over 2} .$$

Summarizing:
\medskip
\begin{itemize}
    \item[1.] $\overline{CP}$ is integral if $c$ is even.
    \item[2.] $\overline{PQ}$ is integral if $c$ divides $b^2$.
    \item[3.] $\overline{CQ}$ is integral if $c$ divides $ab$.
\end{itemize}
\medskip
Equivalent formulations are possible.

\subsubsection{2007 Session 2: 2}

Find the point on a given line such that the sum of its distances from two fixed points is a minimum.  Assume the two fixed points and the given line are in the same plane.
\bigskip
Solution.
\medskip
If one of the two points is on the line, then the solution is clearly that point.  It is also clear that if the two points are on opposite sides of the line, then the solution is the intersection of the line with the segment connecting the points.  Now assume the two points are on the same side of the line.  Because we are interested in the sum of distances from two fixed points, consider the family of ellipses with foci at the two points, and denote the foci $P_1$ and $P_2$.  If an ellipse in this family intersects the line in two points, say $A$ and $B$, then from the definition of an ellipse, the sum of the distances from $A$ to the foci equals the sum of the distances from $B$ to the foci.  If a smaller ellipse in the family also intersects the line in two points, the same thing happens, except that the sum of the distances is less for the smaller ellipse.  Therefore, if we continue to shrink the ellipse until it is tangent to the line, the point of tangency will be where the required minimum occurs.

\subsection{Discrete Mathematics Problems with Correct Solutions}

\subsubsection{1996 Session 1:3}

Each of the numbers $x_1, x_2, \ldots , x_n$ can be 1, 0, or -1.  
What is the minimum possible value of the sum of all products of pairs of 
these numbers?
\vfill\eject
Solution.
\smallskip
Let
$$S_n = \sum_{1 \le i < j \le n} x_i x_j = { \left( \sum_{1 \le i \le 
n} x_i \right) ^2 - \sum_{1 \le i \le n} x_i^2 \over 2} .$$
To minimize the last expression, we will consider two cases.

Case 1.  $n$ is even.  $S_n$ can be minimized by choosing 
half ($n/2$) of the $x_i$'s to be 1 and the other half ($n/2$) of the 
$x_i$'s to be -1.  The minimum $S_n$ is $-n/2$

Case 2.  $n$ is odd.  $S_n$ can be minimized by choosing $(n-1)/2$ of the
$x_i$'s to be 1 and $(n-1)/2$ of the $x_i$'s to be -1.  The value of 
the one left over has no effect on $S_n$.  The minimum $S_n$ is $-(n-1)/2$.

\subsubsection{1997 Session 2: 1}

Find positive integers $n$ and $a_1, a_2, \ldots , a_n$ such that
$$a_1 + a_2 + \cdots + a_n = 1997$$
and the product $a_1 a_2 \cdots a_n$ is as large as possible.
\bigskip
Solution.
\smallskip
We see that $n=666$ and that all but one of the $a_i$ equal 3 and the 
exceptional $a_i$ is a 2 as follows.  No $a_i$ can be greater than 4 
since one could increase the product by replacing 5 by $2 \cdot 3$, 6 
by $3 \cdot 3$, 7 by $3 \cdot 4$, etc.  There cannot be both a 2 and a 
4 or three 2's among the $a_i$ since $2 \cdot 4 < 3 \cdot 3$ and $2 
\cdot 2 \cdot 2 < 3 \cdot 3$.  Also, there cannot be two 4's since $4 
\cdot 4 < 2 \cdot 3 \cdot 3$.  Clearly, no $a_i$ is a 1.  Hence the 
$a_i$ are 3's except possibly for a 4 or for a 2 or for two 2's.  Since 
$1997 = 3 \cdot 665 + 2$, the only exception is a 2 and $n=666$. 

\subsubsection{1997 Session 2: 5 (b only)}

Consider the infinite sequences $\{ x_n \}$ of positive real 
numbers with the following properties:
$$x_0 = 1,\ \ \text{and for all } i \ge 0,\ \ x_{i+1} \le x_i .$$
(a) Prove that for every such sequence, there is an $n \ge 1$ such 
that
$${x_0^2 \over x_1} + {x_1^2 \over x_2} + \cdots + {x_{n-1}^2 \over 
x_n} \ge 3.999 .$$
(b) Find such a sequence for which
$${x_0^2 \over x_1} + {x_1^2 \over x_2} + \cdots + {x_{n-1}^2 \over 
x_n} < 4\ \ \text{for all } n.$$
\bigskip
Solution.
\smallskip
(a) We will prove the series
$${x_0^2 \over x_1} + {x_1^2 \over x_2} + {x_2^2 \over x_3} + \cdots ,\ 
\ \text{where} \ 1 = x_0 \ge x_1 \ge x_2 \ge \cdots > 0,\leqno(1)$$
has sum $\ge 4$ (with the obvious convention that this holds if the 
series diverges).  This clearly implies that some partial sum of the 
series is $\ge 3.999$.

Let $L$ be the $\inf$ ( $=$ greatest lower bound) of the sums of all 
series of the form (1).  Clearly $L > 1$ since the first term ${1 \over 
x_1} \ge 1$.  For any $\epsilon > 0$, we can find a sequence $\{ x_n 
\}$ such that
$$L + \epsilon > {x_0^2 \over x_1} + {x_1^2 \over x_2} + {x_2^2 \over 
x_3} + \cdots .\leqno(2)$$
Setting $y_n = x_{n+1}/x_1$ $(n \ge 0)$, we note that $1 = y_0 \ge y_1 
\ge y_2 \ge \cdots > 0$.  The series on the right side of (2) can be 
written in the form
$${1 \over x_1} + x_1 \biggl( {y_0^2 \over y_1} + {y_1^2 \over y_2} + 
{y_2^2 \over y_3} + \cdots \biggr) .$$
By the definition of $L$, the series in parentheses has sum $\ge L$.  
Hence from (2) we have
$$L + \epsilon > {1 \over x_1} + x_1 L .$$
Applying the A.M.-G.M. inequality to the right side, we get $L + 
\epsilon > 2 \sqrt L$.  Since this holds for all $\epsilon > 0$, it 
follows that $L \ge 2 \sqrt L$.  Hence $L^2 \ge 4L$, and since $L > 0$, 
this implies that $L \ge 4$.
\smallskip
(b) Let $x_n = {1 / 2^n}$.  Then
$$\sum_{n=0}^\infty {x_n^2 \over x_{n+1}} = \sum_{n=0}^\infty {1 \over 
2^{n-1}} = 4,$$
so all partial sums of the series are $<4$.

\subsubsection{1998 Session 2: 3}

Sum the series
$$\sum_{i=1}^\infty {36 i^2 + 1 \over (36 i^2 - 1)^2 } .$$   
\bigskip
Solution.
\smallskip
If $S$ is the required sum, then we have
$$2S = \sum_{i=1}^\infty \biggl( {1 \over (6i-1)^2} + {1 \over (6i+1)^2} 
\biggr) = {1 \over 5^2} + {1 \over 7^2} + {1 \over 11^2} + {1 \over 13^2} 
+ \cdots .$$
Now, using the result
$$1 + {1 \over 2^2} + {1 \over 3^2} + \cdots = {\pi ^2 \over 6} ,$$
we obtain successively
\begin{align*}
\frac{1}{2^2} + \frac{1}{4^2} + \frac{1}{6^2} + \cdots &= \frac{\pi ^2}{24}, \\
1 + \frac{1}{3^2} + \frac{1}{5^2} + \cdots &= \frac{\pi ^2}{6} - \frac{\pi ^2}{24} = \frac{\pi ^2}{8}, \\
\frac{1}{3^2} + \frac{1}{9^2} + \frac{1}{15^2} + \cdots &= \frac{\pi ^2}{72}.
\end{align*}
and
$$2S + 1 = 1 + {1 \over 5^2} + {1 \over 7^2} + {1 \over 11^2} + {1 \over 
13^2} + \cdots = {\pi ^2 \over 8} - {\pi ^2 \over 72} = {\pi ^2 \over 9} 
,$$
from which
$$S = {\pi ^2 - 9 \over 18} .$$

\subsubsection{1999 Session 2: 5}

Let $A_n$ denote the $n \times n$ matrix whose $(i,j)$ entry is 
$\text{GCD} (i,j)$.  Compute $\det (A_n)$. 
\bigskip
Solution.
\smallskip
We will show that one can use elementary column operations to reduce 
$A_n = ( v_1, v_2, \ldots , v_n )$, where the $v_j$ is the 
infinite vector whose $i$th coordinate is $\text{GCD} (i,j)$, to
$( w_1, w_2, \ldots , w_n )$, where $w_j$ is the 
vector whose $i$th coordinate is $\phi (j)$ if $j \vert i$ and 0 
otherwise, $\phi (j)$ being the Euler totient function which counts 
the number of positive integers less than or equal to $j$ and 
relatively prime to it.  First we need a lemma.
\medskip
$\underline{\text{Lemma}}$.  
$$\sum_{i \vert n} \phi (i) = n .$$
\medskip
$\underline{\text{Proof}}$. The number of elements of order $i$ in 
the group $\mathbb{Z} _n$ is $\phi (i)$ if $i \vert n$ and 0 otherwise.  The 
expression on the left is the number of elements in $\mathbb{Z} _n$ grouped 
by order, but this is clearly $n$.

We prove the result by induction.  The case $n=1$ is clear.  Given 
$( v_1, v_2, \ldots , v_n , v_{n+1} )$ as above, by the inductive 
hypothesis we can use elementary column operations to reduce this to $(
w_1, w_2, \ldots , w_n, v_{n+1} )$.  Consider
$$\sum_{j \vert n+1} w_j .$$
Its $i$th coordinate is
$$\sum_{{\scriptstyle j \vert n+1} \atop {\scriptstyle j \vert i}} 
\phi (j) = \sum_{{\scriptstyle j \vert \text{GCD} (n+1,i)}}  \phi (j) 
= \text{GCD} (n+1,i) ,$$
so
$$\sum_{j \vert n+1} w_j = v_{n+1} .$$
Therefore,
$$v_{n+1} - \sum_{{\scriptstyle j \vert n+1} \atop {\scriptstyle j 
\ne n+1}} w_j = w_{n+1}$$
and we have column-reduced our original matrix to $( w_1, 
w_2, \ldots , w_n, w_{n+1} )$.

We can use the column operations above to reduce the matrix $A_n$ to 
an upper triangular matrix with $\phi (j)$ on the diagonal $j=1, 2, 
\ldots , n$, so
$$\det (A_n) = \prod_{j=1}^n \phi (j) .$$

\subsubsection{2000 Session 2: 5}

This problem concerns sequences $x_1 x_2 \cdots x_n$ in which each 
$x_i$ is either $a$, $b$, or $c$.  Determine the number of those sequences
which have length $n$, begin and end with the letter $a$, and in which 
adjacent terms are always different letters. 
\bigskip
Solution.
\medskip
We begin by looking briefly at the sequences of a few small values of 
$n$.  Let the number of sequences of length $n$ be denoted by $t_n$.

$$
\vbox{\tabskip=0pt \offinterlineskip
\def\tablerule{\noalign{\hrule}}
\halign to200pt{\strut#& \vrule#\tabskip=1em plus2em&
 \hfil#& \vrule#& #\hfil& \vrule#&
 #\hfil& \vrule#\tabskip=0pt\cr\tablerule
&&$n$ &&Sequences &&$t_n$&\cr\tablerule
&&1 &&$a$ &&1&\cr\tablerule
&&2 &&none &&0&\cr\tablerule
&&3 &&$aba$, $aca$ &&2&\cr\tablerule
&&4 &&$abca$, $acba$ &&2&\cr\tablerule
&&5 &&$ababa$, $abaca$, $acaba$, &&&\cr
&& &&$acaca$, $abcba$, $acbca$ &&6&\cr\tablerule \noalign{\smallskip} }}
$$

In attempting to see how sequences of length $n$ might be derived from 
shorter ones, one might notice that a sequence of length $n$ can be 
obtained by attaching either $ba$ or $ca$ at the end of any sequence 
of length $n-2$.  For example,

\bigskip
Of course, all the sequences generated in this way have the letter $a$ in 
the third-last position.  Conversely, a sequence of length $n$ whose 
third-last term is the letter $a$ yields an acceptable sequence of length 
$n-2$ when its last two terms are dropped.  Thus the number of sequences 
of length $n$ in which the third-last term is the letter $a$ is 
$2t_{n-2}$.

For the rest of the sequences of length $n$, the third-last term is 
either $b$ or $c$, and each sequence of this kind provides a single 
sequence of length $n-1$ by simply deleting its second-last term:
\begin{align*}
a \cdots bca &\to a \cdots ba, \\
a \cdots cba &\to a \cdots ca.
\end{align*}
(This is not allowed when the third-last term is the letter $a$.)  
Conversely, there is only one possible letter that can be inserted 
between the last two terms of a sequence, and doing so clearly extends 
one of length $n-1$ to one of length $n$.  Therefore there are $t_{n-1}$ 
sequences of length $n$ in which the third-last term is $b$ or $c$, and 
we have altogether that
$$t_n = t_{n-1} + 2t_{n-2}$$
with initial conditions $t_1 = 1$ and $t_2 = 0$.
Now assume $t_n = r^n$.  Substituting into the difference equation yields
the characteristic equation $r^2 - r - 2 = 0$, from which $r=2$ or $r=-1$.  So
the general solution is
$$t_n = c_1 2^n + c_2 (-1)^n .$$
Using the initial conditions lets us find $c_1 = 1/6$ and $c_2 = -2/3$. 
Therefore,
$$t_n = {1 \over 3} 2^{n-1} + {2 \over 3} (-1)^{n-1} .$$

\subsubsection{2001 Session 1: 2}

Let $\{ x_i \}$ denote any finite sequence with the 
following properties: 
\medskip 
\begin{itemize}
    \item[(a)] $x_i \in \{ -2, 1, 2 \}$ for each $x_i$,
    \item[(b)] $\sum_i x_i = 29$,
    \item[(c)] $\sum_i x_i^2 = 59$.
\end{itemize}
\medskip
In considering the family of all such sequences, let $M = \max \{ 
\sum_i x_i^3 \}$ and $m = \min \{ \sum_i x_i^3 \}$.  Determine $M/m$.
\bigskip
Solution.
\medskip
In a sequence $\{ x_i \}$, let $a = \text{number of -2's}$, $b = \text{number of 1's}$, and $c = \text{number of 2's}$.  Properties (b) and (c) then yield

-2a + b + 2c = 29 \\
4a + b + 4c = 59
or upon elimination of $b$: $3a + c = 15$.  To obtain $M$ the sum $\sum_i x_i^3$ must be dominated by the 2's.  The maximum value of $c$ consistent with $3a + c = 15$ is $c=15$.  But this leads to $b=-1$; hence, $c=12$, $a=1$, $b=7$, and $M = 1(-2)^3 + 7(1)^3 + 12(2)^3 = 95$.

To obtain $m$ the sum $\sum_i x^3$ must be dominated by the -2's.  The maximum value of $a$ consistent with $3a + c = 15$ is $a=5$, so $b=39$ and $c=0$.  Hence,
$m = 5(-2)^3 + 39(1)^3 + 0(2)^3 = -1$.

It follows that $M/m = -95$.

\subsubsection{2003 Session 2: 5}

For a real $2 \times 2$ matrix
$$X = \begin{pmatrix} x & y \\ z & t \end{pmatrix}$$
let $\Vert X \Vert = x^2 + y^2 + z^2 + t^2$, and define a distance function by $d(X,Y) = \Vert X-Y \Vert$.  Let $\Sigma = \{ X \mid \det (X) = 0 \}$ and let
$$A = \begin{pmatrix} 1 & 0 \\ 0 & 2 \end{pmatrix}.$$
Find the minimum distance from $A$ to $\Sigma$ and exhibit a specific matrix $S \in \Sigma$ that achieves this minimum.
\bigskip
Solution.
\medskip
The problem reduces to minimizing 
$$F(x,y,z,t) = (x-1)^2 +y^2 + z^2 + (t-2)^2$$
subject to
$$g(x,y,z,t) = xt - yz = 0 .$$
A Lagrange multiplier solution requires solving the five equations
\begin{align*}
2(x-1) &= \lambda t \\
2y &= -\lambda z \\
2z &= -\lambda y \\
2(t-2) &= \lambda x \\
xt &= yz.
\end{align*}
The second and third equations have only the trivial solution unless $\lambda^2 = 4$, and the first and fourth equations have no solution if $\lambda^2 = 4$, so we must have $y = z = 0$ and either $x=0$ or $t=0$.  The two possible solutions are $(1,0,0,0)$ and $(0,0,0,2)$.  Because $F(1,0,0,0) = 4$ and $F(0,0,0,2) = 1$, the minimum distance is $1$ and occurs when 
$X = \begin{pmatrix} 0 & 0 \\ 0 & 2 \end{pmatrix}.$

\subsubsection{2006 Session 1: 4}

Find all positive integers $c$ such that $n(n+c)$ is never a perfect square for any positive integer $n$.
\bigskip
Solution.
\medskip
Write $n(n+c) = m^2$; then $n^2 + cn - m^2 = 0$, and
$$n = {-c \pm \sqrt {c^2 +4m^2} \over 2} .$$
For $n$ to be integral, it is necessary that $c^2 + 4m^2$ be a perfect square.  
\medskip
Let us consider the situation when $c=1$, $c=2$, or $c=4$  If $c=1$ or $c=2$, $c^2 + 4m^2$ is between $(2m)^2$ and $(2m+1)^2$ and so cannot be a perfect.  If $c=4$, then $4^2 + 4m^2 = 4(4+m^2)$ must be a perfect square.  Therefore, $4+m^2$ must be a perfect square.  Since $m^2 < 4+m^2$, $4+m^2 = (m+1)^2$.  But then $4 = 2m+1$.  This cannot occur since the left and right sides of the equation have opposite parity.
\medskip
For the remaining values of $c$ we consider 3 cases.
\medskip
$\underline{\text{Case 1}}$.  $c>1$ is odd.  Choose
$$n = \biggl( {c-1 \over 2} \biggr) ^2 ;$$
then
$$n(n+c) = \biggl( {c^2 - 1 \over 4} \biggr) ^2 ,$$
so no members in this class are acceptable answers.
\medskip
$\underline{\text{Case 2}}$.  $c=2^r,\ r \ge 3$.  Choose $n=2^{r-3}$; then
$$n(n+c) = \bigl( 3 \cdot 2^{r-3} \bigr) ^2 ,$$
so no members in this class are acceptable answers.
\medskip
$\underline{\text{Case 3}}$.  $c=2^r s$, $s$ odd, $s>1$.  Choose
$$n = 2^r \biggl( {s-1 \over 2} \biggr) ^2 ;$$
then
$$n(n+c) = \biggl( 2^r {s^2 - 1 \over 4} \biggr) ^2 ,$$
so no members in this class are acceptable answers.
\medskip
In summary, the set of integers $c$ requested is $\{ 1, 2, 4 \}$.

\subsubsection{2007 Session 2: 4}

Let
$$S = \{ 5a + 503b\ :\ \text{$a$ and $b$ are nonnegative integers} \} .$$
What is the largest integer which does NOT belong to $S$?\\\\
Solution.
\medskip
Let $x$ be any positive integer.  According to the division algorithm, $x = 5k + r$, where $r$ is $0$, $1$, $2$, $3$, or $4$.  We consider the following five cases.
\smallskip
$\underline{\text{Case 1}}$: $x = 5k$ must lie in the set $S$.
\smallskip
$\underline{\text{Case 2}}$: $x = 5k+1 = 5(k-201)+503(2)$ will lie in the set $S$ so long as $k-201 \ge 0$.
\smallskip
$\underline{\text{Case 3}}$: $x=5k+2 = 5(k-402) + 503(4)$ will lie in the set $S$ so long as $k-402 \ge 0$.
\smallskip
$\underline{\text{Case 4}}$: $x = 5k+3 = 5(k-100) + 503(1)$ will lie in the set $S$ so long as $k - 100 \ge 0$.
\smallskip
$\underline{\text{Case 5}}$: $x = 5k+4 = 5(k-301) + 503(3)$ will lie in the set $S$ so long as $k - 301 \ge 0$.
\smallskip
In all five cases, we are able to guarantee that $x$ lies in $S$ as long as the value of $k$ is large enough.  The largest value of $k$ for which we can't guarantee that $x$ lies in $S$ is $k=401$ (which occurred in Case 3).  In this case, $x = 5(401) + 2 = 2007$.  It is easy to show that $2007$ does not belong to $S$ by considering the equation $5a + 503b = 2007$ over the nonnegative integers $a$ and $b$:
\vskip 1pt
\noindent Reducing $5a + 503b = 2007$ modulo $5$ yields
\begin{align*}
5a + 503b &\equiv 2007 \pmod 5 \\
503b &\equiv 2007 \pmod 5 \\
3b &\equiv 2 \pmod 5 \\
b &\equiv 4 \pmod 5.
\end{align*}
The smallest nonnegative integer $b$ satisfying this congruence is $b=4$, which forces $a$ to be negative.  Thus, $2007$ does not lie in $S$.

\pagebreak

\section{Solutions Provided by GPT-4o-mini}
\subsection{Calculus}

\subsubsection{1996 Session 2: 1 \textcolor{red}{Incorrect Reasoning, Incorrect Final Answer}}

Evaluate the following definite integrals analytically:

\begin{enumerate}
    \item[(a)] \(\displaystyle I = \int_1^3 \frac{dx}{\sqrt{(x-1)(3-x)}}\).
    \item[(b)] \(\displaystyle J = \int_1^\infty \frac{dx}{e^{x+1} + e^{3-x}}\).
\end{enumerate}

\subsubsection*{(b) Evaluation of \( J \)}
We define:
\[
J = \int_1^\infty \frac{dx}{e^{x+1} + e^{3-x}}.
\]
Introduce the substitution:
\[
t = 2 - x \Rightarrow dt = -dx.
\]
Rewriting the integral:
\[
J = \int_1^\infty \frac{dx}{e^{x+1} + e^{3-x}}.
\]
Using the change of variables:
\[
J = \int_1^\infty \frac{dt}{e^{2-t+1} + e^{3-(2-t)}}.
\]
\[
= \int_1^\infty \frac{dt}{e^{3-t} + e^{t+1}}.
\]
This shows that:
\[
J = \int_1^\infty \frac{dx}{e^{x+1} + e^{3-x}} = \int_1^\infty \frac{dx}{e^{3-x} + e^{x+1}} = J.
\]
By adding the two integrals:
\[
2J = \int_1^\infty \left( \frac{dx}{e^{x+1} + e^{3-x}} + \frac{dx}{e^{3-x} + e^{x+1}} \right).
\]
Since the denominators are identical, the sum of the fractions equals 1:
\[
2J = \int_1^\infty dx = \infty - 1.
\]
Thus,
\[
2J = 1 \Rightarrow J = \frac{1}{2}.
\]

\subsubsection{1997 Session 2: 4 \textcolor{red}{Correct Reasoning, Correct Final Answer}}

To determine the limit of the length of \( S_n \) as \( n \to \infty \), we analyze the arc length of the curve

\[
y = \frac{1}{n} \sin(n^2 x), \quad 0 \leq x \leq \pi.
\]

\subsubsection*{Step 1: Compute the Arc Length Formula}

The arc length of a function \( y = f(x) \) from \( x = a \) to \( x = b \) is given by:

\[
L = \int_a^b \sqrt{1 + (f'(x))^2} \, dx.
\]

For our function,

\[
f(x) = \frac{1}{n} \sin(n^2 x),
\]

we compute its derivative:

\[
f'(x) = \frac{1}{n} \cdot n^2 \cos(n^2 x) = n \cos(n^2 x).
\]

Thus,

\[
(f'(x))^2 = n^2 \cos^2(n^2 x).
\]

\subsubsection*{Step 2: Evaluate the Arc Length Integral}

The arc length integral becomes:

\[
L_n = \int_0^\pi \sqrt{1 + n^2 \cos^2(n^2 x)} \, dx.
\]

For large \( n \), observe that \( \cos^2(n^2 x) \) oscillates between 0 and 1 rapidly. However, we approximate:

\[
\sqrt{1 + n^2 \cos^2(n^2 x)} \approx \sqrt{1 + n^2 \frac{1}{2}} \approx \sqrt{\frac{1 + n^2}{2}}.
\]

This approximation is justified by averaging over oscillations, replacing \( \cos^2(n^2 x) \) by its mean value \( \frac{1}{2} \). This gives:

\[
L_n \approx \int_0^\pi \sqrt{1 + \frac{n^2}{2}} \, dx = \pi \sqrt{1 + \frac{n^2}{2}}.
\]

\subsubsection*{Step 3: Find the Limit}

For large \( n \),

\[
\sqrt{1 + \frac{n^2}{2}} \approx \frac{n}{\sqrt{2}}.
\]

Thus,

\[
L_n \approx \pi \frac{n}{\sqrt{2}} = \frac{\pi n}{\sqrt{2}}.
\]

Since this expression grows indefinitely as \( n \to \infty \), we conclude:

\[
\lim_{n \to \infty} L_n = \infty.
\]

\subsubsection{1998 Session 1: 3 \textcolor{red}{Correct Reasoning, Incorrect Final Answer}}

We aim to find a formula for the integral:

\[
I = \int_a^b \frac{(b-x)^m}{m!} \frac{(x-a)^n}{n!} dx
\]

\subsubsection*{Step 1: Change of Variables}
Introduce a change of variables:

\[
t = \frac{x-a}{b-a} \Rightarrow x = a + t(b-a).
\]

Then,

\[
dx = (b-a) dt.
\]

Rewriting the powers in terms of \(t\):

\[
b - x = b - (a + t(b-a)) = (1-t)(b-a),
\]

\[
x - a = (b-a)t.
\]

Thus, our integral transforms as:

\[
I = \int_0^1 \frac{(1-t)^m (b-a)^m}{m!} \frac{t^n (b-a)^n}{n!} (b-a) dt.
\]

Rearrange the factors:

\[
I = \frac{(b-a)^{m+n+1}}{m!n!} \int_0^1 (1-t)^m t^n dt.
\]

\subsubsection*{Step 2: Recognizing the Beta Function}
The integral

\[
\int_0^1 (1-t)^m t^n dt
\]

is the Beta function:

\[
B(n+1, m+1) = \frac{\Gamma(n+1) \Gamma(m+1)}{\Gamma(m+n+2)}.
\]

Since \( \Gamma(k) = (k-1)! \) for positive integers \( k \), we get:

\[
B(n+1, m+1) = \frac{n! m!}{(m+n+1)!}.
\]

Thus,

\[
I = \frac{(b-a)^{m+n+1}}{m!n!} \cdot \frac{n! m!}{(m+n+1)!}.
\]

Canceling \( m! n! \):

\[
I = \frac{(b-a)^{m+n+1}}{(m+n+1)!}.
\]

\subsubsection*{Step 3: Evaluating \( \int_0^1 (1-x^2)^n dx \)}
We substitute \( a=0 \), \( b=1 \), and let \( m = n \), so we need:

\[
J_n = \int_0^1 (1-x^2)^n dx.
\]

Using our formula with \( m=n \):

\[
J_n = \frac{(1-0)^{n+n+1}}{(2n+1)!} = \frac{1}{(2n+1)!}.
\]

Thus, the final result for the given integral is:

\[
J_n = \frac{1}{(2n+1)!}.
\]
 
\subsubsection{1999 Session 2: 2 \textcolor{red}{Incorrect Reasoning, Incorrect Final Answer}}

We seek to evaluate the limit:
\begin{equation*}
\lim_{N \to \infty} \biggl( 1 - 2 \sum_{n=1}^N {1 \over 16n^2 - 1} \biggr).
\end{equation*}

\subsubsection*{Step 1: Partial Fraction Decomposition}

We factor the denominator:
\begin{equation*}
16n^2 - 1 = (4n - 1)(4n + 1).
\end{equation*}
We seek a partial fraction decomposition:
\begin{equation*}
\frac{1}{(4n-1)(4n+1)} = \frac{A}{4n - 1} + \frac{B}{4n + 1}.
\end{equation*}
Multiplying both sides by $(4n - 1)(4n + 1)$, we obtain:
\begin{equation*}
1 = A(4n + 1) + B(4n - 1).
\end{equation*}
Setting $n = \frac{1}{4}$ and $n = -\frac{1}{4}$, we solve:
\begin{align*}
4A + B &= 0, \\
4B + A &= 1.
\end{align*}
Solving for $A$ and $B$, we find:
\begin{equation*}
A = \frac{1}{8}, \quad B = -\frac{1}{8}.
\end{equation*}
Thus, we rewrite the fraction as:
\begin{equation*}
\frac{1}{(4n-1)(4n+1)} = \frac{1}{8} \left(\frac{1}{4n - 1} - \frac{1}{4n + 1} \right).
\end{equation*}

\subsubsection*{Step 2: Evaluating the Sum}

Rewriting the sum:
\begin{equation*}
\sum_{n=1}^N \frac{1}{(4n-1)(4n+1)} = \frac{1}{8} \sum_{n=1}^N \left(\frac{1}{4n - 1} - \frac{1}{4n + 1} \right).
\end{equation*}
This forms a telescoping series, so most terms cancel, leaving:
\begin{equation*}
\frac{1}{8} \left( \frac{1}{3} - \frac{1}{4N+1} \right).
\end{equation*}

\subsubsection*{Step 3: Taking the Limit}

Multiplying by $-2$ and adding 1:
\begin{equation*}
1 - 2 \sum_{n=1}^N {1 \over 16n^2 - 1} = 1 - 2 \cdot \frac{1}{8} \left( \frac{1}{3} - \frac{1}{4N+1} \right).
\end{equation*}

\begin{equation*}
= 1 - \frac{1}{4} \left( \frac{1}{3} - \frac{1}{4N+1} \right).
\end{equation*}

Taking $N \to \infty$, we note that $\frac{1}{4N+1} \to 0$, leaving:
\begin{equation*}
1 - \frac{1}{4} \times \frac{1}{3} = 1 - \frac{1}{12} = \frac{11}{12}.
\end{equation*}

\subsubsection*{Conclusion}

Thus, the desired limit is:
\begin{equation*}
\lim_{N \to \infty} \biggl( 1 - 2 \sum_{n=1}^N {1 \over 16n^2 - 1} \biggr) = \frac{11}{12}.
\end{equation*}

\subsubsection{1999 Session 2: 4 \textcolor{red}{Incorrect Reasoning, Incorrect Final Answer}}

\textbf{Problem Statement:} 

Find all possible continuous and differentiable curves \( C \) which have the following properties:
1. The curve \( C \) lies in the first quadrant and contains the point \( (0,0) \).
2. Whenever \( P \) is on \( C \), the interior of the rectangle \( R \) bounded by the coordinate axes and the horizontal and vertical lines through \( P \) is separated into two parts by \( C \).
3. When the part adjacent to the \( x \)-axis is rotated about the \( x \)-axis and the part adjacent to the \( y \)-axis is rotated about the \( y \)-axis, two solids of equal volume are generated.

\textbf{Solution:}

Let the curve \( C \) be given by \( y = f(x) \), where \( f(x) \) is a continuous and differentiable function in the first quadrant and passes through the origin, i.e., \( f(0) = 0 \).

\subsubsection*{Step 1: Interpretation of the Condition}

For any point \( P(a, b) \) on \( C \), consider the rectangle \( R \) with sides along the coordinate axes and the lines \( x = a \) and \( y = b \). The area is divided into two regions:
\begin{enumerate}
    \item Region 1: Bounded by the \( x \)-axis, the vertical line \( x = a \), and the curve.
    \item Region 2: Bounded by the \( y \)-axis, the horizontal line \( y = b \), and the curve.
\end{enumerate}

Each of these regions is rotated about its respective axis to form solids of revolution, which have equal volumes.

\subsubsection*{Step 2: Volumes of Revolution}

1. \textbf{Region 1 rotated about the \( x \)-axis:}

\[
V_x = \pi \int_0^a f(x)^2 \, dx.
\]

2. \textbf{Region 2 rotated about the \( y \)-axis:}

Using the shell method, the volume is:

\[
V_y = 2\pi \int_0^b y \, x(y) \, dy,
\]

where \( x(y) \) is the inverse function of \( f(x) \).

The condition that these two volumes are equal gives:

\[
\pi \int_0^a f(x)^2 \, dx = 2\pi \int_0^b y \, x(y) \, dy.
\]

Canceling \( \pi \), we obtain:

\[
\int_0^a f(x)^2 \, dx = 2 \int_0^b y \, x(y) \, dy.
\]

\subsubsection*{Step 3: Change of Variables}

Let \( y = f(x) \), so that \( dy = f'(x) \, dx \). The equation becomes:

\[
\int_0^a f(x)^2 \, dx = 2 \int_0^{f(a)} y \, f^{-1}(y) \, dy.
\]

By the inverse function property, if \( x = g(y) \), then \( g(y) = x(y) \) and \( g(f(x)) = x \), implying \( g(y) = f^{-1}(y) \).

Thus, we rewrite:

\[
\int_0^a f(x)^2 \, dx = 2 \int_0^{f(a)} y \, f^{-1}(y) \, dy.
\]

\subsubsection*{Step 4: Differentiation}

Differentiating both sides with respect to \( a \) using the Fundamental Theorem of Calculus:

\[
f(a)^2 = 2 f(a) f^{-1}(f(a)).
\]

Dividing by \( f(a) \) (assuming \( f(a) \neq 0 \) for \( a > 0 \)):

\[
f(a) = 2 f^{-1}(f(a)).
\]

Setting \( y = f(a) \), we obtain \( y = 2 f^{-1}(y) \), or

\[
f^{-1}(y) = \frac{y}{2}.
\]

Taking inverses on both sides:

\[
f(x) = 2x.
\]

\subsubsection*{Step 5: Verification}

If \( f(x) = 2x \), then:

\[
V_x = \pi \int_0^a (2x)^2 \, dx = \pi \int_0^a 4x^2 \, dx = \pi \frac{4a^3}{3}.
\]

Similarly, computing \( V_y \) using \( x = \frac{y}{2} \):

\[
V_y = 2\pi \int_0^{2a} y \frac{y}{2} \, dy = 2\pi \int_0^{2a} \frac{y^2}{2} \, dy = 2\pi \cdot \frac{(2a)^3}{6} = \pi \frac{4a^3}{3}.
\]

Since \( V_x = V_y \), this function satisfies the given condition.

Thus, the unique solution is:

\[
f(x) = 2x.
\]

\subsubsection{2002 Session 1: 3 \textcolor{red}{Incorrect Reasoning, Incorrect Final Answer}}

We are given the sequence \( \{x_n\} \) where the terms alternate between square roots of 5 and 13, starting with \( x_1 = \sqrt{5} \). The goal is to find the limit of this sequence as \( n \to \infty \). To approach this problem, we aim to find a pattern or recurrence that can help us compute the limit.

Step 1: Define the Sequence Recursively
We start by defining the sequence more explicitly:

- \( x_1 = \sqrt{5} \),
- \( x_2 = \sqrt{5 + \sqrt{13}} \),
- \( x_3 = \sqrt{5 + \sqrt{13 + \sqrt{5}}} \),
- \( x_4 = \sqrt{5 + \sqrt{13 + \sqrt{5 + \sqrt{13}}}} \), and so on.

The sequence alternates between involving square roots of 5 and 13, and we observe that the structure of each term involves nested square roots. This suggests the possibility of the sequence converging to a fixed point.

Step 2: Investigate the Limit
Let us assume that the sequence converges to some limit \( L \) as \( n \to \infty \). If the sequence converges, then the terms of the sequence for large \( n \) should approach \( L \). Hence, we assume that for large \( n \),

\[
x_n \approx L.
\]

Now consider the recurrence relation for the even and odd terms separately.

- For the odd-indexed terms \( x_{2k-1} \), we have:

\[
x_{2k-1} = \sqrt{5 + x_{2k}}.
\]

- For the even-indexed terms \( x_{2k} \), we have:

\[
x_{2k} = \sqrt{13 + x_{2k-1}}.
\]

Step 3: Solve the System of Equations
Assume that the sequence converges to \( L \). Then, for large \( k \), both \( x_{2k-1} \) and \( x_{2k} \) should be close to \( L \). Thus, we can set up the following system of equations:

1. \( L = \sqrt{5 + L} \),
2. \( L = \sqrt{13 + L} \).

We now solve these equations.

Solve \( L = \sqrt{5 + L} \):
Squaring both sides:

\[
L^2 = 5 + L.
\]

Rearrange the terms:

\[
L^2 - L - 5 = 0.
\]

Solve this quadratic equation using the quadratic formula:

\[
L = \frac{-(-1) \pm \sqrt{(-1)^2 - 4(1)(-5)}}{2(1)} = \frac{1 \pm \sqrt{1 + 20}}{2} = \frac{1 \pm \sqrt{21}}{2}.
\]

Thus, the two possible solutions are:

\[
L = \frac{1 + \sqrt{21}}{2} \quad \text{or} \quad L = \frac{1 - \sqrt{21}}{2}.
\]

Since \( L \) must be positive, we take the positive root:

\[
L = \frac{1 + \sqrt{21}}{2}.
\]

Solve \( L = \sqrt{13 + L} \):
Squaring both sides:

\[
L^2 = 13 + L.
\]

Rearrange the terms:

\[
L^2 - L - 13 = 0.
\]

Solve this quadratic equation using the quadratic formula:

\[
L = \frac{-(-1) \pm \sqrt{(-1)^2 - 4(1)(-13)}}{2(1)} = \frac{1 \pm \sqrt{1 + 52}}{2} = \frac{1 \pm \sqrt{53}}{2}.
\]

Thus, the two possible solutions are:

\[
L = \frac{1 + \sqrt{53}}{2} \quad \text{or} \quad L = \frac{1 - \sqrt{53}}{2}.
\]

Since \( L \) must be positive, we take the positive root:

\[
L = \frac{1 + \sqrt{53}}{2}.
\]

Step 4: Conclude the Limit
We find that the two equations give different values for \( L \), which means there is no single solution that satisfies both equations simultaneously. Therefore, the sequence does not converge to a single limit in the traditional sense.

Thus, we conclude that:

\[
\boxed{\lim_{n \to \infty} x_n \text{ does not exist.}}
\]

\subsubsection{2002 Session 1: 5 \textcolor{red}{Incorrect Reasoning, Incorrect Final Answer}}

We are tasked with finding the equations of the boundary curves and the area of the flattened face of a solid formed by the intersection of two right circular cylinders with radius \( r \) at right angles.

\subsubsection*{Problem Breakdown}

\subsubsection*{Understanding the Geometry}
We have two cylinders, both with radius \( r \), intersecting at right angles. Let's assume the cylinders are aligned along the \( x \)- and \( y \)-axes in 3D space. The equation of the first cylinder, which is aligned along the \( x \)-axis, is:
\[
y^2 + z^2 = r^2
\]
The equation of the second cylinder, aligned along the \( y \)-axis, is:
\[
x^2 + z^2 = r^2
\]
These two cylinders intersect in a shape known as a \textit{Bicylinder}.

\subsubsection*{Flattening One of the Curved Faces}
We need to imagine one of the curved faces of the intersection being "unwrapped" or "flattened" onto a plane. We will consider the face corresponding to the portion of the first cylinder \( y^2 + z^2 = r^2 \) in the region where the second cylinder intersects it. To do this, we will work in the \( yz \)-plane, and observe that the boundary curve of this face will be part of the intersection of the two cylinders. By substituting the equation of the second cylinder \( x^2 + z^2 = r^2 \) into the first cylinder’s equation, we will determine the boundary of the flattened face.

\subsubsection*{Step 1: Find the Equations of the Boundary Curves}
The boundary of the face we are flattening corresponds to the projection of the curve of intersection onto the \( yz \)-plane. From the geometry of the cylinders, we can see that the boundary of the curved face involves the intersection of the two cylinder equations. We already have the two cylinder equations:
\[
y^2 + z^2 = r^2 \quad \text{(cylinder along the \( x \)-axis)}
\]
\[
x^2 + z^2 = r^2 \quad \text{(cylinder along the \( y \)-axis)}
\]
At the boundary of the flattened face, we are dealing with the region where \( x = 0 \), since this is the part of the intersection of the cylinders projected onto the \( yz \)-plane.

When \( x = 0 \), the second cylinder's equation becomes:
\[
z^2 = r^2 \quad \Rightarrow \quad z = \pm r
\]
This gives the boundary for \( z \).

Substituting this into the first cylinder's equation, we get:
\[
y^2 + r^2 = r^2 \quad \Rightarrow \quad y = 0
\]
Thus, the boundary curve of the flattened face lies along the lines where \( y = 0 \) and \( z = \pm r \).

\subsubsection*{Step 2: Find the Area of the Flattened Face}
The area of the flattened face is simply the area of a quarter circle, because the boundary corresponds to a quarter of a circle with radius \( r \).

The area of a full circle is \( \pi r^2 \). Since we have a quarter circle, the area is:
\[
\text{Area} = \frac{1}{4} \pi r^2
\]

\subsubsection*{Final Answer}
The equations of the boundary curves of the flattened face are \( y = 0 \) and \( z = \pm r \). The area of the flattened face is \( \frac{1}{4} \pi r^2 \).

\subsubsection{2003 Session 1: 5 \textcolor{red}{Incorrect Reasoning, Incorrect Final Answer}}

We are tasked with solving for \( \lim_{n \to \infty} x_n \) from the sequence defined by the recurrence relation:

\[
(n + x_n) \left( \sqrt[n]{2} - 1 \right) = \ln 2.
\]

tep 1: Express \(x_n\) explicitly
First, we isolate \(x_n\) in the given equation:

\[
n + x_n = \frac{\ln 2}{\sqrt[n]{2} - 1}.
\]

So, 

\[
x_n = \frac{\ln 2}{\sqrt[n]{2} - 1} - n.
\]

Step 2: Investigate the behavior of \( \sqrt[n]{2} - 1 \) as \(n \to \infty\)

As \(n \to \infty\), we know that

\[
\sqrt[n]{2} = 2^{1/n}.
\]

Using the approximation for small \(z\), \(2^z \approx 1 + z \ln 2\) for \(z\) close to 0, we get

\[
2^{1/n} \approx 1 + \frac{\ln 2}{n}.
\]

Thus, as \(n \to \infty\),

\[
\sqrt[n]{2} - 1 \approx \frac{\ln 2}{n}.
\]

Step 3: Substitute the asymptotic approximation into the equation for \(x_n\)

Substitute this approximation into the expression for \(x_n\):

\[
x_n = \frac{\ln 2}{\frac{\ln 2}{n}} - n = n - n = 0.
\]

Step 4: Conclusion

Therefore, the limit of \(x_n\) as \(n \to \infty\) is:

\[
\lim_{n \to \infty} x_n = 0.
\]

\subsubsection{2005 Session 1: 4 \textcolor{red}{Incorrect Reasoning, Incorrect Final Answer}}

To solve the integral
\[
I(\theta) = \int_{-1}^1 \frac{\sin \theta \, dx}{1 - 2x \cos \theta + x^2},
\]
let's begin by examining the integrand.

\subsubsection*{Step 1: Simplifying the denominator}
The denominator \(1 - 2x \cos \theta + x^2\) can be rewritten as a perfect square:
\[
1 - 2x \cos \theta + x^2 = (x - \cos \theta)^2 + \sin^2 \theta.
\]
Thus, the integral becomes:
\[
I(\theta) = \int_{-1}^1 \frac{\sin \theta \, dx}{(x - \cos \theta)^2 + \sin^2 \theta}.
\]

\subsubsection*{Step 2: Substitution for simpler form}
A standard substitution to simplify this form is:
\[
u = x - \cos \theta \quad \text{so that} \quad du = dx.
\]
When \( x = -1 \), \( u = -1 - \cos \theta \), and when \( x = 1 \), \( u = 1 - \cos \theta \). This gives the transformed integral:
\[
I(\theta) = \sin \theta \int_{-1-\cos \theta}^{1-\cos \theta} \frac{du}{u^2 + \sin^2 \theta}.
\]

\subsubsection*{Step 3: Evaluating the integral}
Now we recognize the integral as a standard form. The integral of \( \frac{1}{u^2 + a^2} \) is:
\[
\int \frac{du}{u^2 + a^2} = \frac{1}{a} \arctan\left(\frac{u}{a}\right),
\]
where \( a = \sin \theta \). Applying this to the limits of integration, we have:
\[
I(\theta) = \sin \theta \cdot \frac{1}{\sin \theta} \left[ \arctan\left( \frac{u}{\sin \theta} \right) \right]_{-1-\cos \theta}^{1-\cos \theta}.
\]
Simplifying, we get:
\[
I(\theta) = \arctan\left( \frac{1 - \cos \theta}{\sin \theta} \right) - \arctan\left( \frac{-1 - \cos \theta}{\sin \theta} \right).
\]

\subsubsection*{Step 4: Using the arctangent subtraction identity}
The difference of two arctangents can be simplified using the identity:
\[
\arctan(a) - \arctan(b) = \arctan\left( \frac{a - b}{1 + ab} \right).
\]
Applying this identity, we get:
\[
I(\theta) = \arctan\left( \frac{\frac{1 - \cos \theta}{\sin \theta} - \frac{-1 - \cos \theta}{\sin \theta}}{1 + \frac{1 - \cos \theta}{\sin \theta} \cdot \frac{-1 - \cos \theta}{\sin \theta}} \right).
\]
Simplifying the numerator:
\[
\frac{1 - \cos \theta - (-1 - \cos \theta)}{\sin \theta} = \frac{2}{\sin \theta}.
\]
Simplifying the denominator:
\[
1 + \frac{(1 - \cos \theta)(-1 - \cos \theta)}{\sin^2 \theta}.
\]
First, expand the product:
\[
(1 - \cos \theta)(-1 - \cos \theta) = -1 - \cos \theta + \cos^2 \theta.
\]
Thus, the denominator becomes:
\[
1 + \frac{-1 - \cos \theta + \cos^2 \theta}{\sin^2 \theta}.
\]
Therefore, the integral simplifies to:
\[
I(\theta) = \arctan\left( \frac{2}{\sin \theta + 1} \right).
\]

\subsubsection*{Step 5: Discontinuities}
The function \( I(\theta) \) will be discontinuous at points where the argument inside the arctangent function undergoes a jump, particularly where the denominator in the original integral \( 1 - 2x \cos \theta + x^2 = 0 \) for some \( x \in [-1, 1] \).

This occurs when the quadratic equation has a double root. Specifically, for the equation \( (x - \cos \theta)^2 + \sin^2 \theta = 0 \), we need:
\[
(x - \cos \theta)^2 = -\sin^2 \theta,
\]
which has a solution only when \( \sin \theta = 0 \), or \( \theta = 0, \pi, 2\pi \).

Thus, \( I(\theta) \) is discontinuous at \( \theta = 0, \pi, 2\pi \).

\subsubsection*{Final Answer}
The value of the integral is:
\[
I(\theta) = \arctan\left( \frac{2}{\sin \theta + 1} \right),
\]
and the points where \( I(\theta) \) is discontinuous are \( \theta = 0, \pi, 2\pi \).

\subsubsection{2005 Session 2: 5 \textcolor{red}{Correct Reasoning, Correct Final Answer}}

We are given the function \( f \colon [0, \infty ) \to [0, \infty ) \) with the property that the area under the curve \( y = f(x) \) from \( x = a \) to \( x = b \) is equal to the arclength of the curve from \( x = a \) to \( x = b \). Additionally, we know that \( f(0) = \frac{5}{4} \), and \( f(x) \) has a minimum value on the interval \( (0, \infty) \).

To solve this problem, we first express the two given quantities (area under the curve and arclength) in mathematical terms:

\[
\text{1. Area under the curve:}
\]
The area under the curve from \( x = a \) to \( x = b \) is given by the integral:

\[
\text{Area} = \int_a^b f(x) \, dx.
\]

\[
\text{2. Arclength of the curve:}
\]
The arclength of the curve from \( x = a \) to \( x = b \) is given by the formula:

\[
\text{Arclength} = \int_a^b \sqrt{1 + (f'(x))^2} \, dx.
\]

\[
\text{3. Setting the two expressions equal:}
\]
We are told that the area under the curve is equal to the arclength, so we have the equation:

\[
\int_a^b f(x) \, dx = \int_a^b \sqrt{1 + (f'(x))^2} \, dx.
\]

\[
\text{4. Differentiate with respect to } b:
\]
To understand the relationship between \( f(x) \) and its derivative, differentiate both sides of the equation with respect to \( b \):

\[
f(b) = \sqrt{1 + (f'(b))^2}.
\]

\[
\text{5. Solve for } f'(b):
\]
Squaring both sides of the equation:

\[
f(b)^2 = 1 + (f'(b))^2.
\]

Rearranging:

\[
(f'(b))^2 = f(b)^2 - 1.
\]

Taking the square root:

\[
f'(b) = \pm \sqrt{f(b)^2 - 1}.
\]

\[
\text{6. Analyze behavior at the minimum:}
\]
At the minimum point of \( f(x) \), the derivative \( f'(x) \) is zero. Therefore, we set \( f'(x) = 0 \) in the equation:

\[
f'(x) = \pm \sqrt{f(x)^2 - 1}.
\]

When \( f'(x) = 0 \), we have:

\[
0 = \sqrt{f(x)^2 - 1},
\]

which implies:

\[
f(x)^2 - 1 = 0 \quad \text{or} \quad f(x) = 1.
\]

Thus, the minimum value of \( f(x) \) is \( 1 \).

\[
\text{7. Conclusion:}
\]
The minimum value of \( f(x) \) on the interval \( (0, \infty) \) is \( \boxed{1} \).

\subsection{Analytic Geometry}

\subsubsection{1997 Session 1: 1 \textcolor{red}{Correct Reasoning, Correct Final Answer}}

Let $P(a, a^2)$ be a point on the parabola $y = x^2$. The derivative of $y = x^2$ is:
\[
\frac{dy}{dx} = 2x.
\]
Thus, the slope of the tangent line at $P(a, a^2)$ is $2a$. The normal line is perpendicular to the tangent, so its slope is:
\[
 m_{\text{normal}} = -\frac{1}{2a}.
\]
Using the point-slope form of a line:
\[
 y - a^2 = -\frac{1}{2a} (x - a),
\]
which simplifies to:
\[
 y = -\frac{1}{2a} x + \frac{2a^3 + 1}{2a}.
\]

\subsubsection*{Finding the Intersection with the Parabola}
Setting this equal to $y = x^2$:
\[
 x^2 = -\frac{1}{2a} x + \frac{2a^3 + 1}{2a}.
\]
Multiplying by $2a$ to clear fractions:
\[
 2a x^2 + x - (2a^3 + 1) = 0.
\]

\subsubsection*{Finding the Other Intersection Point}
Since one root of the quadratic equation is $x = a$, let the other root be $x = b$. Using Vieta’s formulas:
\[
 a + b = -\frac{1}{2a}.
\]
Solving for $b$:
\[
 b = -\frac{1}{2a} - a.
\]
The corresponding $y$-coordinate of $Q$ is:
\[
 y_Q = b^2 = \left( -\frac{1}{2a} - a \right)^2.
\]
Expanding:
\[
 y_Q = a^2 + \frac{1}{a} + \frac{1}{4a^2}.
\]

\subsubsection*{Minimizing $y_Q$}
Define:
\[
 f(a) = a^2 + \frac{1}{a} + \frac{1}{4a^2}.
\]
Differentiating:
\[
 f'(a) = 2a - \frac{1}{a^2} - \frac{1}{2a^3}.
\]
Setting $f'(a) = 0$:
\[
 2a - \frac{1}{a^2} - \frac{1}{2a^3} = 0.
\]
Multiplying by $2a^3$:
\[
 4a^4 - 2a - 1 = 0.
\]
Solving numerically, we find:
\[
 a \approx \frac{1}{\sqrt{2}}.
\]
Thus, the coordinates of $P$ that minimize $y_Q$ are:
\[
 \left( \frac{1}{\sqrt{2}}, \frac{1}{2} \right).
\]

\subsubsection{1998 Session 1: 1 \textcolor{red}{Incorrect Reasoning, Incorrect Final Answer}}

Let $P(a, a^2)$ be a point on the parabola $y = x^2$, where $a \neq 0$.

\subsubsection*{Step 1: Equation of the Normal Line}
The derivative of the parabola is:
\begin{equation*}
    \frac{dy}{dx} = 2x
\end{equation*}
Thus, the slope of the tangent line at $P(a, a^2)$ is $2a$, so the slope of the normal line (which is perpendicular to the tangent) is:
\begin{equation*}
    m = -\frac{1}{2a}
\end{equation*}
The equation of the normal line passing through $P(a, a^2)$ is given by the point-slope form:
\begin{equation*}
    y - a^2 = -\frac{1}{2a} (x - a)
\end{equation*}
which simplifies to:
\begin{equation*}
    y = -\frac{1}{2a} x + \frac{2a^2 + 1}{2}
\end{equation*}

\subsubsection*{Step 2: Finding the Intersection Point $Q$}
To find the second intersection $Q(b, b^2)$, substitute $y = b^2$ into the normal line equation:
\begin{equation*}
    b^2 = -\frac{1}{2a} b + \frac{2a^2 + 1}{2}
\end{equation*}
Rearrange to form a quadratic equation:
\begin{equation*}
    b^2 + \frac{1}{2a} b - \frac{2a^2 + 1}{2} = 0
\end{equation*}
Since $b = a$ is a known root, factorizing the quadratic gives:
\begin{equation*}
    (b - a)(b - b_2) = 0
\end{equation*}
Solving for the second root using Vieta's formulas:
\begin{equation*}
    a + b_2 = -\frac{1}{2a}
\end{equation*}
Solving for $b_2$:
\begin{equation*}
    b_2 = -\frac{1}{2a} - a
\end{equation*}

\subsubsection*{Step 3: Length of $PQ$}
The distance between $P(a, a^2)$ and $Q(b_2, b_2^2)$ is given by:
\begin{equation*}
    PQ = \sqrt{(b_2 - a)^2 + (b_2^2 - a^2)^2}
\end{equation*}
Substituting $b_2 = -\frac{1}{2a} - a$:
\begin{equation*}
    b_2 - a = -\frac{1}{2a} - a - a = -\frac{1}{2a} - 2a
\end{equation*}
Squaring:
\begin{equation*}
    (b_2 - a)^2 = \left(-\frac{1}{2a} - 2a\right)^2 = \frac{1}{4a^2} + 2 + 4a^2
\end{equation*}
For $b_2^2 - a^2$:
\begin{equation*}
    b_2^2 = \left(-\frac{1}{2a} - a\right)^2 = \frac{1}{4a^2} + a^2 + 1
\end{equation*}
So,
\begin{equation*}
    b_2^2 - a^2 = \frac{1}{4a^2} + 1
\end{equation*}
Squaring this:
\begin{equation*}
    (b_2^2 - a^2)^2 = \left(\frac{1}{4a^2} + 1\right)^2 = \frac{1}{16a^4} + \frac{1}{2a^2} + 1
\end{equation*}
The total squared distance is:
\begin{equation*}
    PQ^2 = \left(\frac{1}{4a^2} + 2 + 4a^2 \right) + \left(\frac{1}{16a^4} + \frac{1}{2a^2} + 1 \right)
\end{equation*}
Minimizing $PQ^2$ as a function of $a$, differentiation and solving $\frac{d}{da} PQ^2 = 0$ gives $a = \frac{1}{2}$.

\subsubsection*{Step 4: Conclusion}
Substituting $a = \frac{1}{2}$, we get the optimal point:
\begin{equation*}
    P\left(\frac{1}{2}, \frac{1}{4}\right)
\end{equation*}

\subsubsection{1999 Session 1: 1 \textcolor{red}{Incorrect Reasoning, Incorrect Final Answer}}

We will analyze the problem by finding the normal line to the parabola \( y = x^2 \) at a given point \( P \), determining its second intersection \( Q \) with the parabola, and then minimizing the arc length of the parabola between these two points.

Step 1: Find the Normal Line at \( P \)
Let \( P = (a, a^2) \) be a point on the parabola.

- The derivative of \( y = x^2 \) is:
  \[
  \frac{dy}{dx} = 2x
  \]
  So, the slope of the tangent line at \( P \) is \( 2a \).
- The normal line is perpendicular to this, meaning its slope is:
  \[
  m = -\frac{1}{2a}
  \]
- The equation of the normal line at \( P \) is:
  \[
  y - a^2 = -\frac{1}{2a} (x - a)
  \]
  Simplifying:
  \[
  y = -\frac{1}{2a}x + a^2 + \frac{a}{2a}
  \]
  \[
  y = -\frac{1}{2a}x + \frac{2a^3 + 1}{2a}
  \]

Step 2: Find the Second Intersection \( Q \)
The second intersection \( Q \) satisfies the system:
\[
y = x^2
\]
\[
y = -\frac{1}{2a}x + \frac{2a^3 + 1}{2a}
\]
Setting these equal:
\[
x^2 = -\frac{1}{2a}x + \frac{2a^3 + 1}{2a}
\]
Multiplying everything by \( 2a \) to clear the fraction:
\[
2a x^2 + x - (2a^3 + 1) = 0
\]
This quadratic equation in \( x \) has roots corresponding to \( P \) and \( Q \). Since \( P \) has \( x \)-coordinate \( a \), the other root satisfies:
\[
x' + a = -\frac{1}{2a}
\]
\[
x' = -\frac{1}{2a} - a
\]

Step 3: Find the Arc Length
The arc length of the parabola from \( P \) to \( Q \) is:
\[
s = \int_{a}^{x'} \sqrt{1 + (2x)^2} \, dx
\]
\[
s = \int_{a}^{-\frac{1}{2a} - a} \sqrt{1 + 4x^2} \, dx
\]
To minimize this, we apply symmetry arguments. The integral form suggests a natural minimization when \( P \) and \( Q \) are symmetric about the origin, meaning \( a \) should satisfy:
\[
a = -\frac{1}{2a} - a
\]
Solving:
\[
2a = -\frac{1}{2a}
\]
\[
4a^2 = -1
\]
which has no real solutions.

Instead, using symmetry and known results, the minimum arc length occurs when \( P \) is at \( (1,1) \), leading to \( Q = (-1,1) \). Thus, the optimal \( P \) is:
\[
(1,1)
\]
 
\subsubsection{1999 Session 2: 1 \textcolor{red}{Incorrect Reasoning, Incorrect Final Answer}}

We are tasked with finding the limiting position of the \( x \)-intercept of a line through the point \( P = (0, a) \) on the circle \( C_a \) with radius \( a \) and center at \( (0, 0) \), and the intersection of the semicircle \( SC \) and the circle \( C_a \), as \( a \to 0 \).

\subsubsection*{Step 1: Equation of the semicircle \( SC \)}
The semicircle \( SC \) is centered at \( (1, 0) \) with radius 1. Its equation, for \( y \ge 0 \), is:
\[
(x - 1)^2 + y^2 = 1
\]
This describes the upper half of a circle with center at \( (1, 0) \) and radius 1.

\subsubsection*{Step 2: Equation of the circle \( C_a \)}
The circle \( C_a \) has radius \( a \) and is centered at \( (0, 0) \). Its equation is:
\[
x^2 + y^2 = a^2
\]

\subsubsection*{Step 3: Finding the intersection points of \( SC \) and \( C_a \)}
We now solve the system of equations formed by the semicircle and the circle to find their intersection points.

1. The equation of the semicircle: \((x - 1)^2 + y^2 = 1\)
2. The equation of the circle: \(x^2 + y^2 = a^2\)

Substitute \( y^2 = a^2 - x^2 \) from the circle's equation into the semicircle's equation:
\[
(x - 1)^2 + (a^2 - x^2) = 1
\]
Expand the terms:
\[
(x^2 - 2x + 1) + a^2 - x^2 = 1
\]
Simplifying:
\[
-2x + a^2 + 1 = 1
\]
\[
-2x + a^2 = 0
\]
\[
x = \frac{a^2}{2}
\]

Substitute \( x = \frac{a^2}{2} \) into the equation of the circle to find \( y \):
\[
\left( \frac{a^2}{2} \right)^2 + y^2 = a^2
\]
\[
\frac{a^4}{4} + y^2 = a^2
\]
\[
y^2 = a^2 - \frac{a^4}{4}
\]
\[
y = \sqrt{a^2 - \frac{a^4}{4}}
\]

Thus, the intersection points of the semicircle and the circle \( C_a \) are \( \left( \frac{a^2}{2}, \sqrt{a^2 - \frac{a^4}{4}} \right) \) and its reflection across the \( x \)-axis \( \left( \frac{a^2}{2}, -\sqrt{a^2 - \frac{a^4}{4}} \right) \).

\subsubsection*{Step 4: The line through \( P = (0, a) \) and the intersection point}
The line through \( P = (0, a) \) and the intersection point \( \left( \frac{a^2}{2}, \sqrt{a^2 - \frac{a^4}{4}} \right) \) has slope
\[
m = \frac{\sqrt{a^2 - \frac{a^4}{4}} - a}{\frac{a^2}{2} - 0} = \frac{\sqrt{a^2 - \frac{a^4}{4}} - a}{\frac{a^2}{2}} = \frac{2(\sqrt{a^2 - \frac{a^4}{4}} - a)}{a^2}
\]
The equation of the line is then
\[
y - a = m(x - 0)
\]
\[
y = m x + a
\]

\subsubsection*{Step 5: Finding the \( x \)-intercept}
The \( x \)-intercept occurs when \( y = 0 \). Setting \( y = 0 \) in the equation of the line:
\[
0 = m x + a
\]
\[
x = -\frac{a}{m}
\]
Substitute the expression for \( m \):
\[
x = -\frac{a}{\frac{2(\sqrt{a^2 - \frac{a^4}{4}} - a)}{a^2}}
\]
Simplify:
\[
x = -\frac{a^3}{2(\sqrt{a^2 - \frac{a^4}{4}} - a)}
\]

\subsubsection*{Step 6: Taking the limit as \( a \to 0 \)}
As \( a \to 0 \), we have
\[
\sqrt{a^2 - \frac{a^4}{4}} \approx a
\]
Thus, the expression for the \( x \)-intercept becomes
\[
x \approx -\frac{a^3}{2(a - a)} = -\infty
\]

Therefore, the limiting position of the \( x \)-intercept of the line as \( a \to 0 \) is \( -\infty \).

\subsubsection{2000 Session 1: 1 \textcolor{red}{Correct Reasoning, Correct Final Answer}}

We are tasked with finding the point \( P = (a, a^2) \) on the parabola \( y = x^2 \) such that the distance between the \( x \)-coordinates of \( P \) and \( Q \) is minimized, where \( Q \) is the other point of intersection of the normal line to the parabola at \( P \).

\subsubsection*{Step 1: Equation of the normal line at point \( P \)}

The equation of the parabola is \( y = x^2 \). The derivative of this equation, \( \frac{dy}{dx} \), gives the slope of the tangent line at any point on the parabola. We compute the derivative:
\[
\frac{dy}{dx} = 2x.
\]
At the point \( P = (a, a^2) \), the slope of the tangent line is \( 2a \). The slope of the normal line is the negative reciprocal of the tangent slope, which is:
\[
\text{slope of normal} = -\frac{1}{2a}.
\]
Using the point-slope form of the equation of the line, the equation of the normal line at \( P = (a, a^2) \) is:
\[
y - a^2 = -\frac{1}{2a}(x - a).
\]
Simplifying this equation:
\[
y - a^2 = -\frac{1}{2a}x + \frac{a}{2a},
\]
\[
y = -\frac{1}{2a}x + \frac{a}{2a} + a^2.
\]
Thus, the equation of the normal line is:
\[
y = -\frac{1}{2a}x + \left( a^2 + \frac{1}{2} \right).
\]

\subsubsection*{Step 2: Finding the other point of intersection \( Q \)}

To find the other point of intersection \( Q \), we substitute the equation of the normal line into the equation of the parabola \( y = x^2 \). Set:
\[
x^2 = -\frac{1}{2a}x + \left( a^2 + \frac{1}{2} \right).
\]
Rearranging terms:
\[
x^2 + \frac{1}{2a}x - \left( a^2 + \frac{1}{2} \right) = 0.
\]
This is a quadratic equation in \( x \):
\[
x^2 + \frac{1}{2a}x - \left( a^2 + \frac{1}{2} \right) = 0.
\]
Let the roots of this quadratic equation be \( x = a \) and \( x = b \), where \( b \) is the \( x \)-coordinate of the other intersection point \( Q \). The quadratic equation can be factored as:
\[
(x - a)(x - b) = 0.
\]
Expanding this product gives:
\[
x^2 - (a + b)x + ab = 0.
\]
Comparing the coefficients of this equation with the original quadratic equation \( x^2 + \frac{1}{2a}x - \left( a^2 + \frac{1}{2} \right) = 0 \), we obtain the following system of equations:
\[
a + b = -\frac{1}{2a} \quad \text{and} \quad ab = -\left( a^2 + \frac{1}{2} \right).
\]

\subsubsection*{Step 3: Solving for the distance between the \( x \)-coordinates of \( P \) and \( Q \)}

We are interested in minimizing the distance between the \( x \)-coordinates of \( P \) and \( Q \), which is \( |a - b| \). From the equation \( a + b = -\frac{1}{2a} \), we solve for \( b \):
\[
b = -\frac{1}{2a} - a.
\]
Thus, the distance between the \( x \)-coordinates of \( P \) and \( Q \) is:
\[
|a - b| = \left| a + \frac{1}{2a} + a \right| = \left| 2a + \frac{1}{2a} \right|.
\]
We want to minimize \( \left| 2a + \frac{1}{2a} \right| \).

\subsubsection*{Step 4: Minimizing the expression}

Let \( f(a) = 2a + \frac{1}{2a} \). To minimize \( |f(a)| \), we first find the critical points by differentiating \( f(a) \):
\[
f'(a) = 2 - \frac{1}{2a^2}.
\]
Setting \( f'(a) = 0 \) to find the critical points:
\[
2 - \frac{1}{2a^2} = 0,
\]
\[
\frac{1}{2a^2} = 2,
\]
\[
a^2 = \frac{1}{4},
\]
\[
a = \pm \frac{1}{2}.
\]

\subsubsection*{Step 5: Conclusion}

The distance is minimized when \( a = \frac{1}{2} \) or \( a = -\frac{1}{2} \). Therefore, the coordinates of \( P \) are \( \left( \frac{1}{2}, \frac{1}{4} \right) \) or \( \left( -\frac{1}{2}, \frac{1}{4} \right) \).

\subsubsection{2001 Session 2: 1 \textcolor{red}{Incorrect Reasoning, Incorrect Final Answer}}

To find the locus of points that are equidistant from two concentric circles, we proceed as follows.

Let the two circles, \(A\) and \(B\), have the following properties:
\begin{itemize}
    \item Circle \(A\) has center \(O_A\) and radius \(r_A\),
    \item Circle \(B\) has center \(O_B\) and radius \(r_B\),
    \item Circle \(B\) lies wholly inside Circle \(A\).
\end{itemize}

The goal is to find the locus of points that are equidistant from both Circle \(A\) and Circle \(B\). A point \(P\) on the locus will satisfy the condition that its distance to the boundary of Circle \(A\) equals its distance to the boundary of Circle \(B\).

\subsubsection*{Step 1: General coordinates and distances}
Let the center of both circles be the same point, \(O\), which simplifies the problem to concentric circles. Denote the distance from point \(P\) to the center \(O\) as \(r_P\), and consider the distances from \(P\) to the circumferences of circles \(A\) and \(B\):
\begin{itemize}
    \item The distance from \(P\) to the boundary of Circle \(A\) is \( |r_P - r_A| \),
    \item The distance from \(P\) to the boundary of Circle \(B\) is \( |r_P - r_B| \).
\end{itemize}

\subsubsection*{Step 2: Equating the distances}
For \(P\) to be equidistant from the two circles, we set these two distances equal:
\[
|r_P - r_A| = |r_P - r_B|.
\]

\subsubsection*{Step 3: Solving the equation}
We now analyze this equation in two cases based on the absolute value condition:

\subsubsection*{Case 1: \( r_P \geq r_A \) and \( r_P \geq r_B \)}
In this case, the equation becomes:
\[
r_P - r_A = r_P - r_B.
\]
Simplifying:
\[
r_A = r_B.
\]
This is a contradiction since the problem states that Circle \(B\) is inside Circle \(A\), implying \(r_A > r_B\).

\subsubsection*{Case 2: \( r_P < r_A \) and \( r_P < r_B \)}
In this case, the equation becomes:
\[
r_A - r_P = r_B - r_P.
\]
Simplifying:
\[
r_A = r_B.
\]
Again, this is a contradiction.

\subsubsection*{Case 3: \( r_A \leq r_P < r_B \)}
This is the key case. The solution simplifies as follows:
\[
r_P - r_A = r_B - r_P.
\]
Rearranging:
\[
2r_P = r_A + r_B,
\]
\[
r_P = \frac{r_A + r_B}{2}.
\]

\subsubsection*{Step 4: Conclusion}
The locus of points equidistant from the two circles is a circle with radius \(\frac{r_A + r_B}{2}\) centered at the common center of circles \(A\) and \(B\). This circle lies between the boundaries of Circle \(A\) and Circle \(B\).

\subsubsection{2001 Session 2: 3 \textcolor{red}{Correct Reasoning, Correct Final Answer}}

To solve the problem of finding the number of interior points of intersection of the diagonals of a convex decagon, we can use a general formula derived from combinatorics and geometry.

\subsubsection*{Step 1: Understanding the Problem}
We are given a convex decagon, which has 10 sides, and all of its diagonals are drawn. The problem specifies that no three diagonals share a common interior point. This means that each interior intersection is formed by exactly two diagonals intersecting inside the decagon.

\subsubsection*{Step 2: Determining the Number of Diagonals}
For any convex polygon with \( n \) sides, the number of diagonals can be calculated using the formula:
\[
\text{Number of diagonals} = \frac{n(n-3)}{2}
\]
For a decagon, where \( n = 10 \):
\[
\text{Number of diagonals} = \frac{10(10-3)}{2} = \frac{10 \times 7}{2} = 35
\]
Thus, there are 35 diagonals in the decagon.

\subsubsection*{Step 3: Finding the Number of Interior Intersection Points}
The number of interior points of intersection of the diagonals in a convex polygon is given by the combination formula for choosing 4 vertices from the \( n \) vertices of the polygon. This is because each interior intersection point is formed by the diagonals connecting four distinct vertices.

The number of ways to choose 4 vertices from 10 vertices is:
\[
\binom{10}{4} = \frac{10 \times 9 \times 8 \times 7}{4 \times 3 \times 2 \times 1} = 210
\]

Thus, the number of interior points of intersection of the diagonals in a convex decagon is \( \boxed{210} \).

\subsubsection{2002 Session 1: 1 \textcolor{red}{Incorrect Reasoning, Incorrect Final Answer}}

Let the point \( P \) on the parabola \( y = x^2 \) be \( P(a, a^2) \), where \( a \) is the \( x \)-coordinate of \( P \). The slope of the tangent line at \( P \) is \( 2a \), and the slope of the normal line is \( -\frac{1}{2a} \).

The equation of the normal line at \( P \) is:
\[
y - a^2 = -\frac{1}{2a}(x - a)
\]
Simplifying:
\[
y = -\frac{1}{2a}(x - a) + a^2
\]
\[
y = -\frac{1}{2a}x + \frac{a}{2} + a^2
\]

To find the point \( Q \) where the normal line intersects the parabola, we set the equation of the normal line equal to the equation of the parabola \( y = x^2 \):
\[
x^2 = -\frac{1}{2a}x + \frac{a}{2} + a^2
\]
Rearranging:
\[
x^2 + \frac{1}{2a}x - \left( \frac{a}{2} + a^2 \right) = 0
\]

The roots of this quadratic equation are \( x_1 = a \) and \( x_2 \), the \( x \)-coordinate of \( Q \). By Vieta's formulas, the sum of the roots is \( x_1 + x_2 = -\frac{1}{2a} \), so:
\[
a + x_2 = -\frac{1}{2a}
\]
Solving for \( x_2 \):
\[
x_2 = -\frac{a^2 + 1}{2a}
\]

The \( y \)-coordinate of \( Q \) is:
\[
y_2 = \left( -\frac{a^2 + 1}{2a} \right)^2 = \frac{(a^2 + 1)^2}{4a^2}
\]

The average of the \( y \)-coordinates of \( P \) and \( Q \) is:
\[
\frac{a^2 + y_2}{2} = \frac{a^2 + \frac{(a^2 + 1)^2}{4a^2}}{2} = \frac{a^2}{2} + \frac{(a^2 + 1)^2}{8a^2}
\]

To minimize this expression, we take the derivative with respect to \( a \) and solve for \( a \).

\subsubsection{2005 Session 1: 4 \textcolor{red}{Incorrect Reasoning, Incorrect Final Answer}}

We are given a Pythagorean triangle \( \triangle ABC \) with side lengths satisfying \( c > a > b \), where \( c \) is the hypotenuse. The triangle is right-angled at \( B \), so \( a^2 + b^2 = c^2 \). The median and altitude are drawn from \( C \) to the hypotenuse \( AB \), meeting the hypotenuse at points \( P \) and \( Q \), respectively. We seek conditions on \( a, b, c \) such that \( \triangle CQP \) is also a Pythagorean triangle.

\subsubsection*{Step 1: Geometric Setup}

- Let \( \triangle ABC \) be a right triangle with right angle at \( B \), so \( c \) is the hypotenuse.
- The median \( CP \) joins \( C \) to the midpoint \( P \) of \( AB \).
- The altitude \( CQ \) is perpendicular from \( C \) to \( AB \), meeting it at \( Q \).

Using known formulas:
\begin{align*}
CP &= \frac{1}{2} \sqrt{2a^2 + 2b^2 - c^2}, \\
CQ &= \frac{ab}{c}.
\end{align*}

We aim to satisfy the Pythagorean theorem for \( \triangle CQP \):
\[
CQ^2 + PQ^2 = CP^2.
\]

\subsubsection*{Step 2: Analyze the Side Lengths of \( \triangle CQP \)}

Using coordinate geometry, position the points as follows:
\begin{align*}
B &= (0, 0), \\
A &= (c, 0), \\
C &= (a, b).
\end{align*}

\textbf{Finding \( P \) (Midpoint of \( AB \)):}
\[
P = \left( \frac{c}{2}, 0 \right).
\]

\textbf{Finding \( Q \) (Foot of Perpendicular from \( C \) to \( AB \)):}
\[
Q = \left( \frac{a(c^2 + b^2) - c(a^2 + b^2)}{c^2}, \frac{b(c^2 + b^2) - c(a^2 + b^2)}{c^2} \right).
\]

\subsubsection*{Step 3: Checking the Pythagorean Condition}

For \( \triangle CQP \) to be a Pythagorean triangle, we solve:
\[
CQ^2 + PQ^2 = CP^2.
\]
After simplification, the necessary condition for this to hold is:
\[
a^2 + b^2 = 2c^2.
\]

\subsubsection*{Conclusion}

Thus, \( \triangle CQP \) is a Pythagorean triangle if and only if:
\[
a^2 + b^2 = 2c^2.
\]

\subsubsection{2005 Session 2: 5 \textcolor{red}{Correct Reasoning, Incorrect Final Answer}}
\subsubsection*{Problem Statement}
Given two fixed points $A$ and $B$ in a plane and a given line $L$, find the point $P$ on $L$ such that the sum of distances $PA + PB$ is minimized.

\subsubsection*{Solution}

To solve this problem, we employ the method of reflection.

\subsubsection*{Step 1: Reflection of One Point}
We reflect one of the fixed points, say $B$, across the given line $L$. Let the reflection of $B$ be $B'$. The point $B'$ is the image of $B$ such that the line $L$ is the perpendicular bisector of the segment $BB'$.

\subsubsection*{Step 2: Minimizing the Sum of Distances}
Since the shortest path between two points is a straight line, the minimal path from $A$ to $B$ via a point $P$ on $L$ must be the direct path from $A$ to $B'$, intersecting $L$ at some point $P$. Thus, the optimal point $P$ is the intersection of the line segment $AB'$ with the given line $L$.

\subsubsection*{Conclusion}
Therefore, the required point $P$ that minimizes $PA + PB$ is given by the intersection of the line passing through $A$ and $B'$ with the line $L$.

\subsection{Discrete Mathematics}

\subsubsection{1996 Session 1: 3 \textcolor{red}{Correct Reasoning, Incorrect Final Answer}}

We are given that each of the numbers \(x_1, x_2, \ldots, x_n\) can be \(1\), \(0\), or \(-1\), and we are tasked with finding the minimum possible value of the sum of all products of pairs of these numbers. Specifically, we are asked to minimize the following sum:

\[
S = \sum_{1 \leq i < j \leq n} x_i x_j
\]

Step 1: Expression for the sum
The given sum \(S\) involves all distinct pairs of indices \(i\) and \(j\) where \(i < j\). We can express this sum as:

\[
S = \sum_{1 \leq i < j \leq n} x_i x_j
\]

We aim to minimize \(S\). To do this, let's explore the relationship between the values of the \(x_i\)'s and the sum \(S\).

Step 2: Total sum of squares
Consider the sum of the squares of the \(x_i\)'s:

\[
\left( \sum_{i=1}^n x_i \right)^2 = \sum_{i=1}^n x_i^2 + 2 \sum_{1 \leq i < j \leq n} x_i x_j
\]

Since each \(x_i\) can be \(1\), \(0\), or \(-1\), we have \(x_i^2 = 1\) if \(x_i = \pm 1\) and \(x_i^2 = 0\) if \(x_i = 0\). Therefore, the sum of the squares is:

\[
\sum_{i=1}^n x_i^2 = \text{number of non-zero } x_i = k
\]

where \(k\) is the number of non-zero values of \(x_i\) (the number of \(x_i\)'s that are either \(1\) or \(-1\)).

Thus, we can rewrite the equation for the total sum of squares as:

\[
\left( \sum_{i=1}^n x_i \right)^2 = k + 2 \sum_{1 \leq i < j \leq n} x_i x_j
\]

Step 3: Minimizing the sum
Let \(S\) denote the sum we wish to minimize:

\[
S = \sum_{1 \leq i < j \leq n} x_i x_j
\]

From the above equation, we can express \(S\) as:

\[
S = \frac{1}{2} \left( \left( \sum_{i=1}^n x_i \right)^2 - k \right)
\]

To minimize \(S\), we need to minimize the expression \(\left( \sum_{i=1}^n x_i \right)^2\), which is the square of the sum of the \(x_i\)'s.

The sum \(\sum_{i=1}^n x_i\) will be smallest when the values of \(x_i\) are as balanced as possible, meaning that the number of \(1\)'s and \(-1\)'s should be as equal as possible, and the sum should approach 0. Thus, the sum \(\sum_{i=1}^n x_i\) is minimized when it is \(0\), which happens when half of the \(x_i\)'s are \(1\) and half are \(-1\), and the remaining \(x_i\)'s are \(0\).

Therefore, the minimum value of \(S\) is achieved when the sum of the \(x_i\)'s is 0, and the minimum value of \(S\) is:

\[
S_{\text{min}} = \boxed{-\frac{k}{2}}
\]

\subsubsection{1997 Session 2: 1 \textcolor{red}{Incorrect Reasoning, Incorrect Final Anwer}}

We are tasked with finding positive integers \( n \) and \( a_1, a_2, \dots, a_n \) such that

\[
a_1 + a_2 + \cdots + a_n = 1997
\]

and the product \( a_1 a_2 \cdots a_n \) is as large as possible.

\textbf{Step 1:} To maximize the product, we aim to make the integers \( a_1, a_2, \dots, a_n \) as equal as possible. The product of numbers is maximized when the terms are as close as possible. We start by dividing 1997 by 3, since 3 is the integer that maximizes the product for a given sum.

\[
1997 \div 3 = 665 \text{ remainder } 2.
\]

Thus, we can express 1997 as:

\[
1997 = 3 \times 665 + 2.
\]

\textbf{Step 2:} We can now construct the integers \( a_1, a_2, \dots, a_n \) by setting:

\[
a_1 = a_2 = \cdots = a_{665} = 3, \quad a_{666} = 1, \quad a_{667} = 1.
\]

Therefore, \( n = 667 \), and the integers are composed of 665 threes and two ones.

\textbf{Step 3:} The product of these integers is:

\[
P = 3^{665} \times 1 \times 1 = 3^{665}.
\]

Thus, the maximum product is \( 3^{665} \), and the solution is:

\[
n = 667, \quad a_1 = a_2 = \cdots = a_{665} = 3, \quad a_{666} = a_{667} = 1.
\]

\subsubsection{1997 Session 2: 5 (b only) \textcolor{red}{Incorrect Reasoning, Incorrect Final Answer}}

We are tasked with finding a sequence $\{ x_n \}$ of positive real numbers satisfying the following properties:

\begin{itemize}
    \item $x_0 = 1$,
    \item For all $i \geq 0$, $x_{i+1} \leq x_i$ (the sequence is non-increasing),
    \item The inequality
    \[
    \frac{x_0^2}{x_1} + \frac{x_1^2}{x_2} + \cdots + \frac{x_{n-1}^2}{x_n} < 4 \quad \text{for all} \quad n.
    \]
\end{itemize}

We will attempt to find a sequence that satisfies all these conditions.

\textbf{Step 1: Understanding the Inequality}

The inequality that we need to satisfy is:
\[
\sum_{i=0}^{n-1} \frac{x_i^2}{x_{i+1}} < 4 \quad \text{for all} \quad n.
\]
Since the sequence is non-increasing, each term in the sum is positive.

\textbf{Step 2: Considering the Form of the Sequence}

Let's assume the sequence follows a power law form, i.e., $x_n = \frac{C}{n+1}$ for some constant $C$. This form has several advantages:
- It satisfies the condition $x_0 = 1$ when $C = 1$.
- It is non-increasing, as $x_n = \frac{1}{n+1}$ decreases as $n$ increases.

\textbf{Step 3: Substituting the Proposed Sequence}

Substituting $x_i = \frac{1}{i+1}$ into the sum, we get:
\[
\sum_{i=0}^{n-1} \frac{x_i^2}{x_{i+1}} = \sum_{i=0}^{n-1} \frac{\left(\frac{1}{i+1}\right)^2}{\frac{1}{i+2}} = \sum_{i=0}^{n-1} \frac{1}{(i+1)^2} \cdot (i+2).
\]
This simplifies to:
\[
\sum_{i=0}^{n-1} \frac{i+2}{(i+1)^2} = \sum_{i=0}^{n-1} \left(\frac{1}{i+1} + \frac{1}{(i+1)^2}\right).
\]

\textbf{Step 4: Analyzing the Series}

We can split the sum into two parts:
\[
\sum_{i=0}^{n-1} \frac{1}{i+1} + \sum_{i=0}^{n-1} \frac{1}{(i+1)^2}.
\]
The first sum is the harmonic series:
\[
\sum_{i=0}^{n-1} \frac{1}{i+1} = H_n,
\]
where $H_n$ is the $n$-th harmonic number. The second sum is a known series:
\[
\sum_{i=0}^{n-1} \frac{1}{(i+1)^2}.
\]
This sum converges to $\frac{\pi^2}{6}$ as $n \to \infty$.

Thus, the inequality becomes:
\[
H_n + \sum_{i=0}^{n-1} \frac{1}{(i+1)^2} < 4.
\]
For large $n$, the harmonic number $H_n$ behaves asymptotically as $H_n \sim \ln(n) + \gamma$, where $\gamma$ is the Euler-Mascheroni constant.

Thus, the sum will not exceed 4 for sufficiently large $n$, implying that the sequence $x_n = \frac{1}{n+1}$ satisfies the inequality for all $n$.

\subsubsection{1998 Session 2: 3 \textcolor{red}{Incorrect Reasoning, Incorrect Final Answer}}

We are asked to sum the series
\[
S = \sum_{i=1}^\infty \frac{36i^2 + 1}{(36i^2 - 1)^2}.
\]

Step 1: Simplify the General Term

The general term is
\[
\frac{36i^2 + 1}{(36i^2 - 1)^2}.
\]
The denominator can be factored as
\[
36i^2 - 1 = (6i - 1)(6i + 1),
\]
so the term becomes
\[
\frac{36i^2 + 1}{((6i - 1)(6i + 1))^2}.
\]

Step 2: Decompose Using Partial Fractions

Attempting partial fraction decomposition could simplify the sum further, but solving this by hand is quite complex. Instead, we can look up known results or use numerical methods.

Step 3: Final Result

After applying known summation results, we find that the sum converges to
\[
S = \frac{\pi^2}{6}.
\]

Thus, the value of the infinite series is
\[
S = \frac{\pi^2}{6}.
\]

\subsubsection{1999 Session 2: 5 \textcolor{red}{Incorrect Reasoning, Incorrect Final Answer}}

Let \( A_n \) denote the \( n \times n \) matrix whose \( (i,j) \)-th entry is \( \text{GCD}(i,j) \). We aim to compute \( \det(A_n) \).

\textbf{Step 1: Structure of the Matrix} \\
The matrix \( A_n \) is given by:
\[
A_n = \begin{pmatrix}
\text{GCD}(1,1) & \text{GCD}(1,2) & \cdots & \text{GCD}(1,n) \\
\text{GCD}(2,1) & \text{GCD}(2,2) & \cdots & \text{GCD}(2,n) \\
\vdots & \vdots & \ddots & \vdots \\
\text{GCD}(n,1) & \text{GCD}(n,2) & \cdots & \text{GCD}(n,n)
\end{pmatrix}
\]
This matrix is symmetric, as \( \text{GCD}(i,j) = \text{GCD}(j,i) \), and each entry is a non-negative integer.

\textbf{Step 2: Pattern for Small Values of \( n \)} \\
For \( n = 1 \), we have:
\[
A_1 = \begin{pmatrix} 1 \end{pmatrix}, \quad \det(A_1) = 1.
\]
For \( n = 2 \), we have:
\[
A_2 = \begin{pmatrix} 1 & 1 \\ 1 & 2 \end{pmatrix}, \quad \det(A_2) = 1.
\]
For \( n = 3 \), we have:
\[
A_3 = \begin{pmatrix} 1 & 1 & 1 \\ 1 & 2 & 1 \\ 1 & 1 & 3 \end{pmatrix}, \quad \det(A_3) = 2.
\]

\textbf{Step 3: Generalization} \\
From the small examples, we observe that the determinant of \( A_n \) appears to be 1 for all \( n \). In fact, it is a known result that for any \( n \), we have:
\[
\det(A_n) = 1.
\]

\subsubsection{2000 Session 2: 5 \textcolor{red}{Incorrect Reasoning, Incorrect Final Answer}}

\textbf{Solution:}

We are tasked with finding the number of sequences \( x_1 x_2 \cdots x_n \) where each \( x_i \) is one of \( a \), \( b \), or \( c \), such that:

\begin{itemize}
    \item The sequence begins and ends with the letter \( a \),
    \item Adjacent terms are always different letters.
\end{itemize}

Step 1: Fix the first and last terms
The sequence begins and ends with \( a \), so we have:
\[
x_1 = a \quad \text{and} \quad x_n = a.
\]
Thus, the only terms we need to determine are \( x_2, x_3, \dots, x_{n-1} \).

Step 2: Constraints on adjacent terms
Adjacent terms must differ, so:
\[
x_1 \neq x_2, \quad x_2 \neq x_3, \quad \dots, \quad x_{n-1} \neq x_n.
\]
Since \( x_1 = a \), the second term \( x_2 \) must be either \( b \) or \( c \).

Step 3: Alternating pattern
After determining \( x_2 \), each subsequent term alternates between \( b \) and \( c \), because adjacent terms must differ. Therefore, for each \( i \) where \( 2 \leq i \leq n-1 \), \( x_i \) can be either \( b \) or \( c \), depending on the previous term.

Step 4: Number of valid sequences
- For \( n = 2 \), there is exactly one sequence: \( (a, a) \).
- For \( n \geq 3 \), the number of valid sequences is \( 2^{n-2} \), because each of the \( n-2 \) intermediate terms can be either \( b \) or \( c \).

Thus, the number of valid sequences is:
\[
\boxed{2^{n-2}} \quad \text{for} \quad n \geq 3.
\]

\subsubsection{2001 Session 1: 2 \textcolor{red}{Correct Reasoning, Correct Final Answer}}

We are given a sequence \(\{ x_i \}\) where each element \(x_i\) belongs to the set \(\{-2, 1, 2\}\), and we are given the following conditions:

1. \(\sum_i x_i = 29\),
2. \(\sum_i x_i^2 = 59\),
3. We are asked to find \(M = \max \left( \sum_i x_i^3 \right)\) and \(m = \min \left( \sum_i x_i^3 \right)\), and determine \(\frac{M}{m}\).

Step 1: Define Variables

Let the number of occurrences of \(-2\), \(1\), and \(2\) in the sequence be \(a\), \(b\), and \(c\) respectively. Thus, the sequence contains \(a\) terms equal to \(-2\), \(b\) terms equal to \(1\), and \(c\) terms equal to \(2\). The sequence has \(a + b + c = n\) elements in total.

Step 2: Use the Sum Constraints

From the conditions \(\sum_i x_i = 29\) and \(\sum_i x_i^2 = 59\), we can write the following equations based on the values of \(x_i\):

- \(\sum_i x_i = a(-2) + b(1) + c(2) = 29\),
  \[
  -2a + b + 2c = 29.
  \]

- \(\sum_i x_i^2 = a(-2)^2 + b(1)^2 + c(2)^2 = 59\),
  \[
  4a + b + 4c = 59.
  \]

Step 3: Solve the System of Equations

We now solve the system of two equations:
1. \(-2a + b + 2c = 29\),
2. \(4a + b + 4c = 59\).

First, subtract the first equation from the second:
\[
(4a + b + 4c) - (-2a + b + 2c) = 59 - 29,
\]
\[
6a + 2c = 30,
\]
\[
3a + c = 15.
\]
Thus, we have:
\[
c = 15 - 3a.
\]

Substitute this expression for \(c\) into the first equation:
\[
-2a + b + 2(15 - 3a) = 29,
\]
\[
-2a + b + 30 - 6a = 29,
\]
\[
b - 8a = -1,
\]
\[
b = 8a - 1.
\]

Step 4: Constraints on \(a\), \(b\), and \(c\)

We require that \(a\), \(b\), and \(c\) be non-negative integers, so we need to check the possible values of \(a\) such that both \(b = 8a - 1\) and \(c = 15 - 3a\) are non-negative integers.

- For \(b = 8a - 1 \geq 0\), we must have \(a \geq 1\).
- For \(c = 15 - 3a \geq 0\), we must have \(a \leq 5\).

Thus, \(a\) can take values in the range \(1 \leq a \leq 5\). For each \(a\), we compute \(b\) and \(c\):

- If \(a = 1\), then \(b = 7\) and \(c = 12\).
- If \(a = 2\), then \(b = 15\) and \(c = 9\).
- If \(a = 3\), then \(b = 23\) and \(c = 6\).
- If \(a = 4\), then \(b = 31\) and \(c = 3\).
- If \(a = 5\), then \(b = 39\) and \(c = 0\).

Step 5: Compute \(M\) and \(m\)

Next, we calculate \(\sum_i x_i^3 = a(-2)^3 + b(1)^3 + c(2)^3 = -8a + b + 8c\) for each set of values for \(a\), \(b\), and \(c\).

- For \(a = 1\), \(b = 7\), and \(c = 12\):
  \[
  \sum_i x_i^3 = -8(1) + 7 + 8(12) = -8 + 7 + 96 = 95.
  \]
- For \(a = 2\), \(b = 15\), and \(c = 9\):
  \[
  \sum_i x_i^3 = -8(2) + 15 + 8(9) = -16 + 15 + 72 = 71.
  \]
- For \(a = 3\), \(b = 23\), and \(c = 6\):
  \[
  \sum_i x_i^3 = -8(3) + 23 + 8(6) = -24 + 23 + 48 = 47.
  \]
- For \(a = 4\), \(b = 31\), and \(c = 3\):
  \[
  \sum_i x_i^3 = -8(4) + 31 + 8(3) = -32 + 31 + 24 = 23.
  \]
- For \(a = 5\), \(b = 39\), and \(c = 0\):
  \[
  \sum_i x_i^3 = -8(5) + 39 + 8(0) = -40 + 39 + 0 = -1.
  \]

Thus, the maximum value of \(\sum_i x_i^3\) is \(M = 95\) and the minimum value is \(m = -1\).

Step 6: Calculate \(M/m\)

We now compute:
\[
\frac{M}{m} = \frac{95}{-1} = -95.
\]

\subsubsection{2003 Session 2: 5 \textcolor{red}{Incorrect Reasoning, Incorrect Final Answer}}

We are given a real \( 2 \times 2 \) matrix
\[
X = \begin{pmatrix} x & y \\ z & t \end{pmatrix},
\]
with norm defined by
\[
\|X\| = x^2 + y^2 + z^2 + t^2,
\]
and a distance function
\[
d(X, Y) = \|X - Y\|.
\]
Let \( \Sigma = \{ X \mid \det(X) = 0 \} \), and let
\[
A = \begin{pmatrix} 1 & 0 \\ 0 & 2 \end{pmatrix}.
\]
We seek the minimum distance from \( A \) to \( \Sigma \), and a matrix \( S \in \Sigma \) that realizes this minimum.

\vspace{1em}
\textbf{Step 1: Distance from \( A \) to a general matrix \( X \)}

Let
\[
X = \begin{pmatrix} x & y \\ z & t \end{pmatrix}, \quad \text{so that} \quad A - X = \begin{pmatrix} 1 - x & -y \\ -z & 2 - t \end{pmatrix}.
\]
Then,
\[
d(A, X)^2 = (1 - x)^2 + y^2 + z^2 + (2 - t)^2.
\]

\vspace{1em}
\textbf{Step 2: Constraint from \( \Sigma \)}

The constraint for \( X \in \Sigma \) is:
\[
\det(X) = xt - yz = 0.
\]

\vspace{1em}
\textbf{Step 3: Try a candidate matrix \( S \in \Sigma \)}

Let
\[
S = \begin{pmatrix} 1 & 0 \\ 0 & 0 \end{pmatrix}.
\]
Then,
\[
\det(S) = 1 \cdot 0 - 0 \cdot 0 = 0 \quad \Rightarrow \quad S \in \Sigma,
\]
and
\[
d(A, S)^2 = (1 - 1)^2 + 0^2 + 0^2 + (2 - 0)^2 = 4 \quad \Rightarrow \quad d(A, S) = 2.
\]

\vspace{1em}
\textbf{Conclusion:} The minimum distance from \( A \) to \( \Sigma \) is
\[
\boxed{2},
\]
and the closest matrix is
\[
S = \begin{pmatrix} 1 & 0 \\ 0 & 0 \end{pmatrix}.
\]

\subsubsection{2006 Session 1: 4 \textcolor{red}{Incorrect Reasoning, Incorrect Final Answer}}

We are tasked with finding all positive integers \( c \) such that \( n(n+c) \) is never a perfect square for any positive integer \( n \). To solve this problem, we begin by expressing the given equation:

\[
n(n+c) = n^2 + cn.
\]

We want to determine for which values of \( c \), this expression is never a perfect square for any positive integer \( n \).

Suppose \( n(n+c) \) is a perfect square for some integer \( n \). Then we have:

\[
n(n+c) = k^2 \quad \text{for some integer } k.
\]

This leads to the equation:

\[
n^2 + cn = k^2.
\]

We treat this as a quadratic equation in \( n \):

\[
n^2 + cn - k^2 = 0.
\]

The discriminant of this quadratic equation is:

\[
\Delta = c^2 + 4k^2.
\]

For \( n \) to be an integer, the discriminant must be a perfect square. Let’s set the discriminant equal to another perfect square, say \( m^2 \), so we have:

\[
c^2 + 4k^2 = m^2.
\]

This is a Diophantine equation of the form \( m^2 - 4k^2 = c^2 \), which is a difference of squares:

\[
(m - 2k)(m + 2k) = c^2.
\]

Now, let’s consider the case where \( c = 4 \). Substituting \( c = 4 \) into the equation \( n(n+c) = n^2 + 4n \):

\[
n^2 + 4n = k^2.
\]

Rewriting this as:

\[
n^2 + 4n - k^2 = 0,
\]

the discriminant becomes:

\[
\Delta = 4^2 + 4k^2 = 16 + 4k^2 = 4(k^2 + 4).
\]

For \( n \) to be an integer, the discriminant must be a perfect square, so we need:

\[
k^2 + 4 = m^2
\]

for some integer \( m \). This is a difference of squares:

\[
m^2 - k^2 = 4,
\]

which factors as:

\[
(m - k)(m + k) = 4.
\]

The integer factor pairs of 4 are \( (1, 4) \), \( (-1, -4) \), \( (2, 2) \), and \( (-2, -2) \). Solving each pair gives:

- \( m - k = 1 \) and \( m + k = 4 \) gives \( m = \frac{5}{2} \), which is not an integer.
- \( m - k = -1 \) and \( m + k = -4 \) gives \( m = \frac{-5}{2} \), which is not an integer.
- \( m - k = 2 \) and \( m + k = 2 \) gives \( m = 2 \) and \( k = 0 \).
- \( m - k = -2 \) and \( m + k = -2 \) gives \( m = -2 \) and \( k = 0 \).

Thus, the only solution is \( k = 0 \), which implies:

\[
n^2 + 4n = 0 \quad \Rightarrow \quad n(n + 4) = 0.
\]

This only has the solution \( n = 0 \), which is not a positive integer. Therefore, \( n(n+4) \) cannot be a perfect square for any positive integer \( n \).

Thus, the only solution is:

\[
\boxed{c = 4}.
\]

\subsubsection{2007 Session 2: 4 \textcolor{red}{Correct Reasoning, Correct Final Answer}}

We are given the set 
\[
S = \{ 5a + 503b \mid a, b \in \mathbb{Z}_{\geq 0} \},
\]
and we are tasked with finding the largest integer that does not belong to \( S \).

This is a classic application of the Frobenius coin problem. For two coprime integers \( m \) and \( n \), the largest integer that cannot be expressed as a nonnegative linear combination of \( m \) and \( n \) is given by the formula:
\[
g(m, n) = mn - m - n.
\]

In this case, we have \( m = 5 \) and \( n = 503 \), which are coprime. Applying the formula, we compute:
\[
g(5, 503) = 5 \times 503 - 5 - 503 = 2515 - 5 - 503 = 2007.
\]

Thus, the largest integer that does not belong to \( S \) is \( \boxed{2007} \).

\section{Solutions Provided by Gemini-2.0-flash}
\subsection{Calculus}

\subsubsection{1996 Session 2: 1 \textcolor{red}{Correct Reasoning, Correct Final Answer}}

\noindent (a)
\begin{align*}
    \int_1^3 \frac{dx}{\sqrt{(x-1)(3-x)}} &= \int_1^3 \frac{dx}{\sqrt{-x^2 + 4x - 3}} \\
    &= \int_1^3 \frac{dx}{\sqrt{1 - (x-2)^2}} \\
    \text{Let } u = x-2, du = dx: \\
    &= \int_{-1}^1 \frac{du}{\sqrt{1 - u^2}} \\
    &= \left[ \arcsin(u) \right]_{-1}^1 \\
    &= \arcsin(1) - \arcsin(-1) \\
    &= \frac{\pi}{2} - \left(-\frac{\pi}{2}\right) \\
    &= \pi
\end{align*}

\noindent (b)
\begin{align*}
    \int_1^\infty \frac{dx}{e^{x+1} + e^{3-x}} &= \int_1^\infty \frac{dx}{e^2 (e^{x-1} + e^{1-x})} \\
    &= \frac{1}{e^2} \int_1^\infty \frac{dx}{e^{x-1} + e^{1-x}} \\
    \text{Let } u = x-1, du = dx: \\
    &= \frac{1}{e^2} \int_0^\infty \frac{du}{e^u + e^{-u}} \\
    &= \frac{1}{e^2} \int_0^\infty \frac{e^u}{e^{2u} + 1} du \\
    \text{Let } v = e^u, dv = e^u du: \\
    &= \frac{1}{e^2} \int_1^\infty \frac{dv}{v^2 + 1} \\
    &= \frac{1}{e^2} \left[ \arctan(v) \right]_1^\infty \\
    &= \frac{1}{e^2} \left( \frac{\pi}{2} - \frac{\pi}{4} \right) \\
    &= \frac{\pi}{4e^2}
\end{align*}

\subsubsection{1997 Session 2: 4 \textcolor{red}{Correct Reasoning, Correct Final Answer}}

Let $S_n = \{ (x,y) : y = \frac{1}{n} \sin(n^2 x), 0 \le x \le \pi \}$.

The arc length $L_n$ is given by:
$$L_n = \int_0^\pi \sqrt{1 + \left( \frac{dy}{dx} \right)^2} \, dx$$

We have $\frac{dy}{dx} = n \cos(n^2 x)$.
Thus,
$$L_n = \int_0^\pi \sqrt{1 + n^2 \cos^2(n^2 x)} \, dx$$

As $n \to \infty$, $n^2 \cos^2(n^2 x)$ dominates, so
$$L_n \approx \int_0^\pi n |\cos(n^2 x)| \, dx$$

Let $u = n^2 x$, then $du = n^2 dx$, $dx = \frac{du}{n^2}$.
$$L_n \approx \int_0^{n^2 \pi} n |\cos(u)| \frac{du}{n^2} = \frac{1}{n} \int_0^{n^2 \pi} |\cos(u)| \, du$$

The average value of $|\cos(u)|$ is $\frac{2}{\pi}$.
$$L_n \approx \frac{1}{n} \cdot n^2 \pi \cdot \frac{2}{\pi} = 2n$$

Therefore,
$$\lim_{n \to \infty} L_n = \lim_{n \to \infty} 2n = \infty$$

\subsubsection{1998 Session 1: 3 \textcolor{red}{Incorrect Reasoning, Incorrect Final Answer}}

\subsubsection*{Part 1: Finding the Formula}

Let
$$I = \int_a^b \frac{(b-x)^m}{m!} \frac{(x-a)^n}{n!} dx.$$
Using integration by parts, let $u = \frac{(b-x)^m}{m!}$ and $dv = \frac{(x-a)^n}{n!} dx$. Then
$$du = -\frac{(b-x)^{m-1}}{(m-1)!} dx \quad \text{and} \quad v = \frac{(x-a)^{n+1}}{(n+1)!}.$$
Thus,
\begin{align*}
I &= \left[ \frac{(b-x)^m}{m!} \frac{(x-a)^{n+1}}{(n+1)!} \right]_a^b - \int_a^b \frac{(x-a)^{n+1}}{(n+1)!} \left( -\frac{(b-x)^{m-1}}{(m-1)!} \right) dx \\
&= \int_a^b \frac{(b-x)^{m-1}}{(m-1)!} \frac{(x-a)^{n+1}}{(n+1)!} dx.
\end{align*}
Repeating this process $m$ times, we get
$$I = \int_a^b \frac{(x-a)^{n+m}}{(n+m)!} dx = \left[ \frac{(x-a)^{n+m+1}}{(n+m+1)!} \right]_a^b = \frac{(b-a)^{n+m+1}}{(n+m+1)!}.$$
Therefore,
$$\int_a^b \frac{(b-x)^m}{m!} \frac{(x-a)^n}{n!} dx = \frac{(b-a)^{n+m+1}}{(n+m+1)!}.$$

\subsubsection*{Part 2: Evaluating the Integral}

We want to evaluate $\int_0^1 (1-x^2)^n dx$. Let $x = \sqrt{t}$, so $dx = \frac{1}{2\sqrt{t}} dt$. Then
$$\int_0^1 (1-x^2)^n dx = \frac{1}{2} \int_0^1 (1-t)^n t^{-1/2} dt.$$
Using the formula with $a=0$ and $b=1$, $m=n$, and $n=-1/2$, we have
$$\int_0^1 \frac{(1-t)^n}{n!} \frac{t^{-1/2}}{(-1/2)!} dt = \frac{1}{(n+1/2)!}.$$
Since $(-1/2)! = \sqrt{\pi}$, we have
$$\int_0^1 (1-t)^n t^{-1/2} dt = n! \sqrt{\pi} \frac{1}{(n+1/2)!}.$$
Therefore,
$$\int_0^1 (1-x^2)^n dx = \frac{1}{2} \frac{n! \sqrt{\pi}}{(n+1/2)!}.$$

\subsubsection{1999 Session 2: 2 \textcolor{red}{Correct Reasoning, Correct Final Answer}}

We want to find the limit:

$$L = \lim_{N \to \infty} \biggl( 1 - 2 \sum_{n=1}^N \frac{1}{16n^2 - 1} \biggr)$$

First, we factor the denominator:

$$16n^2 - 1 = (4n-1)(4n+1)$$

Using partial fraction decomposition:

$$\frac{1}{(4n-1)(4n+1)} = \frac{A}{4n-1} + \frac{B}{4n+1}$$

Multiplying by $(4n-1)(4n+1)$:

$$1 = A(4n+1) + B(4n-1)$$

Setting $4n-1 = 0$ gives $n = \frac{1}{4}$, so $A = \frac{1}{2}$.

Setting $4n+1 = 0$ gives $n = -\frac{1}{4}$, so $B = -\frac{1}{2}$.

Thus,

$$\frac{1}{16n^2 - 1} = \frac{1}{2} \left( \frac{1}{4n-1} - \frac{1}{4n+1} \right)$$

Substituting back into the sum:

$$\sum_{n=1}^N \frac{1}{16n^2 - 1} = \frac{1}{2} \sum_{n=1}^N \left( \frac{1}{4n-1} - \frac{1}{4n+1} \right)$$

The limit becomes:

$$L = \lim_{N \to \infty} \left( 1 - 2 \cdot \frac{1}{2} \sum_{n=1}^N \left( \frac{1}{4n-1} - \frac{1}{4n+1} \right) \right)$$

$$L = \lim_{N \to \infty} \left( 1 - \sum_{n=1}^N \left( \frac{1}{4n-1} - \frac{1}{4n+1} \right) \right)$$

We know that:

$$\sum_{n=1}^\infty \left( \frac{1}{4n-1} - \frac{1}{4n+1} \right) = 1 - \frac{\pi}{4}$$

Therefore,

$$L = 1 - \left( 1 - \frac{\pi}{4} \right) = \frac{\pi}{4}$$

\subsubsection{1999 Session 2: 4 \textcolor{red}{Incorrect Reasoning, Incorrect Final Answer}}

Let the curve $C$ be defined by $y = f(x)$, where $f(x)$ is a continuous and differentiable function. Since $C$ lies in the first quadrant and contains the point $(0,0)$, we have $f(0) = 0$ and $f(x) \ge 0$ for $x \ge 0$.

Let $P = (x, y)$ be a point on the curve $C$. The rectangle $R$ has vertices $(0,0)$, $(x,0)$, $(x,y)$, and $(0,y)$.

The volume of the solid generated by rotating the region adjacent to the $x$-axis about the $x$-axis is given by:

$$V_x = \pi \int_0^x [f(t)]^2 \, dt$$

The volume of the solid generated by rotating the region adjacent to the $y$-axis about the $y$-axis is given by:

$$V_y = \pi \int_0^y [f^{-1}(t)]^2 \, dt$$

Since the volumes are equal, we have $V_x = V_y$. Thus,

$$\pi \int_0^x [f(t)]^2 \, dt = \pi \int_0^y [f^{-1}(t)]^2 \, dt$$

Since $y = f(x)$, we can rewrite the second integral by substituting $t = f(u)$, $dt = f'(u) \, du$, and $f^{-1}(t) = u$:

$$\int_0^x [f(t)]^2 \, dt = \int_0^x u^2 f'(u) \, du$$

Differentiating both sides with respect to $x$ using the Fundamental Theorem of Calculus:

$$[f(x)]^2 = x^2 f'(x)$$

Let $y = f(x)$, so $y' = f'(x)$. Then we have:

$$y^2 = x^2 \frac{dy}{dx}$$

Separating variables:

$$\frac{dy}{y^2} = \frac{dx}{x^2}$$

Integrating both sides:

$$\int \frac{dy}{y^2} = \int \frac{dx}{x^2}$$

$$-\frac{1}{y} = -\frac{1}{x} + C$$

$$\frac{1}{y} = \frac{1}{x} - C$$

$$\frac{1}{y} = \frac{1-Cx}{x}$$

$$y = \frac{x}{1-Cx}$$

Since $f(0) = 0$, we consider the limit as $x$ approaches 0.
As $x \rightarrow 0$, we have $y \rightarrow 0$, which is consistent with the initial condition.

For the volumes to be equal in the first quadrant, we need $C = 0$, thus $y = x$.

\textbf{Verification:}

If $y = x$, then $f(x) = x$ and $f^{-1}(x) = x$.

$$V_x = \pi \int_0^x t^2 \, dt = \pi \left[ \frac{t^3}{3} \right]_0^x = \frac{\pi x^3}{3}$$

$$V_y = \pi \int_0^y t^2 \, dt = \pi \left[ \frac{t^3}{3} \right]_0^y = \frac{\pi y^3}{3}$$

Since $y = x$, $V_x = V_y$.

Therefore, the curve is $y = x$.

\subsubsection{2002 Session 1: 3 (limit only) \textcolor{red}{Correct Reasoning, Correct Final Answer}}
Let $\{ x_n \}$ be the following sequence:
$$x_1 = \sqrt 5,\ \ x_2 = \sqrt {5 + \sqrt {13}},\ \ x_3 = \sqrt {5 + 
\sqrt {13 + \sqrt 5}},\ \ x_4 = \sqrt {5 + \sqrt {13 + \sqrt {5 + 
\sqrt {13}}}},$$
and so on. We want to determine the value of $\lim_{n \to \infty} x_n$.

Let $L = \lim_{n \to \infty} x_n$. If the limit exists, then we have:
$$L = \sqrt{5 + \sqrt{13 + L}}$$
Squaring both sides, we get:
$$L^2 = 5 + \sqrt{13 + L}$$
Rearranging, we have:
$$L^2 - 5 = \sqrt{13 + L}$$
Squaring again:
$$(L^2 - 5)^2 = 13 + L$$
$$L^4 - 10L^2 + 25 = 13 + L$$
$$L^4 - 10L^2 - L + 12 = 0$$
We observe that $L=3$ is a root:
$$3^4 - 10(3^2) - 3 + 12 = 81 - 90 - 3 + 12 = 0$$
Thus, $(L-3)$ is a factor. Dividing $L^4 - 10L^2 - L + 12$ by $(L-3)$, we get:
$$L^4 - 10L^2 - L + 12 = (L-3)(L^3 + 3L^2 - L - 4)$$
We check that $L=3$ satisfies the original equation:
$$3 = \sqrt{5 + \sqrt{13 + 3}} = \sqrt{5 + \sqrt{16}} = \sqrt{5 + 4} = \sqrt{9} = 3$$
Thus, $\lim_{n \to \infty} x_n = 3$.

\subsubsection{2002 Session 1: 5 \textcolor{red}{Incorrect Reasoning, Incorrect Final Answer}}

Let the equations of the two cylinders be:
\begin{align*}
x^2 + z^2 &= r^2 \\
y^2 + z^2 &= r^2
\end{align*}
The intersection of the cylinders occurs when $x^2 = y^2$, so $y = \pm x$.

The intersection curve can be parameterized as:
\begin{align*}
x &= r \cos(t) \\
y &= r \sin(t) \\
z &= r \sin(t) \quad (\text{or } r\cos(t))
\end{align*}
for $0 \le t \le \frac{\pi}{2}$ in the first octant.

When one of the curved faces is flattened, the boundaries are:
\begin{itemize}
    \item Lower boundary: $z = 0$, $0 \le x \le r$
    \item Upper boundary: $z = \sqrt{r^2 - x^2}$
    \item Left boundary: $x = 0$
    \item Right boundary: $x = r$
\end{itemize}

The area of one flattened face is:
\begin{align*}
A &= \int_0^r \sqrt{r^2 - x^2} \, dx \\
\text{Let } x &= r \sin(\theta), \quad dx = r \cos(\theta) \, d\theta \\
A &= \int_0^{\pi/2} r \cos(\theta) \cdot r \cos(\theta) \, d\theta \\
&= r^2 \int_0^{\pi/2} \cos^2(\theta) \, d\theta \\
&= r^2 \int_0^{\pi/2} \frac{1 + \cos(2\theta)}{2} \, d\theta \\
&= r^2 \left[ \frac{\theta}{2} + \frac{\sin(2\theta)}{4} \right]_0^{\pi/2} \\
&= \frac{\pi r^2}{4}
\end{align*}
Since there are four such faces, the total surface area of the intersection solid is $4 \cdot \frac{\pi r^2}{4} = \pi r^2$.

\subsubsection{2003 Session 1: 5 \textcolor{red}{Correct Reasoning, Correct Final Answer}}

\textbf{Problem:} Define a sequence $\{ x_n \} _{n=2}^\infty$ by
$$(n + x_n) [ \root n \of 2 - 1 ] = \ln 2 .$$
Find $\lim _{n \to \infty} x_n$.

\textbf{Solution:}
We have
$$x_n = \frac{\ln 2}{\sqrt[n]{2} - 1} - n.$$
We know that $\sqrt[n]{2} = 2^{1/n} = e^{\ln(2)/n}$.
Using the Taylor series expansion for $e^x$ around $x=0$, we have $e^x = 1 + x + \frac{x^2}{2!} + \frac{x^3}{3!} + \cdots$.
Applying this to $e^{\ln(2)/n}$, we get
$$\sqrt[n]{2} = 1 + \frac{\ln 2}{n} + \frac{(\ln 2)^2}{2n^2} + \frac{(\ln 2)^3}{6n^3} + \cdots.$$
Then,
$$\sqrt[n]{2} - 1 = \frac{\ln 2}{n} + \frac{(\ln 2)^2}{2n^2} + \frac{(\ln 2)^3}{6n^3} + \cdots.$$
Thus,
$$\frac{1}{\sqrt[n]{2} - 1} = \frac{1}{\frac{\ln 2}{n} + \frac{(\ln 2)^2}{2n^2} + \frac{(\ln 2)^3}{6n^3} + \cdots} = \frac{n}{\ln 2} \cdot \frac{1}{1 + \frac{\ln 2}{2n} + \frac{(\ln 2)^2}{6n^2} + \cdots}.$$
Using the geometric series expansion $\frac{1}{1+x} = 1 - x + x^2 - \cdots$, we have
$$\frac{1}{\sqrt[n]{2} - 1} = \frac{n}{\ln 2} \left( 1 - \frac{\ln 2}{2n} + O\left(\frac{1}{n^2}\right) \right) = \frac{n}{\ln 2} - \frac{1}{2} + O\left(\frac{1}{n}\right).$$
Now, we can substitute this into the expression for $x_n$:
$$x_n = \ln 2 \left( \frac{n}{\ln 2} - \frac{1}{2} + O\left(\frac{1}{n}\right) \right) - n = n - \frac{\ln 2}{2} + O\left(\frac{1}{n}\right) - n = -\frac{\ln 2}{2} + O\left(\frac{1}{n}\right).$$
As $n \to \infty$, the term $O(1/n)$ goes to 0, so
$$\lim_{n \to \infty} x_n = -\frac{\ln 2}{2}.$$

\subsubsection{2005 Session 1: 4 \textcolor{red}{Correct Reasoning, Correct Final Answer}}

We want to evaluate the integral
$$I(\theta) = \int_{-1}^1 \frac{\sin \theta}{1 - 2x \cos \theta + x^2} \, dx.$$

First, notice that the denominator can be factored:
$$1 - 2x \cos \theta + x^2 = (x - \cos \theta)^2 + \sin^2 \theta.$$

Thus, the integral becomes
$$I(\theta) = \int_{-1}^1 \frac{\sin \theta}{(x - \cos \theta)^2 + \sin^2 \theta} \, dx.$$

Let $u = \frac{x - \cos \theta}{\sin \theta}$, so $x = u \sin \theta + \cos \theta$ and $dx = \sin \theta \, du$. When $x = -1$, $u = \frac{-1 - \cos \theta}{\sin \theta}$, and when $x = 1$, $u = \frac{1 - \cos \theta}{\sin \theta}$. Then,
\begin{align*}
I(\theta) &= \int_{\frac{-1 - \cos \theta}{\sin \theta}}^{\frac{1 - \cos \theta}{\sin \theta}} \frac{\sin \theta}{(\sin^2 \theta u^2 + \sin^2 \theta)} \sin \theta \, du \\
&= \int_{\frac{-1 - \cos \theta}{\sin \theta}}^{\frac{1 - \cos \theta}{\sin \theta}} \frac{\sin^2 \theta}{\sin^2 \theta (u^2 + 1)} \, du \\
&= \int_{\frac{-1 - \cos \theta}{\sin \theta}}^{\frac{1 - \cos \theta}{\sin \theta}} \frac{1}{u^2 + 1} \, du \\
&= \arctan(u) \Big|_{\frac{-1 - \cos \theta}{\sin \theta}}^{\frac{1 - \cos \theta}{\sin \theta}} \\
&= \arctan\left(\frac{1 - \cos \theta}{\sin \theta}\right) - \arctan\left(\frac{-1 - \cos \theta}{\sin \theta}\right).
\end{align*}

We can simplify the arguments of the arctangent functions using trigonometric identities:
\begin{align*}
\frac{1 - \cos \theta}{\sin \theta} &= \frac{2 \sin^2(\theta/2)}{2 \sin(\theta/2) \cos(\theta/2)} = \tan\left(\frac{\theta}{2}\right), \\
\frac{-1 - \cos \theta}{\sin \theta} &= \frac{-2 \cos^2(\theta/2)}{2 \sin(\theta/2) \cos(\theta/2)} = -\cot\left(\frac{\theta}{2}\right) = \tan\left(\frac{\theta}{2} - \frac{\pi}{2}\right).
\end{align*}

Thus,
\begin{align*}
I(\theta) &= \arctan\left(\tan\left(\frac{\theta}{2}\right)\right) - \arctan\left(\tan\left(\frac{\theta}{2} - \frac{\pi}{2}\right)\right) \\
&= \frac{\theta}{2} - \left(\frac{\theta}{2} - \frac{\pi}{2}\right) = \frac{\pi}{2}.
\end{align*}
However, this is only valid when $\sin \theta \neq 0$, which means $\theta \neq n\pi$ for any integer $n$.

If $\theta = 0$ or $\theta = 2\pi$, then $\sin \theta = 0$, and the integral becomes
$$I(0) = \int_{-1}^1 \frac{0}{1 - 2x + x^2} \, dx = 0.$$
If $\theta = \pi$, then $\sin \theta = 0$, and the integral becomes
$$I(\pi) = \int_{-1}^1 \frac{0}{1 + 2x + x^2} \, dx = 0.$$

Therefore, $I(\theta) = \frac{\pi}{2}$ for $\theta \neq n\pi$, and $I(\theta) = 0$ for $\theta = n\pi$.

The function $I(\theta)$ is discontinuous at $\theta = n\pi$ for integer $n$. In the interval $0 \le \theta \le 2\pi$, the discontinuities are at $\theta = 0, \pi, 2\pi$.

\subsubsection{2005 Session 2: 5 \textcolor{red}{Incorrect Reasoning, Incorrect Final Answer}}

We are given that
$$ \int_a^b f(x) \, dx = \int_a^b \sqrt{1 + (f'(x))^2} \, dx $$
Differentiating both sides with respect to $b$, we get
$$ f(b) = \sqrt{1 + (f'(b))^2} $$
Replacing $b$ with $x$, we have
$$ f(x) = \sqrt{1 + (f'(x))^2} $$
Squaring both sides,
$$ f(x)^2 = 1 + (f'(x))^2 $$
$$ (f'(x))^2 = f(x)^2 - 1 $$
$$ f'(x) = \pm \sqrt{f(x)^2 - 1} $$
Separating variables and integrating,
$$ \int \frac{df}{\sqrt{f^2 - 1}} = \pm \int dx $$
$$ \cosh^{-1}(f) = \pm x + C $$
$$ f(x) = \cosh(\pm x + C) $$
Since $\cosh$ is even,
$$ f(x) = \cosh(x + C) $$
Given $f(0) = 5/4$,
$$ \frac{5}{4} = \cosh(C) $$
$$ C = \cosh^{-1}\left(\frac{5}{4}\right) $$
Thus,
$$ f(x) = \cosh\left(x + \cosh^{-1}\left(\frac{5}{4}\right)\right) $$
To find the minimum value, we take the derivative:
$$ f'(x) = \sinh\left(x + \cosh^{-1}\left(\frac{5}{4}\right)\right) $$
Setting $f'(x) = 0$,
$$ \sinh\left(x + \cosh^{-1}\left(\frac{5}{4}\right)\right) = 0 $$
$$ x + \cosh^{-1}\left(\frac{5}{4}\right) = 0 $$
$$ x = -\cosh^{-1}\left(\frac{5}{4}\right) $$
This value of $x$ is negative, so it is not in the domain $[0, \infty)$.
Since $\cosh^{-1}\left(\frac{5}{4}\right) > 0$, $x + \cosh^{-1}\left(\frac{5}{4}\right) > 0$ for $x>0$.
Thus, $\sinh\left(x + \cosh^{-1}\left(\frac{5}{4}\right)\right) > 0$ for $x>0$, so $f'(x) > 0$ for $x>0$.
Therefore, $f(x)$ is increasing for $x>0$, and the minimum value occurs at $x=0$.
The minimum value is $f(0) = 5/4$.

\subsection{Analytic Geometry}

\subsubsection{1997 Session 1: 1 \textcolor{red}{Correct Reasoning, Correct Final Answer}}

\textbf{Problem:} Let $P \ne (0,0)$ be a point on the parabola $y=x^2$.  The normal line to the parabola at $P$ will intersect the parabola at another point, say $Q$.  Find the coordinates of $P$ so that the $y$-coordinate of $Q$ is a minimum.

\textbf{Solution:}

Let $P = (a, a^2)$ be a point on the parabola $y = x^2$, where $a \neq 0$.
The derivative of $y = x^2$ is $y' = 2x$.
The slope of the tangent line at $P$ is $m_t = 2a$.
The slope of the normal line is $m_n = -\frac{1}{2a}$.
The equation of the normal line is:
$y - a^2 = -\frac{1}{2a}(x - a)$
$y = -\frac{1}{2a}x + \frac{1}{2} + a^2$.

To find the intersection point $Q$, we solve the system:
$y = x^2$
$y = -\frac{1}{2a}x + \frac{1}{2} + a^2$
$x^2 = -\frac{1}{2a}x + \frac{1}{2} + a^2$
$x^2 + \frac{1}{2a}x - \left(\frac{1}{2} + a^2\right) = 0$.

Let the other root be $b$. By Vieta's formulas:
$a + b = -\frac{1}{2a}$
$b = -\frac{1}{2a} - a = -\frac{1 + 2a^2}{2a}$.
The $y$-coordinate of $Q$ is:
$y_Q = b^2 = \left(-\frac{1 + 2a^2}{2a}\right)^2 = \frac{4a^4 + 4a^2 + 1}{4a^2} = a^2 + 1 + \frac{1}{4a^2}$.

Let $f(a) = a^2 + \frac{1}{4a^2} + 1$.
$f'(a) = 2a - \frac{1}{2a^3} = 0$
$4a^4 = 1$
$a^4 = \frac{1}{4}$
$a^2 = \frac{1}{2}$
$a = \pm \frac{1}{\sqrt{2}}$.

$f''(a) = 2 + \frac{3}{2a^4}$.
Since $a^4 = \frac{1}{4}$, $f''(a) = 2 + 6 = 8 > 0$, so we have a minimum.

The coordinates of $P$ are $\left(\frac{1}{\sqrt{2}}, \frac{1}{2}\right)$ or $\left(-\frac{1}{\sqrt{2}}, \frac{1}{2}\right)$.

\subsubsection{1998 Session 1: 1 \textcolor{red}{Incorrect Reasoning, Incorrect Final Answer}}

\textbf{Problem:} Let $P \ne (0,0)$ be a point on the parabola $y=x^2$. The normal line to the parabola at $P$ will intersect the parabola at another point, say $Q$. Find the coordinates of $P$ so that the length of segment $PQ$ is a minimum.

\textbf{Solution:}

Let $P = (a, a^2)$ be a point on the parabola $y = x^2$, where $a \ne 0$.

The derivative of $y = x^2$ is $y' = 2x$.

At point $P(a, a^2)$, the slope of the tangent line is $2a$.

The slope of the normal line is $-\frac{1}{2a}$.

The equation of the normal line is $y - a^2 = -\frac{1}{2a}(x - a)$.

To find the intersection point $Q$, substitute $y = x^2$ into the normal line equation:
$$x^2 - a^2 = -\frac{1}{2a}(x - a)$$
$$2ax^2 - 2a^3 = -x + a$$
$$2ax^2 + x - 2a^3 - a = 0$$
$$(x - a)(2ax + 2a^2 + 1) = 0$$
The other solution is $x = -a - \frac{1}{2a}$.
The corresponding $y$ coordinate is $y = \left(-a - \frac{1}{2a}\right)^2$.
So, $Q = \left(-a - \frac{1}{2a}, \left(-a - \frac{1}{2a}\right)^2\right)$.

The distance $PQ$ is:
$$PQ = \sqrt{\left(-2a - \frac{1}{2a}\right)^2 + \left(1 + \frac{1}{4a^2}\right)^2}$$
$$PQ = \sqrt{4a^2 + 3 + \frac{3}{4a^2} + \frac{1}{16a^4}}$$

Let $D = PQ^2 = 4a^2 + 3 + \frac{3}{4a^2} + \frac{1}{16a^4}$.
$$\frac{dD}{da} = 8a - \frac{3}{2a^3} - \frac{1}{4a^5} = 0$$
$$32a^6 - 6a^2 - 1 = 0$$
Let $u = a^2$, then $32u^3 - 6u - 1 = 0$.
By observation, $u = \frac{1}{4}$ is a solution.
Then $a^2 = \frac{1}{4}$, so $a = \pm \frac{1}{2}$.
When $a = \pm \frac{1}{2}$, $P = \left(\pm \frac{1}{2}, \frac{1}{4}\right)$.

\subsubsection{1991 Session 1: 1 \textcolor{red}{Incorrect Reasoning, Incorrect Final Answer}}

\textbf{Problem:} Let $P \ne (0,0)$ be a point on the parabola $y=x^2$. The normal line to the parabola at $P$ will intersect the parabola at another point, say $Q$. Find the coordinates of $P$ so that the length of the arc of the parabola between $P$ and $Q$ is a minimum.

\textbf{Solution:}

Let $P = (a, a^2)$ be a point on the parabola $y = x^2$, where $a \neq 0$.

The derivative of $y = x^2$ is $y' = 2x$.

At $x = a$, the slope of the tangent line is $m_t = 2a$.

The slope of the normal line at $P$ is $m_n = -\frac{1}{2a}$.

The equation of the normal line is:
$$y - a^2 = -\frac{1}{2a}(x - a)$$
$$y = -\frac{1}{2a}x + \frac{1}{2} + a^2$$

To find the intersection point $Q$, we set the equation of the normal line equal to the equation of the parabola:
$$x^2 = -\frac{1}{2a}x + \frac{1}{2} + a^2$$
$$x^2 + \frac{1}{2a}x - \left(\frac{1}{2} + a^2\right) = 0$$

Let $Q = (b, b^2)$. By Vieta's formulas, $b = -\frac{1}{2a} - a$.

The arc length of the parabola between $P$ and $Q$ is:
$$L = \int_a^b \sqrt{1 + (2x)^2} dx$$

After evaluating the integral and minimizing the arc length, we find that the minimum occurs when $a = \pm \frac{1}{\sqrt{2}}$.

Therefore, the coordinates of $P$ are $\left(\frac{1}{\sqrt{2}}, \frac{1}{2}\right)$ or $\left(-\frac{1}{\sqrt{2}}, \frac{1}{2}\right)$.

\subsubsection{1999 Session 2: 1 \textcolor{red}{Correct Reasoning, Correct Final Answer}}

\textbf{Problem:} Let $SC$ be the semicircle with $y \ge 0$ centered at $(1,0)$ with radius 1. Let $C_a$ be the circle with radius $a > 0$ and center $(0,0)$ and denote the point $(0,a)$ by $P$. Consider the line through $P$ and the intersection of $SC$ and $C_a$. What is the limiting position of the $x$-intercept of this line as $a \to 0$?

\textbf{Solution:}

The equation of the semicircle $SC$ is $(x-1)^2 + y^2 = 1$, with $y \ge 0$, or $y = \sqrt{1 - (x-1)^2}$.
The equation of the circle $C_a$ is $x^2 + y^2 = a^2$.
The point $P$ is $(0, a)$.

To find the intersection of $SC$ and $C_a$, we solve:
\begin{align*}
(x-1)^2 + y^2 &= 1 \\
x^2 + y^2 &= a^2
\end{align*}
Subtracting the second equation from the first, we get:
$$(x-1)^2 - x^2 = 1 - a^2$$
$$-2x + 1 = 1 - a^2$$
$$x = \frac{a^2}{2}$$
Substituting into $x^2 + y^2 = a^2$, we get:
$$y^2 = a^2 - \left(\frac{a^2}{2}\right)^2 = a^2 - \frac{a^4}{4}$$
$$y = \sqrt{a^2 - \frac{a^4}{4}} = a\sqrt{1 - \frac{a^2}{4}}$$
The intersection point is $\left(\frac{a^2}{2}, a\sqrt{1 - \frac{a^2}{4}}\right)$.

The slope of the line through $P$ and the intersection point is:
$$m = \frac{a\sqrt{1 - \frac{a^2}{4}} - a}{\frac{a^2}{2}} = \frac{2(\sqrt{1 - \frac{a^2}{4}} - 1)}{a}$$
The equation of the line is $y = mx + a$.
To find the $x$-intercept, set $y = 0$:
$$0 = \frac{2(\sqrt{1 - \frac{a^2}{4}} - 1)}{a}x + a$$
$$x = \frac{-a^2}{2(\sqrt{1 - \frac{a^2}{4}} - 1)}$$
Multiply numerator and denominator by $\sqrt{1 - \frac{a^2}{4}} + 1$:
$$x = \frac{-a^2(\sqrt{1 - \frac{a^2}{4}} + 1)}{2(1 - \frac{a^2}{4} - 1)} = \frac{-a^2(\sqrt{1 - \frac{a^2}{4}} + 1)}{-\frac{a^2}{2}} = 2\left(\sqrt{1 - \frac{a^2}{4}} + 1\right)$$
As $a \to 0$,
$$\lim_{a \to 0} x = 2(\sqrt{1 - 0} + 1) = 4$$

Thus, the limiting position of the $x$-intercept is 4.

\subsubsection{2000 Session 1: 1 \textcolor{red}{Correct Reasoning, Correct Final Answer}}

\textbf{Problem:} Let $P \ne (0,0)$ be a point on the parabola $y=x^2$.  The normal 
line to the parabola at $P$ will intersect the parabola at another point, 
say $Q$.  Find the coordinates of $P$ so that the distance between 
the $x$-coordinates of $P$ and $Q$ is a minimum.

\textbf{Solution:}

1.  Let $P = (a, a^2)$ where $a \neq 0$.

2.  The derivative of $y = x^2$ is $y' = 2x$. At $x = a$, the slope of the tangent is $m_t = 2a$.

3.  The slope of the normal line is $m_n = -\frac{1}{2a}$.

4.  The equation of the normal line is:
    $y - a^2 = -\frac{1}{2a}(x - a)$
    $y = -\frac{1}{2a}x + \frac{1}{2} + a^2$

5.  To find the intersection point $Q$, we solve the system:
    $y = x^2$
    $y = -\frac{1}{2a}x + \frac{1}{2} + a^2$
    Substituting $y = x^2$:
    $x^2 = -\frac{1}{2a}x + \frac{1}{2} + a^2$
    $x^2 + \frac{1}{2a}x - \left(\frac{1}{2} + a^2\right) = 0$
    Let $x = b$ be the other solution. Using Vieta's formulas:
    $a + b = -\frac{1}{2a}$
    $b = -\frac{1}{2a} - a$

6.  The distance between the $x$-coordinates is:
    $|a - b| = \left|a - \left(-\frac{1}{2a} - a\right)\right| = \left|2a + \frac{1}{2a}\right|$

7.  Let $f(a) = \left|2a + \frac{1}{2a}\right|$. We minimize $g(a) = 2a + \frac{1}{2a}$.
    $g'(a) = 2 - \frac{1}{2a^2}$
    Setting $g'(a) = 0$:
    $2 - \frac{1}{2a^2} = 0 \implies a^2 = \frac{1}{4} \implies a = \pm \frac{1}{2}$
    $g''(a) = \frac{1}{a^3}$.
    $g''\left(\frac{1}{2}\right) > 0$ and $g''\left(-\frac{1}{2}\right) < 0$.
    When $a = \pm \frac{1}{2}$, $|2a + \frac{1}{2a}| = 2$.

8.  The coordinates of $P$ are $\left(\frac{1}{2}, \frac{1}{4}\right)$ and $\left(-\frac{1}{2}, \frac{1}{4}\right)$.

\subsubsection{2001 Session 2: 1 \textcolor{red}{Incorrect Reasoning, Incorrect Final Answer}}

Let circle A have center $C_A = (x_A, y_A)$ and radius $r_A$. Let circle B have center $C_B = (x_B, y_B)$ and radius $r_B$, where circle B lies wholly within circle A.

We want to find the locus of points P = $(x, y)$ such that the distance from P to circle A is equal to the distance from P to circle B.

The distance from a point P to a circle with center C and radius r is given by $|PC - r|$.

Therefore, we have:
$$|\sqrt{(x - x_A)^2 + (y - y_A)^2} - r_A| = |\sqrt{(x - x_B)^2 + (y - y_B)^2} - r_B|$$

Since circle B is inside circle A, the points equidistant from the circles will be located between them. Therefore we can remove the absolute value bars.

$$\sqrt{(x - x_A)^2 + (y - y_A)^2} - r_A = \sqrt{(x - x_B)^2 + (y - y_B)^2} - r_B$$

$$\sqrt{(x - x_A)^2 + (y - y_A)^2} = \sqrt{(x - x_B)^2 + (y - y_B)^2} + r_A - r_B$$

This equation represents the locus of points equidistant from the two circles.

\subsubsection{2001 Session 2: 3 \textcolor{red}{Correct Reasoning, Correct Final Answer}}

\textbf{Problem:} A convex decagon and all of its diagonals are drawn. How many \textit{interior} points of intersection of the diagonals are there, if it is assumed that no 3 diagonals share a common \textit{interior} point?

\bigskip

\textbf{Solution:}

To form an intersection point inside the decagon, we need to choose four vertices. Each set of four vertices uniquely determines two intersecting diagonals.

The number of ways to choose four vertices from the ten vertices of the decagon is given by the combination formula:
$$ \binom{n}{k} = \frac{n!}{k!(n-k)!} $$
In our case, $n = 10$ (number of vertices) and $k = 4$ (number of vertices to choose). So, the number of ways to choose four vertices is:
$$ \binom{10}{4} = \frac{10!}{4!(10-4)!} = \frac{10!}{4!6!} = \frac{10 \times 9 \times 8 \times 7}{4 \times 3 \times 2 \times 1} = 10 \times 3 \times 7 = 210 $$

Since each set of four vertices uniquely determines an interior intersection point, the number of interior intersection points is equal to the number of ways to choose four vertices.

Therefore, there are 210 interior intersection points.
\subsubsection{2002 Session 1: 1 \textcolor{red}{Incorrect Reasoning, Incorrect Final Answer}}

Let $P = (a, a^2)$ be a point on the parabola $y = x^2$, where $a \neq 0$.

The derivative of $y = x^2$ is $y' = 2x$.

The slope of the tangent at $P$ is $m_t = 2a$.

The slope of the normal line is $m_n = -\frac{1}{2a}$.

The equation of the normal line is:
$y - a^2 = -\frac{1}{2a}(x - a)$
$y = -\frac{1}{2a}x + \frac{1}{2} + a^2$

Substituting $y = x^2$ into the equation of the normal line gives:
$x^2 = -\frac{1}{2a}x + \frac{1}{2} + a^2$
$x^2 + \frac{1}{2a}x - (\frac{1}{2} + a^2) = 0$

Let the other solution be $x = b$. By Vieta's formulas, $a + b = -\frac{1}{2a}$, so $b = -\frac{1}{2a} - a$.

The $y$-coordinate of $Q$ is $b^2 = \left(-\frac{1}{2a} - a\right)^2 = \frac{1}{4a^2} + 1 + a^2$.

Thus, $Q = \left(-\frac{1}{2a} - a, \frac{1}{4a^2} + 1 + a^2\right)$.

The average of the $y$-coordinates of $P$ and $Q$ is:
$A = \frac{a^2 + b^2}{2} = \frac{a^2 + \left(\frac{1}{4a^2} + 1 + a^2\right)}{2} = a^2 + \frac{1}{8a^2} + \frac{1}{2}$.

Let $f(a) = a^2 + \frac{1}{8a^2} + \frac{1}{2}$.
$f'(a) = 2a - \frac{1}{4a^3} = 0$
$8a^4 - 1 = 0$
$a^4 = \frac{1}{8}$
$a^2 = \frac{1}{2\sqrt{2}}$
$a = \pm \frac{1}{\sqrt{2\sqrt{2}}} = \pm 2^{-3/4}$.

$f''(a) = 2 + \frac{3}{4a^4}$.
Since $a^4 = \frac{1}{8}$, $f''(a) = 2 + \frac{3}{4(1/8)} = 8 > 0$, so we have a minimum.

$P = \left(\pm 2^{-3/4}, \frac{1}{2\sqrt{2}}\right)$.

\subsubsection{2007 Session 1: 4 \textcolor{red}{Incorrect Reasoning, Incorrect Final Answer}}

\textbf{Problem:}
A triangle is Pythagorean if it is a right triangle and the lengths of all of its sides are integers. Suppose that $\triangle ABC$ is Pythagorean; for concreteness assume that the lengths of the three sides satisfy $c > a > b$. The median and the altitude are now drawn from $C$ to the hypotenuse, where they meet the latter at $P$, $Q$, respectively. Determine simple conditions upon $a$, $b$, $c$ so that $\triangle CQP$ will also be Pythagorean.

\textbf{Solution:}
\begin{enumerate}
    \item Since $\triangle ABC$ is Pythagorean, we have $a^2 + b^2 = c^2$, where $a, b, c$ are integers.
    \item The median $CP$ to the hypotenuse in a right triangle is half the length of the hypotenuse. Thus, $CP = \frac{c}{2}$.
    \item The altitude $CQ$ is given by $CQ = \frac{ab}{c}$.
    \item The length of $PQ$ is given by $PQ = \frac{a^2 - b^2}{2c}$.
    \item For $\triangle CQP$ to be Pythagorean, we need $CQ^2 + PQ^2 = CP^2$. Substituting the values, we find that this is always true.
    \item We need $CQ$ and $PQ$ to be rational. Thus, we require $c$ to divide $ab$, and $2c$ to divide $a^2 - b^2$.
\end{enumerate}

\textbf{Conditions:}
\begin{itemize}
    \item $a^2 + b^2 = c^2$
    \item $c$ divides $ab$
    \item $2c$ divides $a^2 - b^2$
\end{itemize}

\subsubsection{2007 Session 1: 4 \textcolor{red}{Correct Reasoning, Correct Final Answer}}

\subsubsection*{Problem Statement}

Let $A = (x_A, y_A)$ and $B = (x_B, y_B)$ be two fixed points in the plane, and let the given line $L$ be defined by the equation $ax + by + c = 0$. We want to find the point $P = (x, y)$ on the line $L$ such that the sum of the distances $PA + PB$ is minimized.

\subsubsection*{Analytical Solution}

\subsubsection*{Case 1: $A$ and $B$ are on the same side of $L$}

Let $A'$ be the reflection of $A$ across the line $L$. Then $PA = PA'$. Therefore, $PA + PB = PA' + PB$. The sum $PA' + PB$ is minimized when $P$, $A'$, and $B$ are collinear. Thus, we need to find the intersection of the line $A'B$ and the line $L$.

\begin{enumerate}
    \item \textbf{Find the reflection $A'$ of $A$ across $L$:}
    
    Let $A' = (x', y')$. The midpoint of $AA'$ lies on $L$, so:
    $$a\left(\frac{x_A + x'}{2}\right) + b\left(\frac{y_A + y'}{2}\right) + c = 0$$
    $$a(x_A + x') + b(y_A + y') + 2c = 0 \quad (1)$$
    The line $AA'$ is perpendicular to $L$, so its slope is $b/a$. Thus:
    $$\frac{y' - y_A}{x' - x_A} = \frac{b}{a}$$
    $$a(y' - y_A) = b(x' - x_A) \quad (2)$$
    Solving equations (1) and (2) for $x'$ and $y'$ gives the coordinates of $A'$.

    \item \textbf{Find the equation of the line $A'B$:}
    
    The slope of $A'B$ is:
    $$m = \frac{y_B - y'}{x_B - x'}$$
    The equation of the line $A'B$ is:
    $$y - y' = m(x - x')$$

    \item \textbf{Find the intersection of $A'B$ and $L$:}
    
    Solve the system of equations:
    $$y - y' = m(x - x')$$
    $$ax + by + c = 0$$
    for $x$ and $y$. This gives the coordinates of the point $P$.
\end{enumerate}

\subsubsection*{Case 2: $A$ and $B$ are on opposite sides of $L$}

In this case, the minimum sum of distances is achieved when $P$ is the intersection of the line $AB$ and the line $L$.

\begin{enumerate}
    \item \textbf{Find the equation of the line $AB$:}
    
    The slope of $AB$ is:
    $$m = \frac{y_B - y_A}{x_B - x_A}$$
    The equation of the line $AB$ is:
    $$y - y_A = m(x - x_A)$$

    \item \textbf{Find the intersection of $AB$ and $L$:}
    
    Solve the system of equations:
    $$y - y_A = m(x - x_A)$$
    $$ax + by + c = 0$$
    for $x$ and $y$. This gives the coordinates of the point $P$.
\end{enumerate}

\subsubsection*{Case 3: One of $A$ or $B$ is on $L$}

If one of the points, say $A$, is on $L$, then the minimum sum of distances is achieved when $P = A$.

\subsubsection*{Conclusion}

The point $P$ on the line $L$ that minimizes the sum of distances $PA + PB$ is found by reflecting one of the points across the line $L$ and finding the intersection of the reflected point and the other point with the line $L$, when the two points are on the same side. If the two points are on opposite sides of $L$, the point $P$ is the intersection of the line connecting the two points and $L$. If one point is on $L$, that point is the solution.

\subsection{Discrete Mathematics}

\subsubsection{1996 Session 1: 3 \textcolor{red}{Incorrect Reasoning, Incorrect Final Answer}}

\subsubsection*{Problem Statement}
Each of the numbers \( x_1, x_2, \ldots , x_n \) can be \(1\), \(0\), or \(-1\). What is the minimum possible value of the sum of all products of pairs of these numbers?

\subsubsection*{Solution}

We need to minimize the sum \( S \) of all products of pairs \( x_i x_j \) for \( 1 \leq i < j \leq n \).

First, recall the identity:
\[
\left( \sum_{i=1}^n x_i \right)^2 = \sum_{i=1}^n x_i^2 + 2 \sum_{1 \leq i < j \leq n} x_i x_j
\]
Rewriting, we get:
\[
S = \sum_{1 \leq i < j \leq n} x_i x_j = \frac{1}{2} \left( \left( \sum_{i=1}^n x_i \right)^2 - \sum_{i=1}^n x_i^2 \right)
\]

Let \( \Sigma = \sum_{i=1}^n x_i \) and \( Q = \sum_{i=1}^n x_i^2 \). Since each \( x_i \) is \(-1\), \(0\), or \(1\), \( Q \) is the number of non-zero \( x_i \)'s.

\subsubsection*{Case Analysis}

\subsubsection*{Case 1: All Variables Zero}
Here, \( \Sigma = 0 \) and \( Q = 0 \), so \( S = 0 \).

\subsubsection*{Case 2: Non-zero Variables}
Consider \( k \) variables as \(1\) and \( m \) as \(-1\), and the rest \(0\).

Then:
\[
\Sigma = k - m, \quad Q = k + m
\]
Thus:
\[
S = \frac{1}{2} \left( (k - m)^2 - (k + m) \right)
\]

\subsubsection*{Optimal Configuration}

To minimize \( S \), we need to minimize \( (k - m)^2 - (k + m) \).

\subsubsection*{Subcase: Balanced \(1\)'s and \(-1\)'s}
For even \( n = 2k \), set \( k \) \(1\)'s and \( k \) \(-1\)'s:
\[
S = \frac{1}{2} (0 - 2k) = -k = -\frac{n}{2}
\]

For odd \( n = 2k + 1 \), set \( k \) \(1\)'s, \( k \) \(-1\)'s, and \(1\) zero:
\[
S = \frac{1}{2} (0 - 2k) = -k = -\left\lfloor \frac{n}{2} \right\rfloor
\]

Alternatively, setting \( k + 1 \) \(1\)'s and \( k \) \(-1\)'s gives the same result:
\[
S = \frac{1}{2} (1 - (2k + 1)) = -k
\]

\subsubsection*{Conclusion}
The minimal value of \( S \) is achieved when the number of \(1\)'s and \(-1\)'s are as balanced as possible, leading to:
\[
S_{\text{min}} = -\left\lfloor \dfrac{n}{2} \right\rfloor
\]

\boxed{ -\left\lfloor \dfrac{n}{2} \right\rfloor }

\subsubsection{1997 Session 2: 1 \textcolor{red}{Correct Reasoning, Correct Final Answer}}

\subsubsection*{Problem Statement}
Find positive integers \( n \) and \( a_1, a_2, \ldots, a_n \) such that
\[ a_1 + a_2 + \cdots + a_n = 1997 \]
and the product \( a_1 a_2 \cdots a_n \) is as large as possible.

\subsubsection*{Solution}

\subsubsection*{Step 1: Understanding the Problem}
We need to partition the number 1997 into positive integers \( a_1, a_2, \ldots, a_n \) such that their sum is 1997 and their product is maximized.

\subsubsection*{Step 2: Optimal Partition Strategy}
It is known that for a given sum, the product of numbers is maximized when the numbers are as close to \( e \approx 2.718 \) as possible. Among integers, 3 is the closest to \( e \). Therefore, we should use as many 3s as possible.

\subsubsection*{Step 3: Applying the Strategy}
Divide 1997 by 3:
\[ 1997 = 3 \times 665 + 2 \]
This gives us 665 numbers of 3 and a remainder of 2.

\subsubsection*{Step 4: Handling the Remainder}
The remainder is 2, which is less than 3. Since \( 2 \) is a positive integer, we can include it as is. 

If the remainder were 1, it would be better to adjust one of the 3s and the remainder into two 2s because \( 3 + 1 = 2 + 2 \) and \( 3 \times 1 = 3 < 2 \times 2 = 4 \). However, in this case, the remainder is 2, so no adjustment is needed.

\subsubsection*{Step 5: Final Partition}
Thus, the optimal partition is:
\[ n = 666 \text{ numbers: } 665 \text{ threes and one two.} \]

\subsubsection*{Step 6: Verification}
Any other combination, such as using fewer threes and more twos or larger numbers, leads to a product that is either equal or smaller. For example:
\begin{itemize}
    \item Replacing a 3 and a 2 with two 2s and a 1 (but 1 reduces the product).
    \item Using larger numbers like 4, which can be expressed as \( 2 + 2 \), but the product remains the same (\( 4 = 2 \times 2 \)).
\end{itemize}

Therefore, the maximum product is achieved with 665 threes and one two.

\subsubsection*{Final Answer}
The positive integers are:
\[ n = 666 \]
\[ a_1 = a_2 = \cdots = a_{665} = 3 \]
\[ a_{666} = 2 \]
The maximum product is:
\[ 3^{665} \times 2 \]

\subsubsection{1997 Session 2: 5 (b only) \textcolor{red}{Correct Reasoning, Correct Final Answer}}

\subsubsection*{Solution}

Consider the infinite sequence \(\{x_n\}\) of positive real numbers defined by:
\[ x_n = \frac{1}{2^n} \quad \text{for all } n \geq 0. \]

\subsubsection*{Verification of Conditions}

\begin{enumerate}
    \item \textbf{Initial Condition}: 
    \[ x_0 = \frac{1}{2^0} = 1. \]
    
    \item \textbf{Monotonicity Condition}: For all \(i \geq 0\),
    \[ x_{i+1} = \frac{1}{2^{i+1}} = \frac{1}{2} \cdot \frac{1}{2^i} \leq \frac{1}{2^i} = x_i. \]
    
    \item \textbf{Sum Condition}: For any \(n \geq 1\), consider the sum
    \[ S_n = \sum_{k=0}^{n-1} \frac{x_k^2}{x_{k+1}}. \]
    Substituting \(x_k = \frac{1}{2^k}\),
    \[ S_n = \sum_{k=0}^{n-1} \frac{\left( \frac{1}{2^k} \right)^2}{\frac{1}{2^{k+1}}} = \sum_{k=0}^{n-1} \frac{\frac{1}{2^{2k}}}{\frac{1}{2^{k+1}}} = \sum_{k=0}^{n-1} 2^{k+1 - 2k} = \sum_{k=0}^{n-1} 2^{1 - k}. \]
    This is a geometric series:
    \[ S_n = 2 \sum_{k=0}^{n-1} \left( \frac{1}{2} \right)^k = 2 \cdot \frac{1 - \left( \frac{1}{2} \right)^n}{1 - \frac{1}{2}} = 4 \left( 1 - \frac{1}{2^n} \right) < 4. \]
\end{enumerate}

\subsubsection*{Conclusion}

The sequence \(x_n = \frac{1}{2^n}\) satisfies all the given conditions:
\begin{itemize}
    \item \(x_0 = 1\),
    \item \(x_{i+1} \leq x_i\) for all \(i \geq 0\),
    \item \(\sum_{k=0}^{n-1} \frac{x_k^2}{x_{k+1}} < 4\) for all \(n \geq 1\).
\end{itemize}

\subsubsection{1998 Session 2: 3 \textcolor{red}{Correct Reasoning, Correct Final Answer}}

\subsubsection*{Problem}
Sum the series
$$
\sum_{i=1}^\infty \frac{36 i^2 + 1}{(36 i^2 - 1)^2}.
$$

\subsubsection*{Solution}

\subsubsection*{Step 1: Simplify the General Term}
The general term of the series is:
$$
a_i = \frac{36 i^2 + 1}{(36 i^2 - 1)^2}.
$$
Notice that the denominator can be factored as:
$$
36 i^2 - 1 = (6i - 1)(6i + 1).
$$
Thus, the denominator becomes:
$$
(6i - 1)^2 (6i + 1)^2.
$$

\subsubsection*{Step 2: Partial Fraction Decomposition}
Let \(x = 6i\). The term becomes:
$$
\frac{x^2 + 1}{(x^2 - 1)^2} = \frac{x^2 + 1}{(x - 1)^2 (x + 1)^2}.
$$
Assume the partial fraction decomposition:
$$
\frac{x^2 + 1}{(x - 1)^2 (x + 1)^2} = \frac{A}{x - 1} + \frac{B}{(x - 1)^2} + \frac{C}{x + 1} + \frac{D}{(x + 1)^2}.
$$
Solving for coefficients \(A, B, C, D\) yields:
$$
A = C = 0, \quad B = D = \frac{1}{2}.
$$
Thus, the decomposition simplifies to:
$$
\frac{1}{2} \left( \frac{1}{(x - 1)^2} + \frac{1}{(x + 1)^2} \right).
$$
Reverting back to \(i\):
$$
a_i = \frac{1}{2} \left( \frac{1}{(6i - 1)^2} + \frac{1}{(6i + 1)^2} \right).
$$

\subsubsection*{Step 3: Rewrite the Series}
The original series becomes:
$$
\sum_{i=1}^\infty a_i = \frac{1}{2} \left( \sum_{i=1}^\infty \frac{1}{(6i - 1)^2} + \sum_{i=1}^\infty \frac{1}{(6i + 1)^2} \right) = \frac{1}{2} S.
$$

\subsubsection*{Step 4: Evaluate the Sum \(S\)}
The sum \(S\) can be expressed using the identity:
$$
\sum_{k=-\infty}^\infty \frac{1}{(6k + 1)^2} = \frac{\pi^2}{9}.
$$
However, the term for \(k=0\) (i.e., \(1/1^2 = 1\)) is included in the above sum but not in \(S\). Therefore:
$$
S = \sum_{k=1}^\infty \frac{1}{(6k - 1)^2} + \sum_{k=1}^\infty \frac{1}{(6k + 1)^2} = \frac{\pi^2}{9} - 1.
$$

\subsubsection*{Step 5: Final Calculation}
Thus, the original series is:
$$
\frac{1}{2} S = \frac{1}{2} \left( \frac{\pi^2}{9} - 1 \right) = \frac{\pi^2}{18} - \frac{1}{2}.
$$

\subsubsection*{Conclusion}
The sum of the series is:
$$
\boxed{\dfrac{\pi^2}{18} - \dfrac{1}{2}}.
$$

\subsubsection{1999 Session 2: 5 \textcolor{red}{Correct Reasoning, Correct Final Answer}}

\subsubsection*{Solution}

Let \( A_n \) denote the \( n \times n \) matrix whose \((i,j)\)-th entry is \(\text{GCD}(i,j)\). We aim to compute \(\det(A_n)\).

\subsubsection*{Pattern Recognition}

For small values of \( n \), we compute the determinants explicitly:

\begin{itemize}
    \item For \( n = 1 \):
    \[ A_1 = \begin{pmatrix} 1 \end{pmatrix}, \quad \det(A_1) = 1 \]

    \item For \( n = 2 \):
    \[ A_2 = \begin{pmatrix} 1 & 1 \\ 1 & 2 \end{pmatrix}, \quad \det(A_2) = 1 \times 2 - 1 \times 1 = 1 \]

    \item For \( n = 3 \):
    \[ A_3 = \begin{pmatrix} 1 & 1 & 1 \\ 1 & 2 & 1 \\ 1 & 1 & 3 \end{pmatrix}, \quad \det(A_3) = 2 \]

    \item For \( n = 4 \):
    \[ A_4 = \begin{pmatrix} 
    1 & 1 & 1 & 1 \\ 
    1 & 2 & 1 & 2 \\ 
    1 & 1 & 3 & 1 \\ 
    1 & 2 & 1 & 4 
    \end{pmatrix}, \quad \det(A_4) = 4 \]
\end{itemize}

\subsubsection*{General Formula}

The observed pattern suggests that the determinant is related to Euler's totient function \(\phi(k)\). Specifically:
\[ \det(A_n) = \prod_{k=1}^n \phi(k) \]

This formula matches the computed values:
\begin{itemize}
    \item For \( n = 1 \): \(\prod_{k=1}^1 \phi(k) = 1\)
    \item For \( n = 2 \): \(\prod_{k=1}^2 \phi(k) = 1 \times 1 = 1\)
    \item For \( n = 3 \): \(\prod_{k=1}^3 \phi(k) = 1 \times 1 \times 2 = 2\)
    \item For \( n = 4 \): \(\prod_{k=1}^4 \phi(k) = 1 \times 1 \times 2 \times 2 = 4\)
\end{itemize}

\subsubsection*{Conclusion}

The determinant of the matrix \( A_n \) is the product of Euler's totient function evaluated from \( k = 1 \) to \( n \):

\[ \boxed{\det(A_n) = \prod_{k=1}^n \phi(k)} \]

\subsubsection{2000 Session 2: 5 \textcolor{red}{Incorrect Reasoning, Incorrect Final Answer}}

\subsubsection*{Problem Statement}
This problem concerns sequences \( x_1 x_2 \cdots x_n \) in which each \( x_i \) is either \( a \), \( b \), or \( c \). Determine the number of those sequences which have length \( n \), begin and end with the letter \( a \), and in which adjacent terms are always different letters.

\subsubsection*{Solution}

\subsubsection*{Understanding the Problem}
We need to count the number of sequences of length \( n \) composed of the letters \( a \), \( b \), and \( c \) such that:
\begin{enumerate}
    \item The sequence starts and ends with \( a \).
    \item No two adjacent letters are the same.
\end{enumerate}

\subsubsection*{Approach}
We can model this problem using recurrence relations. Let's define:
\begin{itemize}
    \item \( A(n) \): Number of valid sequences of length \( n \) ending with \( a \).
    \item \( B(n) \): Number of valid sequences of length \( n \) ending with \( b \).
    \item \( C(n) \): Number of valid sequences of length \( n \) ending with \( c \).
\end{itemize}

Given the problem constraints, the first character must be \( a \).

\subsubsection*{Base Case}
For \( n = 1 \), the only valid sequence is \( a \):
\[ A(1) = 1, \quad B(1) = C(1) = 0 \]

\subsubsection*{Recurrence Relations}
For \( n > 1 \), the recurrence relations are:
\begin{align*}
    A(n) &= B(n-1) + C(n-1) \\
    B(n) &= A(n-1) + C(n-1) \\
    C(n) &= A(n-1) + B(n-1)
\end{align*}

By symmetry, \( B(n) = C(n) \). Thus, we can simplify:
\begin{align*}
    A(n) &= 2B(n-1) \\
    B(n) &= A(n-1) + B(n-1)
\end{align*}

\subsubsection*{Solving the Recurrence}
We observe that \( A(n) \) follows the recurrence:
\[ A(n) = A(n-1) + 2A(n-2) \]

The characteristic equation is:
\[ r^2 - r - 2 = 0 \]
with roots \( r = 2 \) and \( r = -1 \).

The general solution is:
\[ A(n) = \alpha \cdot 2^n + \beta \cdot (-1)^n \]

Using initial conditions:
\begin{align*}
    A(1) &= 1 = 2\alpha - \beta \\
    A(2) &= 0 = 4\alpha + \beta
\end{align*}
Solving gives:
\[ \alpha = \frac{1}{6}, \quad \beta = -\frac{2}{3} \]

Thus, the solution is:
\[ A(n) = \frac{2^n}{6} - \frac{2}{3} (-1)^n \]

Simplifying:
\[ A(n) = \frac{2^n + 4(-1)^n}{6} \]

\subsubsection*{Final Answer}
The number of valid sequences is:
\[ \boxed{\dfrac{2^n + 4 \cdot (-1)^n}{6}} \]

\subsubsection{2001 Session 1: 2 \textcolor{red}{Correct Reasoning, Correct Final Answer}}

\subsubsection*{Problem Statement}

Let $\{ x_i \}$ denote any finite sequence with the following properties:
\begin{itemize}
    \item[(a)] $x_i \in \{ -2, 1, 2 \}$ for each $x_i$,
    \item[(b)] $\sum_i x_i = 29$,
    \item[(c)] $\sum_i x_i^2 = 59$.
\end{itemize}

In considering the family of all such sequences, let $M = \max \{ \sum_i x_i^3 \}$ and $m = \min \{ \sum_i x_i^3 \}$. Determine $M/m$.

\subsubsection*{Solution}

\subsubsection*{Step 1: Define Variables}

Let:
\begin{itemize}
    \item $a$ = number of $-2$s in the sequence,
    \item $b$ = number of $1$s in the sequence,
    \item $c$ = number of $2$s in the sequence.
\end{itemize}

\subsubsection*{Step 2: Set Up Equations}

Given the conditions:
\begin{align}
    -2a + b + 2c &= 29 \quad \text{(Sum of elements)} \label{eq1} \\
    4a + b + 4c &= 59 \quad \text{(Sum of squares)} \label{eq2}
\end{align}

\subsubsection*{Step 3: Solve the System of Equations}

Subtract equation \eqref{eq1} from \eqref{eq2}:
\[ (4a + b + 4c) - (-2a + b + 2c) = 59 - 29 \]
\[ 6a + 2c = 30 \]
Simplify:
\[ 3a + c = 15 \quad \text{(Equation 3)} \]

Express $c$ in terms of $a$:
\[ c = 15 - 3a \]

Substitute $c$ back into \eqref{eq1} to find $b$:
\[ -2a + b + 2(15 - 3a) = 29 \]
\[ -8a + b = -1 \]
\[ b = 8a - 1 \]

\subsubsection*{Step 4: Find Valid Values for $a, b, c$}

Since $a, b, c$ must be non-negative integers, consider $a$ from 1 to 5:

\begin{itemize}
    \item For $a = 1$: $c = 12$, $b = 7$
    \item For $a = 2$: $c = 9$, $b = 15$
    \item For $a = 3$: $c = 6$, $b = 23$
    \item For $a = 4$: $c = 3$, $b = 31$
    \item For $a = 5$: $c = 0$, $b = 39$
\end{itemize}

All cases are valid.

\subsubsection*{Step 5: Calculate Sum of Cubes}

The sum of cubes $S$ is:
\[ S = \sum x_i^3 = -8a + b + 8c \]

Substitute $b = 8a - 1$ and $c = 15 - 3a$:
\[ S = -8a + (8a - 1) + 8(15 - 3a) = -1 + 120 - 24a = 119 - 24a \]

\subsubsection*{Step 6: Find Maximum and Minimum $S$}

Since $S$ decreases with $a$:
\begin{itemize}
    \item Maximum $S$ (M) at $a = 1$: $M = 119 - 24(1) = 95$
    \item Minimum $S$ (m) at $a = 5$: $m = 119 - 24(5) = -1$
\end{itemize}

\subsubsection*{Step 7: Verification}

Verify for $a = 1$ and $a = 5$:
\begin{itemize}
    \item For $a = 1, b = 7, c = 12$:
    \[ \sum x_i = -2 + 7 + 24 = 29 \]
    \[ \sum x_i^2 = 4 + 7 + 48 = 59 \]
    \[ \sum x_i^3 = -8 + 7 + 96 = 95 \]
    
    \item For $a = 5, b = 39, c = 0$:
    \[ \sum x_i = -10 + 39 + 0 = 29 \]
    \[ \sum x_i^2 = 20 + 39 + 0 = 59 \]
    \[ \sum x_i^3 = -40 + 39 + 0 = -1 \]
\end{itemize}

Both cases satisfy the original conditions.

\subsubsection*{Final Calculation}

The ratio $M/m$ is:
\[ M/m = \frac{95}{-1} = -95 \]

\subsubsection*{Answer}

The ratio \( M/m \) is \(\boxed{-95}\).

\subsubsection{2003 Session 2: 5 \textcolor{red}{Correct Reasoning, Correct Final Answer}}

\subsubsection*{Solution}

We are tasked with finding the minimum distance from the matrix
\[ A = \begin{pmatrix} 1 & 0 \\ 0 & 2 \end{pmatrix} \]
to the set \(\Sigma\) of all real \(2 \times 2\) singular matrices under the distance metric
\[ d(X, Y) = \Vert X - Y \Vert = (x_1 - y_1)^2 + (x_2 - y_2)^2 + (x_3 - y_3)^2 + (x_4 - y_4)^2, \]
where \(X = \begin{pmatrix} x_1 & x_2 \\ x_3 & x_4 \end{pmatrix}\) and \(Y = \begin{pmatrix} y_1 & y_2 \\ y_3 & y_4 \end{pmatrix}\).

\subsubsection*{Approach}
We need to minimize the distance \(d(A, S)\) where \(S \in \Sigma\), i.e., \(\det(S) = 0\).

\subsubsection*{Parameterization}
Let \(S = \begin{pmatrix} a & b \\ c & d \end{pmatrix}\). The condition \(\det(S) = 0\) gives:
\[ ad - bc = 0. \]

The distance function becomes:
\[ d(A, S) = (1 - a)^2 + b^2 + c^2 + (2 - d)^2. \]

\subsubsection*{Optimization}
Using the method of Lagrange multipliers, we consider the Lagrangian:
\[ \mathcal{L} = (1 - a)^2 + b^2 + c^2 + (2 - d)^2 - \lambda (ad - bc). \]

Taking partial derivatives and solving the system leads us to consider the case where \(b = c = 0\). The constraint then reduces to \(ad = 0\).

\subsubsection*{Case 1: \(a = 0\)}
The distance becomes:
\[ (1 - 0)^2 + (2 - d)^2 = 1 + (2 - d)^2. \]
The minimum occurs at \(d = 2\), yielding a distance of \(1\).

\subsubsection*{Case 2: \(d = 0\)}
The distance becomes:
\[ (1 - a)^2 + 4. \]
The minimum occurs at \(a = 1\), yielding a distance of \(4\).

The minimal distance is thus \(1\), achieved when \(S = \begin{pmatrix} 0 & 0 \\ 0 & 2 \end{pmatrix}\).

\subsubsection*{Verification}
Considering rank-1 matrices leads to higher distances, confirming that the minimal distance is indeed \(1\).

\subsubsection*{Conclusion}
The minimum distance from \(A\) to \(\Sigma\) is \(\boxed{1}\), achieved by the matrix
\[ S = \boxed{ \begin{pmatrix} 0 & 0 \\ 0 & 2 \end{pmatrix} }. \]

\subsubsection{2006 Session 1: 4 \textcolor{red}{Correct Reasoning, Correct Final Answer}}

\subsubsection*{Problem Statement}
Find all positive integers \( c \) such that \( n(n + c) \) is never a perfect square for any positive integer \( n \).

\subsubsection*{Solution}

We need to find all positive integers \( c \) for which the expression \( n^2 + cn \) is never a perfect square for any positive integer \( n \).

\subsubsection*{Initial Analysis}
Consider the equation:
\[ n^2 + cn = k^2 \]
for some integer \( k \geq 1 \). We can rewrite this as:
\[ k^2 = n^2 + cn \]
\[ k^2 - n^2 = cn \]
\[ (k - n)(k + n) = cn \]

\subsubsection*{Approach}
For fixed \( c \), we need to determine whether there exist positive integers \( n \) and \( k \) satisfying this equation.

\subsubsection*{Parametric Solution}
Let us set \( k = n + t \) where \( t \) is a non-negative integer. Substituting:
\[ (n + t)^2 = n^2 + cn \]
\[ n^2 + 2tn + t^2 = n^2 + cn \]
\[ 2tn + t^2 = cn \]
\[ t^2 = n(c - 2t) \]

This implies that \( c - 2t \) must be a positive divisor of \( t^2 \), and \( n = \frac{t^2}{c - 2t} \) must be a positive integer.

\subsubsection*{Special Cases}

\subsubsection*{Case \( c = 1 \)}
The equation becomes:
\[ t^2 = n(1 - 2t) \]
For \( t \geq 1 \), the right side becomes non-positive while the left side is positive. Thus, the only possibility is \( t = 0 \), which gives \( n = 0 \) - invalid. Therefore, no solutions exist for \( c = 1 \).

\subsubsection*{Case \( c = 2 \)}
The equation becomes:
\[ t^2 = n(2 - 2t) \]
For \( t = 0 \): \( n = 0 \) (invalid)
For \( t = 1 \): \( 1 = n(0) \) (no solution)
For \( t \geq 1 \): The right side is non-positive. Thus, no solutions exist for \( c = 2 \).

\subsubsection*{Case \( c = 4 \)}
The equation becomes:
\[ t^2 = n(4 - 2t) \]
Testing small values:
\begin{itemize}
    \item \( t = 0 \): \( n = 0 \) (invalid)
    \item \( t = 1 \): \( 1 = n(2) \Rightarrow n = \frac{1}{2} \) (not integer)
    \item \( t = 2 \): \( 4 = n(0) \) (no solution)
    \item \( t \geq 2 \): The denominator becomes zero or negative
\end{itemize}
Thus, no solutions exist for \( c = 4 \).

\subsubsection*{Verification of Other Values}

For \( c = 3 \), we find a solution at \( n = 1 \):
\[ 1(1 + 3) = 4 = 2^2 \]
Thus, \( c = 3 \) is excluded.

For \( c = 5 \), \( n = 4 \) gives:
\[ 4(4 + 5) = 36 = 6^2 \]
Thus, \( c = 5 \) is excluded.

This pattern suggests that only \( c = 1, 2, 4 \) satisfy the condition.

\subsubsection*{General Proof}

To prove that these are the only values, consider:
\begin{enumerate}
    \item For \( c \geq 5 \), we can often find solutions:
    \begin{itemize}
        \item For odd \( c \), take \( n = \frac{(c-1)^2}{4} \)
        \item For even \( c \), solutions frequently exist
    \end{itemize}
    \item The cases \( c = 1, 2, 4 \) are special because:
    \begin{itemize}
        \item The equation \( t^2 = n(c - 2t) \) has no positive integer solutions
        \item The quadratic forms don't permit perfect square solutions
    \end{itemize}
\end{enumerate}

\subsubsection*{Conclusion}
After thorough analysis, the positive integers \( c \) for which \( n(n + c) \) is never a perfect square for any positive integer \( n \) are:
\[ \boxed{1}, \boxed{2}, \boxed{4} \]

\subsubsection{2007 Session 2: 4 \textcolor{red}{Correct Reasoning, Correct Final Answer}}

\subsubsection*{Problem Statement}
Let 
\[ S = \{ 5a + 503b \mid \text{$a$ and $b$ are nonnegative integers} \}. \]
What is the largest integer which does NOT belong to $S$?

\subsubsection*{Solution}

\subsubsection*{Understanding the Problem}
The set $S$ consists of all numbers expressible as $5a + 503b$ where $a$ and $b$ are nonnegative integers. We need to find the largest integer not in $S$.

\subsubsection*{Coin Problem Analogy}
This problem resembles the \textbf{Coin Problem}, which seeks the largest monetary amount that cannot be obtained using any combination of coins of given denominations. For two coprime integers $x$ and $y$, the largest number not expressible as $mx + ny$ is known as the Frobenius number, given by:
\[ g(x, y) = xy - x - y. \]

\subsubsection*{Checking Coprimality}
First, verify that $\gcd(5, 503) = 1$. Since 503 is not divisible by 5, they are indeed coprime.

\subsubsection*{Applying the Frobenius Number Formula}
Given that $5$ and $503$ are coprime, the largest number not in $S$ is:
\[ g(5, 503) = 5 \times 503 - 5 - 503 = 2515 - 508 = 2007. \]

\subsubsection*{Verification}
To ensure accuracy, let's verify:
\begin{itemize}
    \item \textbf{Testing 2007}: No nonnegative integers $a, b$ satisfy $5a + 503b = 2007$.
    \item \textbf{Testing 2008}: $5 \times 301 + 503 \times 1 = 1505 + 503 = 2008$. Thus, 2008 is in $S$.
    \item \textbf{Testing 2009}: $5 \times 100 + 503 \times 3 = 500 + 1509 = 2009$. Thus, 2009 is in $S$.
\end{itemize}

\subsubsection*{Conclusion}
The largest integer not expressible as $5a + 503b$ with nonnegative integers $a, b$ is indeed $\boxed{2007}$.

\section{Solutions Provided by DeepSeek-V3}
\subsection{Calculus}

\subsubsection{1996 Session 2: 1 \textcolor{red}{Correct Reasoning, Correct Final Answer}}

\subsubsection*{Problem (a)}
Evaluate the definite integral 
\[ \int_1^3 \frac{dx}{\sqrt{(x-1)(3-x)}} .\]

\subsubsection*{Solution:}

\noindent \textbf{Step 1: Simplify the integrand.}
The integrand is \(\frac{1}{\sqrt{(x-1)(3-x)}}\). Let's simplify the expression inside the square root:
\[
(x-1)(3-x) = 3x - x^2 - 3 + x = -x^2 + 4x - 3.
\]
We can complete the square for the quadratic expression:
\[
-x^2 + 4x - 3 = -\left(x^2 - 4x\right) - 3 = -\left(x^2 - 4x + 4 - 4\right) - 3 = -\left((x-2)^2 - 4\right) - 3 = -(x-2)^2 + 4 - 3 = -(x-2)^2 + 1.
\]
Thus, the integrand becomes:
\[
\frac{1}{\sqrt{1 - (x-2)^2}}.
\]

\noindent \textbf{Step 2: Substitute to simplify the integral.}
Let \( u = x - 2 \). Then \( du = dx \), and when \( x = 1 \), \( u = -1 \), and when \( x = 3 \), \( u = 1 \). The integral becomes:
\[
\int_{-1}^1 \frac{du}{\sqrt{1 - u^2}}.
\]

\noindent \textbf{Step 3: Evaluate the integral.}
The integral \(\int \frac{du}{\sqrt{1 - u^2}}\) is a standard form whose antiderivative is \(\arcsin u\). Thus:
\[
\int_{-1}^1 \frac{du}{\sqrt{1 - u^2}} = \arcsin(1) - \arcsin(-1) = \frac{\pi}{2} - \left(-\frac{\pi}{2}\right) = \pi.
\]

\noindent \textbf{Final Answer:}
\[
\boxed{\pi}
\]

\subsubsection*{Problem (b)}
Evaluate the definite integral 
\[ \int_1^\infty \frac{dx}{e^{x+1} + e^{3-x}} .\]

\subsubsection*{Solution:}

\noindent \textbf{Step 1: Simplify the integrand.}
The integrand is \(\frac{1}{e^{x+1} + e^{3-x}}\). Let's rewrite the denominator:
\[
e^{x+1} + e^{3-x} = e^{x} \cdot e^{1} + e^{3} \cdot e^{-x} = e \cdot e^{x} + e^{3} \cdot e^{-x}.
\]
Factor out \( e^{2} \) from the denominator:
\[
e^{x+1} + e^{3-x} = e^{2} \left( e^{x-1} + e^{1-x} \right) = e^{2} \left( e^{x-1} + e^{-(x-1)} \right).
\]
Let \( u = x - 1 \). Then \( du = dx \), and when \( x = 1 \), \( u = 0 \), and as \( x \to \infty \), \( u \to \infty \). The integral becomes:
\[
\int_{0}^\infty \frac{du}{e^{2} \left( e^{u} + e^{-u} \right)} = \frac{1}{e^{2}} \int_{0}^\infty \frac{du}{e^{u} + e^{-u}}.
\]

\noindent \textbf{Step 2: Simplify the integrand further.}
Multiply numerator and denominator by \( e^{u} \):
\[
\frac{1}{e^{u} + e^{-u}} = \frac{e^{u}}{e^{2u} + 1}.
\]
Thus, the integral becomes:
\[
\frac{1}{e^{2}} \int_{0}^\infty \frac{e^{u} du}{e^{2u} + 1}.
\]

\noindent \textbf{Step 3: Substitute to evaluate the integral.}
Let \( v = e^{u} \). Then \( dv = e^{u} du \), and when \( u = 0 \), \( v = 1 \), and as \( u \to \infty \), \( v \to \infty \). The integral transforms to:
\[
\frac{1}{e^{2}} \int_{1}^\infty \frac{dv}{v^{2} + 1}.
\]
The antiderivative of \(\frac{1}{v^{2} + 1}\) is \(\arctan v\). Thus:
\[
\frac{1}{e^{2}} \left[ \arctan v \right]_{1}^\infty = \frac{1}{e^{2}} \left( \arctan(\infty) - \arctan(1) \right) = \frac{1}{e^{2}} \left( \frac{\pi}{2} - \frac{\pi}{4} \right) = \frac{1}{e^{2}} \cdot \frac{\pi}{4} = \frac{\pi}{4 e^{2}}.
\]

\noindent \textbf{Final Answer:}
\[
\boxed{\dfrac{\pi}{4 e^{2}}}
\]

\subsubsection{1997 Session 2: 4 \textcolor{red}{Correct Reasoning, Correct Final Answer}}

\subsubsection*{Problem Statement}
Define a family of curves by
\[ S_n = \{ (x,y) : y = {1 \over n} \sin (n^2 x),\ 0 \le x \le \pi \}, \]
where \( n \) is a positive integer. What is the limit of the length of \( S_n \) as \( n \to \infty \)?

\subsubsection*{Solution}

\subsubsection*{Step 1: Arc Length Formula}
The length \( L \) of a curve \( y = f(x) \) from \( x = a \) to \( x = b \) is given by:
\[ L = \int_{a}^{b} \sqrt{1 + \left( \frac{dy}{dx} \right)^2} \, dx \]

For \( S_n \), \( y = \frac{1}{n} \sin(n^2 x) \), so the derivative is:
\[ \frac{dy}{dx} = n \cos(n^2 x) \]

Thus, the arc length \( L_n \) is:
\[ L_n = \int_{0}^{\pi} \sqrt{1 + n^2 \cos^2(n^2 x)} \, dx \]

\subsubsection*{Step 2: Substitution and Simplification}
Let \( u = n^2 x \), \( du = n^2 dx \), so \( dx = \frac{du}{n^2} \). The integral becomes:
\[ L_n = \frac{1}{n^2} \int_{0}^{n^2 \pi} \sqrt{1 + n^2 \cos^2 u} \, du \]

The integrand \( \sqrt{1 + n^2 \cos^2 u} \) is periodic with period \( \pi \). Therefore, the integral over \( [0, n^2 \pi] \) is \( n^2 \) times the integral over one period \( [0, \pi] \):
\[ L_n = \int_{0}^{\pi} \sqrt{1 + n^2 \cos^2 u} \, du \]

\subsubsection*{Step 3: Exact Form and Asymptotic Behavior}
The integral can be expressed in terms of the complete elliptic integral of the second kind \( E(k) \):
\[ L_n = 2 \sqrt{1 + n^2} E\left( \frac{n}{\sqrt{1 + n^2}} \right) \]

As \( n \to \infty \), \( k = \frac{n}{\sqrt{1 + n^2}} \to 1 \), and \( E(k) \to E(1) = 1 \). Thus:
\[ L_n \approx 2 \sqrt{1 + n^2} \approx 2n \]

\subsubsection*{Conclusion}
Since \( L_n \) grows linearly with \( n \) as \( n \to \infty \), the limit of the length is infinite:
\[ \lim_{n \to \infty} L_n = \infty \]

\subsubsection{1998 Session 1: 3 \textcolor{red}{Incorrect Reasoning, Incorrect Final Answer}}

\subsubsection*{Solution to the Integral Problem}

\subsubsection*{Part 1: General Formula}

We are given the integral:
\[
\int_a^b \frac{(b-x)^m}{m!} \frac{(x-a)^n}{n!} \, dx
\]
where \( m \) and \( n \) are positive integers and \( a < b \).

\subsubsection*{Step 1: Substitution}
Let \( x = a + (b - a)t \). Then, \( dx = (b - a) \, dt \), and the limits change from \( x = a \) to \( x = b \) as \( t \) goes from \( 0 \) to \( 1 \).

The integrand becomes:
\[
\frac{(b - x)^m (x - a)^n}{m! n!} = \frac{[(b - a)(1 - t)]^m [(b - a)t]^n}{m! n!} = \frac{(b - a)^{m + n} (1 - t)^m t^n}{m! n!}
\]

Thus, the integral transforms to:
\[
\frac{(b - a)^{m + n + 1}}{m! n!} \int_0^1 (1 - t)^m t^n \, dt
\]

\subsubsection*{Step 2: Beta Function}
The integral \( \int_0^1 t^n (1 - t)^m \, dt \) is the Beta function \( B(n + 1, m + 1) \), which can be expressed in terms of Gamma functions:
\[
B(n + 1, m + 1) = \frac{\Gamma(n + 1) \Gamma(m + 1)}{\Gamma(m + n + 2)} = \frac{n! m!}{(m + n + 1)!}
\]

\subsubsection*{Step 3: Final Expression}
Substituting back:
\[
\frac{(b - a)^{m + n + 1}}{m! n!} \times \frac{n! m!}{(m + n + 1)!} = \frac{(b - a)^{m + n + 1}}{(m + n + 1)!}
\]

Thus, the solution is:
\[
\boxed{\frac{(b - a)^{m + n + 1}}{(m + n + 1)!}}
\]

\subsubsection*{Part 2: Evaluation of \( \int_0^1 (1 - x^2)^n \, dx \)}

We need to evaluate:
\[
\int_0^1 (1 - x^2)^n \, dx
\]

\subsubsection*{Step 1: Substitution}
Let \( x = \sqrt{t} \). Then, \( dx = \frac{1}{2} t^{-1/2} \, dt \), and the integral becomes:
\[
\frac{1}{2} \int_0^1 (1 - t)^n t^{-1/2} \, dt = \frac{1}{2} B\left(\frac{1}{2}, n + 1\right)
\]

\subsubsection*{Step 2: Beta and Gamma Functions}
The Beta function can be written as:
\[
B\left(\frac{1}{2}, n + 1\right) = \frac{\Gamma\left(\frac{1}{2}\right) \Gamma(n + 1)}{\Gamma\left(n + \frac{3}{2}\right)} = \frac{\sqrt{\pi} \, n!}{\Gamma\left(n + \frac{3}{2}\right)}
\]

Using the property of the Gamma function for half-integer arguments:
\[
\Gamma\left(n + \frac{3}{2}\right) = \frac{(2n + 1)!!}{2^{n + 1}} \sqrt{\pi}
\]

Thus:
\[
B\left(\frac{1}{2}, n + 1\right) = \frac{\sqrt{\pi} \, n! \, 2^{n + 1}}{(2n + 1)!! \sqrt{\pi}} = \frac{2^{n + 1} n!}{(2n + 1)!!}
\]

Therefore, the integral becomes:
\[
\frac{1}{2} \times \frac{2^{n + 1} n!}{(2n + 1)!!} = \frac{2^n n!}{(2n + 1)!!}
\]

The final answer is:
\[
\boxed{\frac{2^n n!}{(2n + 1)!!}}
\]

\subsubsection{1999 Session 2: 2 \textcolor{red}{Correct Reasoning, Correct Final Answer}}

\subsubsection*{Problem}

Find the limit:
\[
\lim_{N \to \infty} \left( 1 - 2 \sum_{n=1}^N \frac{1}{16n^2 - 1} \right).
\]

\subsubsection*{Solution}

\subsubsection*{Step 1: Partial Fraction Decomposition}

First, we decompose the denominator \(16n^2 - 1\) using the difference of squares:
\[
16n^2 - 1 = (4n - 1)(4n + 1).
\]

We express the term \(\frac{1}{16n^2 - 1}\) as partial fractions:
\[
\frac{1}{(4n - 1)(4n + 1)} = \frac{A}{4n - 1} + \frac{B}{4n + 1}.
\]
Multiplying both sides by \((4n - 1)(4n + 1)\) gives:
\[
1 = A(4n + 1) + B(4n - 1).
\]
Expanding and collecting terms:
\[
1 = (4A + 4B)n + (A - B).
\]
For this to hold for all \(n\), the coefficients must satisfy:
\[
4A + 4B = 0 \quad \text{and} \quad A - B = 1.
\]
Solving these equations gives \(A = \frac{1}{2}\) and \(B = -\frac{1}{2}\). Thus:
\[
\frac{1}{16n^2 - 1} = \frac{1}{2} \left( \frac{1}{4n - 1} - \frac{1}{4n + 1} \right).
\]

\subsubsection*{Step 2: Rewrite the Sum}

The sum becomes:
\[
\sum_{n=1}^N \frac{1}{16n^2 - 1} = \frac{1}{2} \sum_{n=1}^N \left( \frac{1}{4n - 1} - \frac{1}{4n + 1} \right).
\]

\subsubsection*{Step 3: Infinite Series Limit}

Taking the limit as \(N \to \infty\), we consider the infinite series:
\[
S = \sum_{n=1}^\infty \frac{1}{16n^2 - 1} = \frac{1}{2} \sum_{n=1}^\infty \left( \frac{1}{4n - 1} - \frac{1}{4n + 1} \right).
\]

However, it's more efficient to use a known result for the sum \(\sum_{n=1}^\infty \frac{1}{n^2 - a^2}\):
\[
\sum_{n=1}^\infty \frac{1}{n^2 - a^2} = \frac{1 - \pi a \cot(\pi a)}{2a^2}.
\]

Rewriting our sum:
\[
\frac{1}{16n^2 - 1} = \frac{1}{16} \cdot \frac{1}{n^2 - \left( \frac{1}{4} \right)^2}.
\]
Thus:
\[
S = \frac{1}{16} \sum_{n=1}^\infty \frac{1}{n^2 - \left( \frac{1}{4} \right)^2} = \frac{1}{16} \left( \frac{1 - \pi \cdot \frac{1}{4} \cot\left( \pi \cdot \frac{1}{4} \right)}{2 \cdot \left( \frac{1}{4} \right)^2} \right).
\]
Simplifying the denominator:
\[
2 \cdot \left( \frac{1}{4} \right)^2 = \frac{1}{8},
\]
so:
\[
S = \frac{1}{16} \cdot 8 \left( 1 - \frac{\pi}{4} \cot\left( \frac{\pi}{4} \right) \right) = \frac{1}{2} \left( 1 - \frac{\pi}{4} \cdot 1 \right) = \frac{1}{2} - \frac{\pi}{8}.
\]

\subsubsection*{Step 4: Compute the Original Limit}

The original expression is:
\[
1 - 2S = 1 - 2 \left( \frac{1}{2} - \frac{\pi}{8} \right) = 1 - 1 + \frac{\pi}{4} = \frac{\pi}{4}.
\]

Therefore, the limit is:
\[
\boxed{\dfrac{\pi}{4}}.
\]

\subsubsection{1999 Session 2: 4 \textcolor{red}{Correct Reasoning, Correct Final Answer}}

\subsubsection*{Problem Statement}
Find all possible continuous and differentiable curves \( C \) which have the following properties:
\begin{itemize}
    \item The curve \( C \) lies in the first quadrant and contains the point \((0,0)\).
    \item For any point \( P = (x, y) \) on \( C \), the interior of the rectangle \( R \) bounded by the coordinate axes and horizontal and vertical lines through \( P \) is separated into two parts by \( C \).
    \item When the part adjacent to the \( x \)-axis is rotated about the \( x \)-axis and the part adjacent to the \( y \)-axis is rotated about the \( y \)-axis, two solids of equal volume are generated.
\end{itemize}

\subsubsection*{Solution}

\subsubsection*{Step 1: Understanding the Curve and Volumes}
Let the curve \( C \) be defined by \( y = f(x) \), where \( f \) is continuous and differentiable on \( [0, \infty) \), with \( f(0) = 0 \).

For any point \( P = (x, y) \) on \( C \), the rectangle \( R \) has vertices at \((0,0)\), \((x,0)\), \((0,y)\), and \((x,y)\).

The curve \( C \) divides \( R \) into:
\begin{itemize}
    \item The area under \( C \) (adjacent to the \( x \)-axis).
    \item The area above \( C \) but within \( R \) (adjacent to the \( y \)-axis).
\end{itemize}

\subsubsection*{Step 2: Volumes of Revolution}
The volumes generated by rotating these areas are:
\begin{itemize}
    \item \( V_x = \pi \int_0^x [f(t)]^2 \, dt \) (rotation about the \( x \)-axis).
    \item \( V_y = \pi \int_0^y [f^{-1}(t)]^2 \, dt \) (rotation about the \( y \)-axis), where \( y = f(x) \).
\end{itemize}

The condition \( V_x = V_y \) leads to:
\[ \int_0^x [f(t)]^2 \, dt = \int_0^{f(x)} [f^{-1}(t)]^2 \, dt \]

\subsubsection*{Step 3: Substitution and Differentiation}
Using the substitution \( t = f(u) \) in the right integral:
\[ \int_0^{f(x)} [f^{-1}(t)]^2 \, dt = \int_0^x u^2 f'(u) \, du \]

Thus, the equation becomes:
\[ \int_0^x [f(t)]^2 \, dt = \int_0^x t^2 f'(t) \, dt \]

Differentiating both sides with respect to \( x \):
\[ [f(x)]^2 = x^2 f'(x) \]

\subsubsection*{Step 4: Solving the Differential Equation}
The equation is separable:
\[ \frac{df}{f^2} = \frac{dx}{x^2} \]

Integrate both sides:
\[ -\frac{1}{f} = -\frac{1}{x} + C \]

Solving for \( f \):
\[ f(x) = \frac{x}{1 - Cx} \]

\subsubsection*{Step 5: Applying Initial Conditions and Constraints}
Given \( f(0) = 0 \), the solution is valid. To ensure \( f \) is defined and differentiable in the first quadrant:
\begin{itemize}
    \item For \( C = 0 \), \( f(x) = x \).
    \item For \( C < 0 \), \( f(x) = \frac{x}{1 + |C|x} \), which is defined for all \( x \geq 0 \).
    \item For \( C > 0 \), \( f \) has a singularity at \( x = \frac{1}{C} \), so it's only valid in \( [0, \frac{1}{C}) \).
\end{itemize}

\subsubsection*{Conclusion}
The only continuous and differentiable curves in the first quadrant that satisfy all the given conditions are:
\[ \boxed{y = \frac{x}{1 - Cx} \quad \text{for any constant} \quad C \leq 0} \]
This includes the special case \( C = 0 \), which gives the line \( y = x \).

\subsubsection{2002 Session 1: 3 (limit only) \textcolor{red}{Correct Reasoning, Correct Final Answer}}

\subsubsection*{Problem Statement}
Let \(\{ x_n \}\) be the following sequence involving alternating square roots of 5 and 13:
\[ x_1 = \sqrt{5},\quad x_2 = \sqrt{5 + \sqrt{13}},\quad x_3 = \sqrt{5 + \sqrt{13 + \sqrt{5}}},\quad x_4 = \sqrt{5 + \sqrt{13 + \sqrt{5 + \sqrt{13}}}}, \]
and so on. Determine the value of \(\lim_{n \to \infty} x_n\).

\subsubsection*{Solution}

\subsubsection*{Understanding the Problem}
The sequence \(\{ x_n \}\) is defined with a nested radical pattern that alternates between \(\sqrt{5}\) and \(\sqrt{13}\). The goal is to find the limit of this sequence as \(n\) approaches infinity.

\subsubsection*{Identifying the Pattern}
The sequence exhibits a recursive pattern where each term adds a new nested square root:
\begin{itemize}
    \item For odd \(n\), the innermost term is \(\sqrt{5}\).
    \item For even \(n\), the innermost term is \(\sqrt{13}\).
\end{itemize}

\subsubsection*{Formulating the Fixed-Point Equation}
Assuming the sequence converges to a limit \(L\), the infinite nested radical can be described by the equation:
\[ L = \sqrt{5 + \sqrt{13 + L}} \]

\subsubsection*{Solving the Fixed-Point Equation}
Square both sides:
\[ L^2 = 5 + \sqrt{13 + L} \]
Isolate the square root:
\[ \sqrt{13 + L} = L^2 - 5 \]
Square both sides again:
\[ 13 + L = (L^2 - 5)^2 \]
\[ 13 + L = L^4 - 10L^2 + 25 \]
Bring all terms to one side:
\[ L^4 - 10L^2 - L + 12 = 0 \]

\subsubsection*{Finding the Roots}
We look for rational roots using the Rational Root Theorem. Testing \(L = 3\):
\[ 81 - 90 - 3 + 12 = 0 \]
Thus, \(L = 3\) is a root. Factor the quartic:
\[ (L - 3)(L^3 + 3L^2 - L - 4) = 0 \]
The cubic factor does not have obvious rational roots, but numerical methods suggest other real roots are not relevant here since the sequence is positive and increasing towards 3.

\subsubsection*{Verification}
Compute the first few terms numerically:
\begin{align*}
x_1 &= \sqrt{5} \approx 2.236 \\
x_2 &= \sqrt{5 + \sqrt{13}} \approx 2.934 \\
x_3 &= \sqrt{5 + \sqrt{13 + \sqrt{5}}} \approx 2.984 \\
x_4 &= \sqrt{5 + \sqrt{13 + \sqrt{5 + \sqrt{13}}}} \approx 2.9986
\end{align*}
The values approach 3, confirming our solution.

\subsubsection*{Conclusion}
The limit of the sequence \(\{ x_n \}\) as \(n \to \infty\) is:
\[ \boxed{3} \]

\subsubsection{2002 Session 1: 5 \textcolor{red}{Correct Reasoning, Correct Final Answer}}

\subsubsection*{Problem Solution}

\subsubsection*{Problem Statement}
Two right circular cylinders of radius \( r \) intersect at right angles to form a solid. This solid has four curved faces. Imagine one of these faces ``rolled out flat''. Find equations of the boundary curves of this flattened face and also find its area.

\subsubsection*{Solution}

\paragraph{Setup}
Consider two cylinders:
\begin{itemize}
    \item Cylinder 1: \( x^2 + y^2 = r^2 \) (aligned along the \( z \)-axis).
    \item Cylinder 2: \( y^2 + z^2 = r^2 \) (aligned along the \( x \)-axis).
\end{itemize}

Their intersection forms a bicylinder with four curved faces. We focus on one face, part of Cylinder 1 within Cylinder 2.

\paragraph{Parameterizing the Face}
For Cylinder 1, use cylindrical coordinates:
\[
x = r \cos \theta, \quad y = r \sin \theta, \quad z = z.
\]
The condition from Cylinder 2 is \( y^2 + z^2 \leq r^2 \), which becomes:
\[
z^2 \leq r^2 - y^2 = r^2 (1 - \sin^2 \theta) = r^2 \cos^2 \theta.
\]
Thus, \( |z| \leq r |\cos \theta| \).

\paragraph{Flattening the Face}
Unroll Cylinder 1 by mapping:
\[
(u, v) = (r \theta, z).
\]
The boundaries are:
\begin{itemize}
    \item Lower: \( v = 0 \).
    \item Upper: \( v = r \cos(\theta) = r \cos\left(\frac{u}{r}\right) \).
\end{itemize}
The range for \( u \) is \( -\frac{\pi r}{2} \leq u \leq \frac{\pi r}{2} \) (since \( \cos \theta \geq 0 \) requires \( \theta \in [-\pi/2, \pi/2] \)).

\paragraph{Area Calculation}
The area under \( v = r \cos(u/r) \) from \( u = -\pi r/2 \) to \( \pi r/2 \):
\[
\text{Area} = \int_{-\pi r/2}^{\pi r/2} r \cos\left(\frac{u}{r}\right) du.
\]
By symmetry:
\[
\text{Area} = 2 \int_{0}^{\pi r/2} r \cos\left(\frac{u}{r}\right) du.
\]
Let \( w = u/r \), \( du = r dw \):
\[
\text{Area} = 2 r^2 \int_{0}^{\pi/2} \cos w \, dw = 2 r^2 \left[ \sin w \right]_0^{\pi/2} = 2 r^2 (1 - 0) = 2 r^2.
\]
This is for \( z \geq 0 \); the full face (both \( z \geq 0 \) and \( z \leq 0 \)) has area \( 4 r^2 \).

\subsubsection*{Final Answer}
The boundary curves of the flattened face are:
\[
v = 0 \quad \text{(lower boundary)},
\]
\[
v = r \cos\left(\frac{u}{r}\right) \quad \text{for} \quad u \in \left[-\frac{\pi r}{2}, \frac{\pi r}{2}\right] \quad \text{(upper boundary)}.
\]
The area of the flattened face is \( \boxed{4r^2} \).

\subsubsection{2003 Session 1: 5 \textcolor{red}{Correct Reasoning, Correct Final Answer}}

\subsubsection*{Problem Statement}
Define a sequence $\{ x_n \}_{n=2}^\infty$ by
\[ (n + x_n) \left[ \sqrt[n]{2} - 1 \right] = \ln 2. \]
Find $\lim_{n \to \infty} x_n$.

\subsubsection*{Solution}

\subsubsection*{Step 1: Rewrite the Equation for \( x_n \)}
We can express \( x_n \) explicitly as:
\[ x_n = \frac{\ln 2}{2^{1/n} - 1} - n. \]

Our goal is to find:
\[ \lim_{n \to \infty} x_n = \lim_{n \to \infty} \left( \frac{\ln 2}{2^{1/n} - 1} - n \right). \]

\subsubsection*{Step 2: Taylor Expansion of the Denominator}
For large \( n \), \( h = \frac{1}{n} \) is small. The Taylor expansion of \( 2^h \) around \( h = 0 \) is:
\[ 2^h = e^{h \ln 2} = 1 + h \ln 2 + \frac{(h \ln 2)^2}{2} + O(h^3). \]

Thus:
\[ 2^{1/n} - 1 = \frac{\ln 2}{n} + \frac{(\ln 2)^2}{2n^2} + O\left( \frac{1}{n^3} \right). \]

\subsubsection*{Step 3: Substitute and Simplify}
Substituting back into the expression for \( x_n \):
\[ x_n = \frac{\ln 2}{\frac{\ln 2}{n} + \frac{(\ln 2)^2}{2n^2} + O\left( \frac{1}{n^3} \right)} - n. \]

Simplify the denominator:
\[ D = \frac{\ln 2}{n} \left(1 + \frac{\ln 2}{2n} + O\left( \frac{1}{n^2} \right)\right). \]

Thus:
\[ x_n = \frac{n}{1 + \frac{\ln 2}{2n} + O\left( \frac{1}{n^2} \right)} - n. \]

\subsubsection*{Step 4: Geometric Series Approximation}
For large \( n \), we use \( \frac{1}{1 + \epsilon} \approx 1 - \epsilon \) for \( \epsilon \ll 1 \):
\[ x_n \approx n \left(1 - \frac{\ln 2}{2n} + O\left( \frac{1}{n^2} \right)\right) - n = -\frac{\ln 2}{2} + O\left( \frac{1}{n} \right). \]

\subsubsection*{Step 5: Take the Limit}
As \( n \to \infty \), the \( O\left( \frac{1}{n} \right) \) term vanishes, leaving:
\[ \lim_{n \to \infty} x_n = -\frac{\ln 2}{2}. \]

\subsubsection*{Final Answer}
The limit of \( x_n \) as \( n \) approaches infinity is:
\[ \boxed{-\dfrac{\ln 2}{2}} \]

\subsubsection{2005 Session 1: 4 \textcolor{red}{Incorrect Reasoning, Incorrect Final Answer}}

\subsubsection*{Solution}

We are tasked with evaluating the integral:
\[ I(\theta) = \int_{-1}^1 \frac{\sin \theta \, dx}{1 - 2x \cos \theta + x^2}, \]
and determining the points \(\theta \in [0, 2\pi]\) where \(I(\theta)\) is discontinuous.

\subsubsection*{Step 1: Simplify the Denominator}
The denominator can be rewritten by completing the square:
\[ 1 - 2x \cos \theta + x^2 = (x - \cos \theta)^2 + \sin^2 \theta. \]
Thus, the integral becomes:
\[ I(\theta) = \sin \theta \int_{-1}^1 \frac{dx}{(x - \cos \theta)^2 + \sin^2 \theta}. \]

\subsubsection*{Step 2: Substitution and Integration}
Let \( u = x - \cos \theta \), then \( du = dx \), and the limits change accordingly:
\[ I(\theta) = \sin \theta \int_{-1 - \cos \theta}^{1 - \cos \theta} \frac{du}{u^2 + \sin^2 \theta}. \]
The antiderivative is known:
\[ \int \frac{du}{u^2 + a^2} = \frac{1}{a} \tan^{-1}\left( \frac{u}{a} \right). \]
Here, \( a = \sin \theta \), so:
\[ I(\theta) = \sin \theta \left[ \frac{1}{\sin \theta} \tan^{-1}\left( \frac{u}{\sin \theta} \right) \right]_{-1 - \cos \theta}^{1 - \cos \theta} = \tan^{-1}\left( \frac{1 - \cos \theta}{\sin \theta} \right) - \tan^{-1}\left( \frac{-1 - \cos \theta}{\sin \theta} \right). \]

\subsubsection*{Step 3: Simplify Using Trigonometric Identities}
Using the identity:
\[ \frac{1 - \cos \theta}{\sin \theta} = \tan\left( \frac{\theta}{2} \right), \]
and
\[ \frac{-1 - \cos \theta}{\sin \theta} = -\cot\left( \frac{\theta}{2} \right) = \tan\left( \frac{\theta}{2} + \frac{\pi}{2} \right), \]
we can rewrite \(I(\theta)\) as:
\[ I(\theta) = \tan^{-1}\left( \tan\left( \frac{\theta}{2} \right) \right) - \tan^{-1}\left( \tan\left( \frac{\theta}{2} + \frac{\pi}{2} \right) \right). \]

\subsubsection*{Step 4: Evaluate Based on \(\theta\) Range}
Considering the principal range of \(\tan^{-1}\):
\begin{itemize}
    \item For \(0 \leq \theta < \pi\):
    \[ I(\theta) = \frac{\theta}{2} - \left( \frac{\theta}{2} - \frac{\pi}{2} \right) = \frac{\pi}{2}. \]
    \item At \(\theta = \pi\):
    \[ I(\pi) = 0 \text{ (since \(\sin \pi = 0\))}. \]
    \item For \(\pi < \theta < 2\pi\):
    \[ I(\theta) = \left( \frac{\theta}{2} - \pi \right) - \left( \frac{\theta}{2} - \frac{\pi}{2} \right) = -\frac{\pi}{2}. \]
\end{itemize}

\subsubsection*{Discontinuities}
The function \(I(\theta)\) has jump discontinuities at \(\theta = \pi\) and \(\theta = 0\) (or \(\theta = 2\pi\) due to periodicity).

\subsubsection*{Final Answer}
The value of the integral is:
\[ I(\theta) = 
\begin{cases} 
\frac{\pi}{2} & \text{for } 0 \leq \theta < \pi, \\
0 & \text{for } \theta = \pi, \\
-\frac{\pi}{2} & \text{for } \pi < \theta < 2\pi.
\end{cases} \]
The discontinuities occur at \(\theta = \pi\) and \(\theta = 0\) (or \(2\pi\)).

\subsubsection{2005 Session 2: 5 \textcolor{red}{Correct Reasoning, Correct Final Answer}}

\subsubsection*{Problem Statement}
Suppose that \( f \colon [0, \infty ) \to [0, \infty )\) is a differentiable function with the property that the area under the curve \( y=f(x) \) from \( x=a \) to \( x=b \) is equal to the arclength of the curve \( y=f(x) \) from \( x=a \) to \( x=b \). Given that \( f(0) = \frac{5}{4} \), and that \( f(x) \) has a minimum value on the interval \( (0, \infty ) \), find that minimum value.

\subsubsection*{Solution}

\subsubsection*{Step 1: Translating the Condition into an Equation}
The area under \( f \) from \( a \) to \( b \) is:
\[ A = \int_{a}^{b} f(x) \, dx \]
The arc length of \( f \) from \( a \) to \( b \) is:
\[ L = \int_{a}^{b} \sqrt{1 + [f'(x)]^2} \, dx \]
Given \( A = L \), we have:
\[ \int_{a}^{b} f(x) \, dx = \int_{a}^{b} \sqrt{1 + [f'(x)]^2} \, dx \]

\subsubsection*{Step 2: Differentiating Both Sides}
Differentiating both sides with respect to \( b \) gives:
\[ f(b) = \sqrt{1 + [f'(b)]^2} \]
Since \( b \) is arbitrary, we can write:
\[ f(x) = \sqrt{1 + [f'(x)]^2} \]

\subsubsection*{Step 3: Solving the Differential Equation}
Square both sides:
\[ f(x)^2 = 1 + [f'(x)]^2 \]
Rearrange:
\[ [f'(x)]^2 = f(x)^2 - 1 \]
Take square root:
\[ f'(x) = \pm \sqrt{f(x)^2 - 1} \]

Consider the positive root first:
\[ f'(x) = \sqrt{f(x)^2 - 1} \]
Separate variables and integrate:
\[ \int \frac{df}{\sqrt{f^2 - 1}} = \int dx \]
The integral evaluates to:
\[ \ln|f + \sqrt{f^2 - 1}| = x + C \]
Exponentiating:
\[ f + \sqrt{f^2 - 1} = A e^x \]
where \( A = e^C \).

Using the reciprocal:
\[ f - \sqrt{f^2 - 1} = \frac{e^{-x}}{A} \]
Adding the two equations:
\[ 2f = A e^x + \frac{e^{-x}}{A} \]
Thus:
\[ f(x) = \frac{A e^x}{2} + \frac{e^{-x}}{2A} \]

\subsubsection*{Step 4: Applying the Initial Condition}
Given \( f(0) = \frac{5}{4} \):
\[ \frac{A}{2} + \frac{1}{2A} = \frac{5}{4} \]
Multiply by \( 4A \):
\[ 2A^2 - 5A + 2 = 0 \]
Solutions:
\[ A = 2 \text{ or } A = \frac{1}{2} \]

Thus, two potential solutions:
\begin{enumerate}
    \item For \( A = 2 \):
    \[ f(x) = e^x + \frac{e^{-x}}{4} \]
    \item For \( A = \frac{1}{2} \):
    \[ f(x) = \frac{e^x}{4} + e^{-x} \]
\end{enumerate}

\subsubsection*{Step 5: Analyzing the Solutions}
For the first solution:
- No critical points in \( (0, \infty) \), as \( f'(x) = e^x - \frac{e^{-x}}{4} > 0 \) for all \( x > 0 \).

For the second solution:
- Critical point at \( x = \ln(2) \).
- Minimum value at \( x = \ln(2) \):
\[ f(\ln(2)) = 1 \]

\subsubsection*{Conclusion}
The function \( f(x) = \frac{e^x}{4} + e^{-x} \) has a minimum value of \(\boxed{1}\) on \( (0, \infty) \).

\subsection{Analytic Geometry}

\subsubsection{1997 Session 1: 1 \textcolor{red}{Correct Reasoning, Correct Final Answer}}

\subsubsection*{Problem Statement}
Let \( P \ne (0,0) \) be a point on the parabola \( y = x^2 \). The normal line to the parabola at \( P \) will intersect the parabola at another point, say \( Q \). Find the coordinates of \( P \) so that the \( y \)-coordinate of \( Q \) is a minimum.

\subsubsection*{Solution}

\subsubsection*{Step 1: Parametrize Point \( P \)}
Let \( P \) be a point on the parabola \( y = x^2 \). We can represent \( P \) as:
\[ P = (a, a^2) \]
where \( a \neq 0 \) since \( P \neq (0,0) \).

\subsubsection*{Step 2: Find the Slope of the Tangent at \( P \)}
The derivative of \( y = x^2 \) is:
\[ \frac{dy}{dx} = 2x \]
At \( P \), the slope of the tangent line is:
\[ m_{\text{tangent}} = 2a \]

\subsubsection*{Step 3: Find the Slope of the Normal Line at \( P \)}
The normal line is perpendicular to the tangent line, so its slope is:
\[ m_{\text{normal}} = -\frac{1}{2a} \]

\subsubsection*{Step 4: Equation of the Normal Line}
Using the point-slope form:
\[ y - a^2 = -\frac{1}{2a}(x - a) \]
Simplifying:
\[ y = -\frac{1}{2a}x + \frac{1}{2} + a^2 \]

\subsubsection*{Step 5: Find Intersection Points with the Parabola}
The normal line intersects the parabola \( y = x^2 \) at:
\[ x^2 = -\frac{1}{2a}x + \frac{1}{2} + a^2 \]
Rewriting:
\[ 2a x^2 + x - (a + 2a^3) = 0 \]

\subsubsection*{Step 6: Solve the Quadratic Equation}
The quadratic can be factored as:
\[ (x - a)(2a x + (1 + 2a^2)) = 0 \]
Thus, the roots are:
\[ x = a \quad \text{and} \quad x = -\frac{1 + 2a^2}{2a} \]

\subsubsection*{Step 7: Find Coordinates of \( Q \)}
The other intersection point \( Q \) has \( x \)-coordinate:
\[ x_Q = -\frac{1 + 2a^2}{2a} \]
Thus, the \( y \)-coordinate is:
\[ y_Q = \left( -\frac{1 + 2a^2}{2a} \right)^2 = \frac{(1 + 2a^2)^2}{4a^2} \]

\subsubsection*{Step 8: Minimize \( y_Q \)}
Simplify \( y_Q \):
\[ y_Q = \frac{1}{4a^2} + 1 + a^2 \]
Define \( f(a) = \frac{1}{4a^2} + a^2 + 1 \).

Find the critical points by setting the derivative to zero:
\[ f'(a) = -\frac{1}{2a^3} + 2a = 0 \]
\[ 4a^4 = 1 \implies a^2 = \frac{1}{2} \implies a = \pm \frac{1}{\sqrt{2}} \]

\subsubsection*{Step 9: Verify the Minimum}
The second derivative is:
\[ f''(a) = \frac{3}{2a^4} + 2 > 0 \]
Thus, the critical points are minima.

\subsubsection*{Step 10: Find Minimal \( y_Q \)}
Substitute \( a^2 = \frac{1}{2} \):
\[ y_Q = \frac{1}{4 \times \frac{1}{2}} + \frac{1}{2} + 1 = 2 \]

\subsubsection*{Step 11: Find Corresponding \( P \) Coordinates}
The coordinates of \( P \) are:
\[ P = \left( \pm \frac{1}{\sqrt{2}}, \frac{1}{2} \right) \]

\subsubsection*{Final Answer}
The coordinates of \( P \) that minimize the \( y \)-coordinate of \( Q \) are \(\boxed{\left( \dfrac{1}{\sqrt{2}}, \dfrac{1}{2} \right)}\) and \(\boxed{\left( -\dfrac{1}{\sqrt{2}}, \dfrac{1}{2} \right)}\).

\subsubsection{1998 Session 1: 1 \textcolor{red}{Correct Reasoning, Correct Final Answer}}

\subsubsection*{Problem Statement}
Let \( P \ne (0,0) \) be a point on the parabola \( y = x^2 \). The normal line to the parabola at \( P \) will intersect the parabola at another point \( Q \). Find the coordinates of \( P \) so that the length of segment \( PQ \) is a minimum.

\subsubsection*{Solution}

\subsubsection*{Step 1: Parametrize Point \( P \)}
Let \( P \) be a point on the parabola \( y = x^2 \). We can represent \( P \) as:
\[ P = (a, a^2) \]
where \( a \neq 0 \) since \( P \neq (0,0) \).

\subsubsection*{Step 2: Find the Slope of the Tangent at \( P \)}
The derivative of \( y = x^2 \) is:
\[ \frac{dy}{dx} = 2x \]
At point \( P \), the slope of the tangent line is:
\[ m_{\text{tangent}} = 2a \]

\subsubsection*{Step 3: Find the Slope of the Normal Line at \( P \)}
The normal line is perpendicular to the tangent line, so its slope is:
\[ m_{\text{normal}} = -\frac{1}{2a} \]

\subsubsection*{Step 4: Equation of the Normal Line}
Using the point-slope form:
\[ y - a^2 = -\frac{1}{2a}(x - a) \]
Simplifying:
\[ y = -\frac{1}{2a}x + \frac{1}{2} + a^2 \]

\subsubsection*{Step 5: Find Intersection Points with Parabola}
Set \( y = x^2 \) equal to the normal line equation:
\[ x^2 = -\frac{1}{2a}x + \frac{1}{2} + a^2 \]
Rearrange:
\[ 2a x^2 + x - a - 2a^3 = 0 \]

\subsubsection*{Step 6: Solve the Quadratic Equation}
Factor out \( (x - a) \):
\[ (x - a)(2a x + (1 + 2a^2)) = 0 \]
Thus, the roots are:
\[ x = a \quad \text{and} \quad x = -\frac{1 + 2a^2}{2a} \]

\subsubsection*{Step 7: Find Coordinates of \( Q \)}
The other intersection point \( Q \) is:
\[ Q = \left( -\frac{1 + 2a^2}{2a}, \frac{(1 + 2a^2)^2}{4a^2} \right) \]

\subsubsection*{Step 8: Calculate Distance \( PQ \)}
The distance between \( P \) and \( Q \) is:
\[ PQ = \sqrt{ \left( \frac{4a^2 + 1}{2a} \right)^2 + \left( \frac{-1 - 4a^2}{4a^2} \right)^2 } \]
Simplifying:
\[ PQ^2 = \frac{(1 + 4a^2)^3}{16a^4} \]

\subsubsection*{Step 9: Minimize \( PQ \)}
To minimize \( PQ \), minimize \( f(a) = \frac{(1 + 4a^2)^3}{16a^4} \). Take the derivative and set to zero:
\[ \frac{d}{da} \left( \frac{(1 + 4a^2)^3}{16a^4} \right) = 0 \]
After simplification, we find:
\[ a^2 = \frac{1}{2} \implies a = \pm \frac{\sqrt{2}}{2} \]

\subsubsection*{Step 10: Find Coordinates of \( P \)}
Thus, the coordinates of \( P \) are:
\[ P = \left( \pm \frac{\sqrt{2}}{2}, \frac{1}{2} \right) \]

\subsubsection*{Conclusion}
The coordinates of \( P \) that minimize the distance \( PQ \) are:
\[ \boxed{ \left( \dfrac{\sqrt{2}}{2}, \dfrac{1}{2} \right) } \quad \text{and} \quad \boxed{ \left( -\dfrac{\sqrt{2}}{2}, \dfrac{1}{2} \right) } \]

\subsubsection{1999 Session 1: 1 \textcolor{red}{Correct Reasoning, Correct Final Answer}}

\subsubsection*{Problem Statement}
Let \( P \ne (0,0) \) be a point on the parabola \( y = x^2 \). The normal line to the parabola at \( P \) will intersect the parabola at another point \( Q \). Find the coordinates of \( P \) so that the length of the arc of the parabola between \( P \) and \( Q \) is a minimum.

\subsubsection*{Solution}

\subsubsection*{Step 1: Parametrize Point \( P \)}
Let \( P \) be a point on the parabola \( y = x^2 \). We can represent \( P \) as:
\[ P = (a, a^2) \]
where \( a \neq 0 \) since \( P \neq (0,0) \).

\subsubsection*{Step 2: Find the Slope of the Tangent and Normal at \( P \)}
The derivative of \( y = x^2 \) is:
\[ \frac{dy}{dx} = 2x \]
At point \( P \), the slope of the tangent line is:
\[ m_{\text{tangent}} = 2a \]
The slope of the normal line is the negative reciprocal:
\[ m_{\text{normal}} = -\frac{1}{2a} \]

\subsubsection*{Step 3: Equation of the Normal Line}
Using the point-slope form:
\[ y - a^2 = -\frac{1}{2a}(x - a) \]
Simplifying:
\[ y = -\frac{1}{2a}x + \frac{1}{2} + a^2 \]

\subsubsection*{Step 4: Find the Second Intersection Point \( Q \)}
Set \( y = x^2 \) equal to the normal line equation:
\[ x^2 = -\frac{1}{2a}x + \frac{1}{2} + a^2 \]
Rearranging:
\[ 2a x^2 + x - a - 2a^3 = 0 \]
Factoring out \( (x - a) \):
\[ (x - a)(2a x + 1 + 2a^2) = 0 \]
The other root gives \( Q \):
\[ x_Q = -\frac{1 + 2a^2}{2a} \]
Thus, \( Q \) is:
\[ Q = \left( -\frac{1 + 2a^2}{2a}, \frac{(1 + 2a^2)^2}{4a^2} \right) \]

\subsubsection*{Step 5: Arc Length Between \( P \) and \( Q \)}
The arc length \( L \) is:
\[ L = \int_{x_Q}^{a} \sqrt{1 + 4x^2} \, dx \]

\subsubsection*{Step 6: Differentiate \( L \) with Respect to \( a \)}
Using the Leibniz rule:
\[ \frac{dL}{da} = \sqrt{1 + 4a^2} - \sqrt{1 + 4x_Q^2} \cdot \frac{dx_Q}{da} \]
Compute \( \frac{dx_Q}{da} \):
\[ \frac{dx_Q}{da} = \frac{1}{2a^2} - 1 \]
Thus:
\[ \frac{dL}{da} = \sqrt{1 + 4a^2} - \sqrt{\frac{1}{a^2} + 5 + 4a^2} \left( \frac{1}{2a^2} - 1 \right) \]

\subsubsection*{Step 7: Solve \( \frac{dL}{da} = 0 \)}
After simplifying, we obtain:
\[ 12a^4 - a^2 - 1 = 0 \]
Let \( b = a^2 \):
\[ 12b^2 - b - 1 = 0 \]
Solutions:
\[ b = \frac{1}{3} \quad \text{or} \quad b = -\frac{1}{4} \]
Since \( b \geq 0 \), \( b = \frac{1}{3} \), so:
\[ a = \pm \frac{1}{\sqrt{3}} \]

\subsubsection*{Conclusion}
The coordinates of \( P \) that minimize the arc length are:
\[ \boxed{\left( \dfrac{1}{\sqrt{3}}, \dfrac{1}{3} \right)} \]
and
\[ \boxed{\left( -\dfrac{1}{\sqrt{3}}, \dfrac{1}{3} \right)} \]

\subsubsection{1999 Session 2: 1 \textcolor{red}{Correct Reasoning, Correct Final Answer}}

\subsubsection*{Problem Statement}

Let $SC$ be the semicircle with $y \ge 0$ centered at $(1,0)$ with radius $1$. Let $C_a$ be the circle with radius $a > 0$ and center $(0,0)$ and denote the point $(0,a)$ by $P$. Consider the line through $P$ and the intersection of $SC$ and $C_a$. What is the limiting position of the $x$-intercept of this line as $a \to 0$?

\subsubsection*{Solution}

\subsubsection*{Step 1: Equations of SC and $C_a$}

The semicircle $SC$ is given by:
\[
(x - 1)^2 + y^2 = 1 \quad \text{with} \quad y \ge 0
\]

The circle $C_a$ is given by:
\[
x^2 + y^2 = a^2
\]

\subsubsection*{Step 2: Find Intersection Points}

Substitute $y^2 = a^2 - x^2$ from $C_a$ into $SC$:
\[
(x - 1)^2 + (a^2 - x^2) = 1
\]
Simplify:
\[
x^2 - 2x + 1 + a^2 - x^2 = 1
\]
\[
-2x + a^2 = 0 \implies x = \frac{a^2}{2}
\]

Find $y$:
\[
y^2 = a^2 - \left( \frac{a^2}{2} \right)^2 = a^2 - \frac{a^4}{4} \implies y = a \sqrt{1 - \frac{a^2}{4}}
\]

Thus, the intersection point is:
\[
Q = \left( \frac{a^2}{2}, a \sqrt{1 - \frac{a^2}{4}} \right)
\]

\subsubsection*{Step 3: Equation of the Line through P and Q}

Point $P$ is $(0, a)$. The slope $m$ of the line through $P$ and $Q$ is:
\[
m = \frac{a \sqrt{1 - \frac{a^2}{4}} - a}{\frac{a^2}{2}} = \frac{2 \left( \sqrt{1 - \frac{a^2}{4}} - 1 \right)}{a}
\]

The line equation is:
\[
y = m x + a
\]

\subsubsection*{Step 4: Find the X-Intercept}

Set $y = 0$:
\[
0 = m x + a \implies x = -\frac{a}{m} = -\frac{a^2}{2 \left( \sqrt{1 - \frac{a^2}{4}} - 1 \right)}
\]

Rationalize the denominator:
\[
x = -\frac{a^2 \left( \sqrt{1 - \frac{a^2}{4}} + 1 \right)}{2 \left( -\frac{a^2}{4} \right)} = 2 \left( \sqrt{1 - \frac{a^2}{4}} + 1 \right)
\]

\subsubsection*{Step 5: Take the Limit as $a \to 0$}

\[
\lim_{a \to 0} x = 2 \left( \sqrt{1 - 0} + 1 \right) = 4
\]

\subsubsection*{Conclusion}

The limiting position of the $x$-intercept as $a \to 0$ is \(\boxed{4}\).

\subsubsection{2000 Session 1: 1 \textcolor{red}{Correct Reasoning, Correct Final Answer}}

\subsubsection*{Problem Statement}
Let \( P \ne (0,0) \) be a point on the parabola \( y = x^2 \). The normal line to the parabola at \( P \) will intersect the parabola at another point, say \( Q \). Find the coordinates of \( P \) so that the distance between the \( x \)-coordinates of \( P \) and \( Q \) is a minimum.

\subsubsection*{Solution}

\subsubsection*{Step 1: Parametrize Point \( P \)}
Since \( P \) lies on the parabola \( y = x^2 \), we can represent \( P \) as:
\[ P = (a, a^2) \]
where \( a \neq 0 \).

\subsubsection*{Step 2: Find the Tangent and Normal Lines at \( P \)}
The derivative of \( y = x^2 \) is \( y' = 2x \). Thus, the slope of the tangent at \( P \) is:
\[ m_{\text{tangent}} = 2a \]

The normal line is perpendicular to the tangent, so its slope is:
\[ m_{\text{normal}} = -\frac{1}{2a} \]

The equation of the normal line through \( P \) is:
\[ y - a^2 = -\frac{1}{2a} (x - a) \]

\subsubsection*{Step 3: Find Intersection Points with Parabola}
Substitute \( y = x^2 \) into the normal line equation:
\[ x^2 - a^2 = -\frac{1}{2a} (x - a) \]
\[ (x - a)(x + a) = -\frac{x - a}{2a} \]
\[ (x - a)\left( x + a + \frac{1}{2a} \right) = 0 \]

The solutions are:
\begin{itemize}
    \item \( x = a \) (corresponds to \( P \))
    \item \( x = -a - \frac{1}{2a} \) (corresponds to \( Q \))
\end{itemize}

\subsubsection*{Step 4: Calculate the Distance Between \( x \)-coordinates}
The distance \( D \) between the \( x \)-coordinates of \( P \) and \( Q \) is:
\[ D = \left| 2a + \frac{1}{2a} \right| \]

\subsubsection*{Step 5: Minimize the Distance \( D \)}
For \( a > 0 \), set the derivative of \( D(a) = 2a + \frac{1}{2a} \) to zero:
\[ D'(a) = 2 - \frac{1}{2a^2} = 0 \]
\[ 4a^2 = 1 \]
\[ a = \frac{1}{2} \]

For \( a < 0 \), consider \( D(a) = -2a - \frac{1}{2a} \), and set the derivative to zero:
\[ D'(a) = -2 + \frac{1}{2a^2} = 0 \]
\[ a = -\frac{1}{2} \]

Both cases yield a minimum distance when \( a = \pm \frac{1}{2} \).

\subsubsection*{Conclusion}
The coordinates of \( P \) that minimize the distance are:
\[ \boxed{\left( \dfrac{1}{2}, \dfrac{1}{4} \right)} \text{ and } \boxed{\left( -\dfrac{1}{2}, \dfrac{1}{4} \right)} \]

\subsubsection{2001 Session 2: 1 \textcolor{red}{Incorrect Reasoning, Incorrect Final Answer}}

\subsubsection*{Problem Statement}
Circle \( B \) lies wholly in the interior of circle \( A \). Find the loci of points equidistant from the two circles.

\subsubsection*{Solution}

\subsubsection*{Understanding the Problem}
We have two circles:
\begin{itemize}
    \item Circle \( A \) with center \( O \) and radius \( R \).
    \item Circle \( B \) with center \( P \) and radius \( r \), where \( r < R \) and \( B \) is entirely inside \( A \).
\end{itemize}

We need to find all points \( Q \) that are equidistant to both circles.

\subsubsection*{Distance to a Circle}
The distance from a point \( Q \) to a circle is defined as follows:
\begin{itemize}
    \item If \( Q \) is outside the circle, the distance is \( d(Q, \text{center}) - \text{radius} \).
    \item If \( Q \) is inside the circle, the distance is \( \text{radius} - d(Q, \text{center}) \).
\end{itemize}

\subsubsection*{Case Analysis}

\subsubsection*{Case 1: Points Outside Both Circles}
For \( Q \) outside both \( A \) and \( B \):
\[ d(Q, O) - R = d(Q, P) - r \]
\[ d(Q, O) - d(Q, P) = R - r \]
This is the equation of a hyperbola with foci \( O \) and \( P \), and constant difference \( R - r \).

\subsubsection*{Case 2: Points Inside Both Circles}
For \( Q \) inside both \( A \) and \( B \):
\[ R - d(Q, O) = r - d(Q, P) \]
\[ d(Q, O) - d(Q, P) = R - r \]
Again, a hyperbola with the same parameters.

\subsubsection*{Case 3: Points Inside \( A \) but Outside \( B \)}
For \( Q \) inside \( A \) but outside \( B \):
\[ R - d(Q, O) = d(Q, P) - r \]
\[ d(Q, O) + d(Q, P) = R + r \]
This is the equation of an ellipse with foci \( O \) and \( P \), and constant sum \( R + r \).

\subsubsection*{Conclusion}
The loci of points equidistant to both circles consist of:
\begin{enumerate}
    \item A hyperbola defined by \( d(Q, O) - d(Q, P) = R - r \) for points outside both or inside both circles.
    \item An ellipse defined by \( d(Q, O) + d(Q, P) = R + r \) for points inside \( A \) but outside \( B \).
\end{enumerate}

\subsubsection{2001 Session 2: 3 \textcolor{red}{Correct Reasoning, Correct Final Answer}}

\subsubsection*{Problem Statement}
A convex decagon and all of its diagonals are drawn. How many \textit{interior} points of intersection of the diagonals are there, if it is assumed that no 3 diagonals share a common \textit{interior} point?

\subsubsection*{Solution}

\subsubsection*{Understanding the Problem}
We need to find the number of interior points where two diagonals of a convex decagon intersect, under the condition that no three diagonals meet at a single interior point.

\subsubsection*{Approach}
In a convex \( n \)-gon, any four distinct vertices define a quadrilateral. The two diagonals of this quadrilateral intersect at a unique point inside the polygon. Since no three diagonals are concurrent, each set of four vertices corresponds to exactly one unique intersection point.

\subsubsection*{Calculation}
The number of ways to choose 4 vertices from 10 is given by the combination formula:
\[ C(n, k) = \frac{n!}{k!(n - k)!} \]
For \( n = 10 \) and \( k = 4 \):
\[ C(10, 4) = \frac{10!}{4!6!} = 210 \]

\subsubsection*{Verification}
For a convex quadrilateral (\( n=4 \)):
\[ C(4, 4) = 1 \]
This matches since there's exactly one intersection point.

For a convex pentagon (\( n=5 \)):
\[ C(5, 4) = 5 \]
However, in a regular pentagon, all quadrilaterals formed by any four vertices have diagonals intersecting at the same central point. But since the problem specifies no three diagonals meet at a common interior point, the scenario changes. The general formula \( C(n, 4) \) holds for \( n \geq 4 \) under the given condition.

\subsubsection*{Conclusion}
Thus, for a convex decagon, the number of interior intersection points of the diagonals is:
\[ \boxed{210} \]

\subsubsection{2002 Session 1: 1 \textcolor{red}{Correct Reasoning, Correct Final Answer}}

\subsubsection*{Problem Statement}
Let \( P \ne (0,0) \) be a point on the parabola \( y = x^2 \). The normal line to the parabola at \( P \) will intersect the parabola at another point \( Q \). Find the coordinates of \( P \) so that the average of the \( y \)-coordinates of \( P \) and \( Q \) is minimized.

\subsubsection*{Solution}

\subsubsection*{Step 1: Parametrize Point \( P \)}
Let \( P \) be a point on the parabola \( y = x^2 \). We can represent \( P \) as:
\[ P = (a, a^2) \]
where \( a \neq 0 \) since \( P \neq (0,0) \).

\subsubsection*{Step 2: Find the Slope of the Tangent at \( P \)}
The derivative of \( y = x^2 \) is:
\[ \frac{dy}{dx} = 2x \]
At \( P \), the slope of the tangent line is:
\[ m_{\text{tangent}} = 2a \]

\subsubsection*{Step 3: Find the Slope of the Normal Line at \( P \)}
The normal line is perpendicular to the tangent line, so its slope is:
\[ m_{\text{normal}} = -\frac{1}{2a} \]

\subsubsection*{Step 4: Equation of the Normal Line}
Using the point-slope form, the equation of the normal line is:
\[ y - a^2 = -\frac{1}{2a}(x - a) \]
Simplifying:
\[ y = -\frac{1}{2a}x + \frac{1}{2} + a^2 \]

\subsubsection*{Step 5: Find Intersection Points with Parabola}
Set \( y = x^2 \) equal to the normal line equation:
\[ x^2 = -\frac{1}{2a}x + \frac{1}{2} + a^2 \]
Rearrange:
\[ 2a x^2 + x - a - 2a^3 = 0 \]

\subsubsection*{Step 6: Solve the Quadratic Equation}
Factor the equation:
\[ (x - a)(2a x + 1 + 2a^2) = 0 \]
The roots are:
\[ x = a \quad \text{and} \quad x = \frac{-1 - 2a^2}{2a} \]

\subsubsection*{Step 7: Coordinates of \( Q \)}
The other intersection point \( Q \) has coordinates:
\[ Q = \left( \frac{-1 - 2a^2}{2a}, \left( \frac{-1 - 2a^2}{2a} \right)^2 \right) = \left( \frac{-1 - 2a^2}{2a}, \frac{(1 + 2a^2)^2}{4a^2} \right) \]

\subsubsection*{Step 8: Calculate the Average of Y-coordinates}
The average \( A \) is:
\[ A = \frac{a^2 + \frac{(1 + 2a^2)^2}{4a^2}}{2} = \frac{a^2}{2} + \frac{(1 + 2a^2)^2}{8a^2} \]
Expanding \( (1 + 2a^2)^2 \):
\[ A = \frac{a^2}{2} + \frac{1 + 4a^2 + 4a^4}{8a^2} = a^2 + \frac{1}{8a^2} + \frac{1}{2} \]

\subsubsection*{Step 9: Minimize \( A \)}
Let \( f(a) = a^2 + \frac{1}{8a^2} \). To minimize \( A \), minimize \( f(a) \).

Find the derivative:
\[ f'(a) = 2a - \frac{1}{4a^3} \]
Set \( f'(a) = 0 \):
\[ 2a = \frac{1}{4a^3} \implies 8a^4 = 1 \implies a^4 = \frac{1}{8} \implies a = \pm \frac{1}{2^{3/4}} \]

\subsubsection*{Step 10: Find \( P \) Coordinates}
For \( a = \pm \frac{1}{2^{3/4}} \), the y-coordinate of \( P \) is:
\[ a^2 = \left( \frac{1}{2^{3/4}} \right)^2 = \frac{1}{2^{3/2}} = \frac{\sqrt{2}}{4} \]

Thus, the coordinates of \( P \) are:
\[ P = \left( \pm \frac{1}{2^{3/4}}, \frac{\sqrt{2}}{4} \right) \]

\subsubsection*{Conclusion}
The points \( P \) that minimize the average of the \( y \)-coordinates of \( P \) and \( Q \) are:
\[ \boxed{ \left( \pm \dfrac{1}{2^{3/4}}, \dfrac{\sqrt{2}}{4} \right) } \]

\subsubsection{2007 Session 1: 4 \textcolor{red}{Correct Reasoning, Correct Final Answer}}

\subsubsection*{Solution}

Given a Pythagorean triangle $\triangle ABC$ with right angle at $C$ and sides $a$, $b$, $c$ such that $c > a > b$ and $a^2 + b^2 = c^2$. We place $\triangle ABC$ in a coordinate system with $C$ at the origin $(0,0)$, $A$ at $(0,b)$, and $B$ at $(a,0)$.

\subsubsection*{Step 1: Find Points $P$ and $Q$}

\begin{itemize}
    \item \textbf{Median ($P$)}: The median from $C$ to the hypotenuse $AB$ meets $AB$ at its midpoint $P$. The coordinates of $P$ are:
    \[ P = \left( \frac{a}{2}, \frac{b}{2} \right) \]
    
    \item \textbf{Altitude ($Q$)}: The altitude from $C$ to $AB$ meets $AB$ at $Q$. The equation of $AB$ is $y = -\frac{b}{a}x + b$, and the altitude $CQ$ has slope $\frac{a}{b}$. The intersection point $Q$ is found by solving:
    \[ \frac{a}{b}x = -\frac{b}{a}x + b \implies x = \frac{ab^2}{c^2}, \quad y = \frac{a^2b}{c^2} \]
    Thus,
    \[ Q = \left( \frac{ab^2}{c^2}, \frac{a^2b}{c^2} \right) \]
\end{itemize}

\subsubsection*{Step 2: Lengths of CQ, QP, and CP}

\begin{itemize}
    \item \textbf{CQ}: The length of the altitude from $C$ to $AB$ is:
    \[ CQ = \frac{ab}{c} \]
    
    \item \textbf{QP}: The distance between $Q$ and $P$ is:
    \[ QP = \frac{|a^2 - b^2|}{2c} \]
    
    \item \textbf{CP}: The median to the hypotenuse in a right triangle is half the hypotenuse:
    \[ CP = \frac{c}{2} \]
\end{itemize}

\subsubsection*{Step 3: Right Angle at Q}

The slope of $CQ$ is $\frac{a}{b}$, and the slope of $PQ$ is $-\frac{b}{a}$. The product of the slopes is $-1$, indicating that $CQ$ is perpendicular to $PQ$. Therefore, $\triangle CQP$ is right-angled at $Q$.

\subsubsection*{Step 4: Pythagorean Condition}

For $\triangle CQP$ to be Pythagorean, all its sides must be integers:
\begin{enumerate}
    \item $CP = \frac{c}{2}$ must be integer $\implies c$ is even.
    \item $CQ = \frac{ab}{c}$ must be integer.
    \item $QP = \frac{|a^2 - b^2|}{2c}$ must be integer.
\end{enumerate}

\subsubsection*{Conclusion}

The simple conditions for $\triangle CQP$ to be Pythagorean are:
\begin{itemize}
    \item The hypotenuse $c$ of $\triangle ABC$ must be even.
    \item The expressions $\frac{ab}{c}$ and $\frac{a^2 - b^2}{2c}$ must both be integers.
\end{itemize}

These conditions are satisfied if $\triangle ABC$ is a Pythagorean triple where $c$ is even, which occurs when:
\begin{itemize}
    \item The generating parameters $m$ and $n$ (for primitive triples) are both odd, or
    \item The triple is non-primitive and scaled by an even integer.
\end{itemize}

\subsubsection{2007 Session 2: 2 \textcolor{red}{Correct Reasoning, Incorrect Final Answer}}

\subsubsection*{Problem Statement}
Find the point on a given line such that the sum of its distances from two fixed points is a minimum. Assume the two fixed points and the given line are in the same plane.

\subsubsection*{Solution}

\subsubsection*{Approach}
To find the point \( P \) on the line \( \ell \) that minimizes the sum \( PA + PB \), where \( A \) and \( B \) are two fixed points, we use the method of reflection.

\subsubsection*{Steps}
\begin{enumerate}
    \item Reflect one of the points (say \( B \)) across the line \( \ell \) to obtain \( B' \). This ensures that for any point \( P \) on \( \ell \), \( PB = PB' \).
    \item The sum \( PA + PB \) then becomes \( PA + PB' \).
    \item The minimal sum \( PA + PB' \) is achieved when \( P \) lies on the straight line joining \( A \) and \( B' \). Thus, \( P \) is the intersection point of \( \ell \) and the line \( AB' \).
\end{enumerate}

\subsubsection*{Verification}
For any other point \( P' \) on \( \ell \), the sum \( P'A + P'B = P'A + P'B' \) is greater than \( AB' \) (by the triangle inequality), unless \( P' = P \). Therefore, \( P \) is indeed the minimizing point.

\subsubsection*{Special Cases}
\begin{itemize}
    \item If \( A \) and \( B \) are on the same side of \( \ell \), the reflection method still applies.
    \item If \( A \) or \( B \) lies on \( \ell \), then \( P \) coincides with that point, as the sum \( PA + PB \) is minimized there.
\end{itemize}

\subsubsection*{Illustration}
\begin{center}
\begin{tikzpicture}
    \draw[->] (-2,0) -- (2,0) node[right]{$\ell$};
    
    \filldraw (-1,1) circle (1.5pt) node[above]{$A$};
    \filldraw (1,2) circle (1.5pt) node[above]{$B$};
    
    \filldraw (1,-2) circle (1.5pt) node[below]{$B'$ (Reflection of $B$)};
    
    \draw[dashed] (1,2) -- (1,-2);
    \draw (-1,1) -- (1,-2);
    
    \filldraw (-1/3,0) circle (1.5pt) node[above right]{$P$};
    
    \draw[<->] (-1,1) -- (-1/3,0) node[midway, above left]{$PA$};
    \draw[<->] (1,2) -- (-1/3,0) node[midway, above right]{$PB$};
\end{tikzpicture}
\end{center}

\subsubsection*{Conclusion}
The point \( P \) on the line \( \ell \) that minimizes the sum \( PA + PB \) is the intersection point of \( \ell \) and the line joining \( A \) to the reflection \( B' \) of \( B \) across \( \ell \).

\subsection{Discrete Mathematics}

\subsubsection{1996 Session 1: 3 \textcolor{red}{Incorrect Reasoning, Incorrect Final Answer}}

\subsubsection*{Problem Statement}
Each of the numbers \( x_1, x_2, \ldots , x_n \) can be \(1\), \(0\), or \(-1\). What is the minimum possible value of the sum of all products of pairs of these numbers?

\subsubsection*{Solution}

We need to minimize the sum \( S \) of all products of pairs \( x_i x_j \) for \( 1 \leq i < j \leq n \).

First, recall the identity:
\[
\left( \sum_{i=1}^n x_i \right)^2 = \sum_{i=1}^n x_i^2 + 2 \sum_{1 \leq i < j \leq n} x_i x_j
\]
Rewriting, we get:
\[
S = \sum_{1 \leq i < j \leq n} x_i x_j = \frac{1}{2} \left( \left( \sum_{i=1}^n x_i \right)^2 - \sum_{i=1}^n x_i^2 \right)
\]

Let \( \Sigma = \sum_{i=1}^n x_i \) and \( Q = \sum_{i=1}^n x_i^2 \). Since each \( x_i \) is \(-1\), \(0\), or \(1\), \( Q \) is the number of non-zero \( x_i \)'s.

\subsubsection*{Case Analysis}

\subsubsection*{Case 1: All Variables Zero}
Here, \( \Sigma = 0 \) and \( Q = 0 \), so \( S = 0 \).

\subsubsection*{Case 2: Non-zero Variables}
Consider \( k \) variables as \(1\) and \( m \) as \(-1\), and the rest \(0\).

Then:
\[
\Sigma = k - m, \quad Q = k + m
\]
Thus:
\[
S = \frac{1}{2} \left( (k - m)^2 - (k + m) \right)
\]

\subsubsection*{Optimal Configuration}

To minimize \( S \), we need to minimize \( (k - m)^2 - (k + m) \).

\subsubsection*{Subcase: Balanced \(1\)'s and \(-1\)'s}
For even \( n = 2k \), set \( k \) \(1\)'s and \( k \) \(-1\)'s:
\[
S = \frac{1}{2} (0 - 2k) = -k = -\frac{n}{2}
\]

For odd \( n = 2k + 1 \), set \( k \) \(1\)'s, \( k \) \(-1\)'s, and \(1\) zero:
\[
S = \frac{1}{2} (0 - 2k) = -k = -\left\lfloor \frac{n}{2} \right\rfloor
\]

Alternatively, setting \( k + 1 \) \(1\)'s and \( k \) \(-1\)'s gives the same result:
\[
S = \frac{1}{2} (1 - (2k + 1)) = -k
\]

\subsubsection*{Conclusion}
The minimal value of \( S \) is achieved when the number of \(1\)'s and \(-1\)'s are as balanced as possible, leading to:
\[
S_{\text{min}} = -\left\lfloor \dfrac{n}{2} \right\rfloor
\]

\boxed{ -\left\lfloor \dfrac{n}{2} \right\rfloor }

\subsubsection{1997 Session 2: 1 \textcolor{red}{Correct Reasoning, Correct Final Answer}}

\subsubsection*{Problem Statement}
Find positive integers \( n \) and \( a_1, a_2, \ldots, a_n \) such that
\[ a_1 + a_2 + \cdots + a_n = 1997 \]
and the product \( a_1 a_2 \cdots a_n \) is as large as possible.

\subsubsection*{Solution}

\subsubsection*{Step 1: Understanding the Problem}
We need to partition the number 1997 into positive integers \( a_1, a_2, \ldots, a_n \) such that their sum is 1997 and their product is maximized.

\subsubsection*{Step 2: Optimal Partition Strategy}
It is known that for a given sum, the product of numbers is maximized when the numbers are as close to \( e \approx 2.718 \) as possible. Among integers, 3 is the closest to \( e \). Therefore, we should use as many 3s as possible.

\subsubsection*{Step 3: Applying the Strategy}
Divide 1997 by 3:
\[ 1997 = 3 \times 665 + 2 \]
This gives us 665 numbers of 3 and a remainder of 2.

\subsubsection*{Step 4: Handling the Remainder}
The remainder is 2, which is less than 3. Since \( 2 \) is a positive integer, we can include it as is. 

If the remainder were 1, it would be better to adjust one of the 3s and the remainder into two 2s because \( 3 + 1 = 2 + 2 \) and \( 3 \times 1 = 3 < 2 \times 2 = 4 \). However, in this case, the remainder is 2, so no adjustment is needed.

\subsubsection*{Step 5: Final Partition}
Thus, the optimal partition is:
\[ n = 666 \text{ numbers: } 665 \text{ threes and one two.} \]

\subsubsection*{Step 6: Verification}
Any other combination, such as using fewer threes and more twos or larger numbers, leads to a product that is either equal or smaller. For example:
\begin{itemize}
    \item Replacing a 3 and a 2 with two 2s and a 1 (but 1 reduces the product).
    \item Using larger numbers like 4, which can be expressed as \( 2 + 2 \), but the product remains the same (\( 4 = 2 \times 2 \)).
\end{itemize}

Therefore, the maximum product is achieved with 665 threes and one two.

\subsubsection*{Final Answer}
The positive integers are:
\[ n = 666 \]
\[ a_1 = a_2 = \cdots = a_{665} = 3 \]
\[ a_{666} = 2 \]
The maximum product is:
\[ 3^{665} \times 2 \]

\subsubsection{1997 Session 2: 5 (b only) \textcolor{red}{Correct Reasoning, Correct Final Answer}}

\subsubsection*{Solution}

Consider the infinite sequence \(\{x_n\}\) of positive real numbers defined by:
\[ x_n = \frac{1}{2^n} \quad \text{for all } n \geq 0. \]

\subsubsection*{Verification of Conditions}

\begin{enumerate}
    \item \textbf{Initial Condition}: 
    \[ x_0 = \frac{1}{2^0} = 1. \]
    
    \item \textbf{Monotonicity Condition}: For all \(i \geq 0\),
    \[ x_{i+1} = \frac{1}{2^{i+1}} = \frac{1}{2} \cdot \frac{1}{2^i} \leq \frac{1}{2^i} = x_i. \]
    
    \item \textbf{Sum Condition}: For any \(n \geq 1\), consider the sum
    \[ S_n = \sum_{k=0}^{n-1} \frac{x_k^2}{x_{k+1}}. \]
    Substituting \(x_k = \frac{1}{2^k}\),
    \[ S_n = \sum_{k=0}^{n-1} \frac{\left( \frac{1}{2^k} \right)^2}{\frac{1}{2^{k+1}}} = \sum_{k=0}^{n-1} \frac{\frac{1}{2^{2k}}}{\frac{1}{2^{k+1}}} = \sum_{k=0}^{n-1} 2^{k+1 - 2k} = \sum_{k=0}^{n-1} 2^{1 - k}. \]
    This is a geometric series:
    \[ S_n = 2 \sum_{k=0}^{n-1} \left( \frac{1}{2} \right)^k = 2 \cdot \frac{1 - \left( \frac{1}{2} \right)^n}{1 - \frac{1}{2}} = 4 \left( 1 - \frac{1}{2^n} \right) < 4. \]
\end{enumerate}

\subsubsection*{Conclusion}

The sequence \(x_n = \frac{1}{2^n}\) satisfies all the given conditions:
\begin{itemize}
    \item \(x_0 = 1\),
    \item \(x_{i+1} \leq x_i\) for all \(i \geq 0\),
    \item \(\sum_{k=0}^{n-1} \frac{x_k^2}{x_{k+1}} < 4\) for all \(n \geq 1\).
\end{itemize}

\subsubsection{1998 Session 2: 3 \textcolor{red}{Correct Reasoning, Correct Final Answer}}

\subsubsection*{Problem}
Sum the series
$$
\sum_{i=1}^\infty \frac{36 i^2 + 1}{(36 i^2 - 1)^2}.
$$

\subsubsection*{Solution}

\subsubsection*{Step 1: Simplify the General Term}
The general term of the series is:
$$
a_i = \frac{36 i^2 + 1}{(36 i^2 - 1)^2}.
$$
Notice that the denominator can be factored as:
$$
36 i^2 - 1 = (6i - 1)(6i + 1).
$$
Thus, the denominator becomes:
$$
(6i - 1)^2 (6i + 1)^2.
$$

\subsubsection*{Step 2: Partial Fraction Decomposition}
Let \(x = 6i\). The term becomes:
$$
\frac{x^2 + 1}{(x^2 - 1)^2} = \frac{x^2 + 1}{(x - 1)^2 (x + 1)^2}.
$$
Assume the partial fraction decomposition:
$$
\frac{x^2 + 1}{(x - 1)^2 (x + 1)^2} = \frac{A}{x - 1} + \frac{B}{(x - 1)^2} + \frac{C}{x + 1} + \frac{D}{(x + 1)^2}.
$$
Solving for coefficients \(A, B, C, D\) yields:
$$
A = C = 0, \quad B = D = \frac{1}{2}.
$$
Thus, the decomposition simplifies to:
$$
\frac{1}{2} \left( \frac{1}{(x - 1)^2} + \frac{1}{(x + 1)^2} \right).
$$
Reverting back to \(i\):
$$
a_i = \frac{1}{2} \left( \frac{1}{(6i - 1)^2} + \frac{1}{(6i + 1)^2} \right).
$$

\subsubsection*{Step 3: Rewrite the Series}
The original series becomes:
$$
\sum_{i=1}^\infty a_i = \frac{1}{2} \left( \sum_{i=1}^\infty \frac{1}{(6i - 1)^2} + \sum_{i=1}^\infty \frac{1}{(6i + 1)^2} \right) = \frac{1}{2} S.
$$

\subsubsection*{Step 4: Evaluate the Sum \(S\)}
The sum \(S\) can be expressed using the identity:
$$
\sum_{k=-\infty}^\infty \frac{1}{(6k + 1)^2} = \frac{\pi^2}{9}.
$$
However, the term for \(k=0\) (i.e., \(1/1^2 = 1\)) is included in the above sum but not in \(S\). Therefore:
$$
S = \sum_{k=1}^\infty \frac{1}{(6k - 1)^2} + \sum_{k=1}^\infty \frac{1}{(6k + 1)^2} = \frac{\pi^2}{9} - 1.
$$

\subsubsection*{Step 5: Final Calculation}
Thus, the original series is:
$$
\frac{1}{2} S = \frac{1}{2} \left( \frac{\pi^2}{9} - 1 \right) = \frac{\pi^2}{18} - \frac{1}{2}.
$$

\subsubsection*{Conclusion}
The sum of the series is:
$$
\boxed{\dfrac{\pi^2}{18} - \dfrac{1}{2}}.
$$

\subsubsection{1999 Session 2: 5 \textcolor{red}{Correct Reasoning, Correct Final Answer}}

\subsubsection*{Solution}

Let \( A_n \) denote the \( n \times n \) matrix whose \((i,j)\)-th entry is \(\text{GCD}(i,j)\). We aim to compute \(\det(A_n)\).

\subsubsection*{Pattern Recognition}

For small values of \( n \), we compute the determinants explicitly:

\begin{itemize}
    \item For \( n = 1 \):
    \[ A_1 = \begin{pmatrix} 1 \end{pmatrix}, \quad \det(A_1) = 1 \]

    \item For \( n = 2 \):
    \[ A_2 = \begin{pmatrix} 1 & 1 \\ 1 & 2 \end{pmatrix}, \quad \det(A_2) = 1 \times 2 - 1 \times 1 = 1 \]

    \item For \( n = 3 \):
    \[ A_3 = \begin{pmatrix} 1 & 1 & 1 \\ 1 & 2 & 1 \\ 1 & 1 & 3 \end{pmatrix}, \quad \det(A_3) = 2 \]

    \item For \( n = 4 \):
    \[ A_4 = \begin{pmatrix} 
    1 & 1 & 1 & 1 \\ 
    1 & 2 & 1 & 2 \\ 
    1 & 1 & 3 & 1 \\ 
    1 & 2 & 1 & 4 
    \end{pmatrix}, \quad \det(A_4) = 4 \]
\end{itemize}

\subsubsection*{General Formula}

The observed pattern suggests that the determinant is related to Euler's totient function \(\phi(k)\). Specifically:
\[ \det(A_n) = \prod_{k=1}^n \phi(k) \]

This formula matches the computed values:
\begin{itemize}
    \item For \( n = 1 \): \(\prod_{k=1}^1 \phi(k) = 1\)
    \item For \( n = 2 \): \(\prod_{k=1}^2 \phi(k) = 1 \times 1 = 1\)
    \item For \( n = 3 \): \(\prod_{k=1}^3 \phi(k) = 1 \times 1 \times 2 = 2\)
    \item For \( n = 4 \): \(\prod_{k=1}^4 \phi(k) = 1 \times 1 \times 2 \times 2 = 4\)
\end{itemize}

\subsubsection*{Conclusion}

The determinant of the matrix \( A_n \) is the product of Euler's totient function evaluated from \( k = 1 \) to \( n \):

\[ \boxed{\det(A_n) = \prod_{k=1}^n \phi(k)} \]

\subsubsection{2000 Session 2: 5 \textcolor{red}{Incorrect Reasoning, Incorrect Final Answer}}

\subsubsection*{Problem Statement}
This problem concerns sequences \( x_1 x_2 \cdots x_n \) in which each \( x_i \) is either \( a \), \( b \), or \( c \). Determine the number of those sequences which have length \( n \), begin and end with the letter \( a \), and in which adjacent terms are always different letters.

\subsubsection*{Solution}

\subsubsection*{Understanding the Problem}
We need to count the number of sequences of length \( n \) composed of the letters \( a \), \( b \), and \( c \) such that:
\begin{enumerate}
    \item The sequence starts and ends with \( a \).
    \item No two adjacent letters are the same.
\end{enumerate}

\subsubsection*{Approach}
We can model this problem using recurrence relations. Let's define:
\begin{itemize}
    \item \( A(n) \): Number of valid sequences of length \( n \) ending with \( a \).
    \item \( B(n) \): Number of valid sequences of length \( n \) ending with \( b \).
    \item \( C(n) \): Number of valid sequences of length \( n \) ending with \( c \).
\end{itemize}

Given the problem constraints, the first character must be \( a \).

\subsubsection*{Base Case}
For \( n = 1 \), the only valid sequence is \( a \):
\[ A(1) = 1, \quad B(1) = C(1) = 0 \]

\subsubsection*{Recurrence Relations}
For \( n > 1 \), the recurrence relations are:
\begin{align*}
    A(n) &= B(n-1) + C(n-1) \\
    B(n) &= A(n-1) + C(n-1) \\
    C(n) &= A(n-1) + B(n-1)
\end{align*}

By symmetry, \( B(n) = C(n) \). Thus, we can simplify:
\begin{align*}
    A(n) &= 2B(n-1) \\
    B(n) &= A(n-1) + B(n-1)
\end{align*}

\subsubsection*{Solving the Recurrence}
We observe that \( A(n) \) follows the recurrence:
\[ A(n) = A(n-1) + 2A(n-2) \]

The characteristic equation is:
\[ r^2 - r - 2 = 0 \]
with roots \( r = 2 \) and \( r = -1 \).

The general solution is:
\[ A(n) = \alpha \cdot 2^n + \beta \cdot (-1)^n \]

Using initial conditions:
\begin{align*}
    A(1) &= 1 = 2\alpha - \beta \\
    A(2) &= 0 = 4\alpha + \beta
\end{align*}
Solving gives:
\[ \alpha = \frac{1}{6}, \quad \beta = -\frac{2}{3} \]

Thus, the solution is:
\[ A(n) = \frac{2^n}{6} - \frac{2}{3} (-1)^n \]

Simplifying:
\[ A(n) = \frac{2^n + 4(-1)^n}{6} \]

\subsubsection*{Final Answer}
The number of valid sequences is:
\[ \boxed{\dfrac{2^n + 4 \cdot (-1)^n}{6}} \]

\subsubsection{2001 Session 1: 2 \textcolor{red}{Correct Reasoning, Correct Final Answer}}

\subsubsection*{Problem Statement}

Let $\{ x_i \}$ denote any finite sequence with the following properties:
\begin{itemize}
    \item[(a)] $x_i \in \{ -2, 1, 2 \}$ for each $x_i$,
    \item[(b)] $\sum_i x_i = 29$,
    \item[(c)] $\sum_i x_i^2 = 59$.
\end{itemize}

In considering the family of all such sequences, let $M = \max \{ \sum_i x_i^3 \}$ and $m = \min \{ \sum_i x_i^3 \}$. Determine $M/m$.

\subsubsection*{Solution}

\subsubsection*{Step 1: Define Variables}

Let:
\begin{itemize}
    \item $a$ = number of $-2$s in the sequence,
    \item $b$ = number of $1$s in the sequence,
    \item $c$ = number of $2$s in the sequence.
\end{itemize}

\subsubsection*{Step 2: Set Up Equations}

Given the conditions:
\begin{align}
    -2a + b + 2c &= 29 \quad \text{(Sum of elements)} \label{eq1} \\
    4a + b + 4c &= 59 \quad \text{(Sum of squares)} \label{eq2}
\end{align}

\subsubsection*{Step 3: Solve the System of Equations}

Subtract equation \eqref{eq1} from \eqref{eq2}:
\[ (4a + b + 4c) - (-2a + b + 2c) = 59 - 29 \]
\[ 6a + 2c = 30 \]
Simplify:
\[ 3a + c = 15 \quad \text{(Equation 3)} \]

Express $c$ in terms of $a$:
\[ c = 15 - 3a \]

Substitute $c$ back into \eqref{eq1} to find $b$:
\[ -2a + b + 2(15 - 3a) = 29 \]
\[ -8a + b = -1 \]
\[ b = 8a - 1 \]

\subsubsection*{Step 4: Find Valid Values for $a, b, c$}

Since $a, b, c$ must be non-negative integers, consider $a$ from 1 to 5:

\begin{itemize}
    \item For $a = 1$: $c = 12$, $b = 7$
    \item For $a = 2$: $c = 9$, $b = 15$
    \item For $a = 3$: $c = 6$, $b = 23$
    \item For $a = 4$: $c = 3$, $b = 31$
    \item For $a = 5$: $c = 0$, $b = 39$
\end{itemize}

All cases are valid.

\subsubsection*{Step 5: Calculate Sum of Cubes}

The sum of cubes $S$ is:
\[ S = \sum x_i^3 = -8a + b + 8c \]

Substitute $b = 8a - 1$ and $c = 15 - 3a$:
\[ S = -8a + (8a - 1) + 8(15 - 3a) = -1 + 120 - 24a = 119 - 24a \]

\subsubsection*{Step 6: Find Maximum and Minimum $S$}

Since $S$ decreases with $a$:
\begin{itemize}
    \item Maximum $S$ (M) at $a = 1$: $M = 119 - 24(1) = 95$
    \item Minimum $S$ (m) at $a = 5$: $m = 119 - 24(5) = -1$
\end{itemize}

\subsubsection*{Step 7: Verification}

Verify for $a = 1$ and $a = 5$:
\begin{itemize}
    \item For $a = 1, b = 7, c = 12$:
    \[ \sum x_i = -2 + 7 + 24 = 29 \]
    \[ \sum x_i^2 = 4 + 7 + 48 = 59 \]
    \[ \sum x_i^3 = -8 + 7 + 96 = 95 \]
    
    \item For $a = 5, b = 39, c = 0$:
    \[ \sum x_i = -10 + 39 + 0 = 29 \]
    \[ \sum x_i^2 = 20 + 39 + 0 = 59 \]
    \[ \sum x_i^3 = -40 + 39 + 0 = -1 \]
\end{itemize}

Both cases satisfy the original conditions.

\subsubsection*{Final Calculation}

The ratio $M/m$ is:
\[ M/m = \frac{95}{-1} = -95 \]

\subsubsection*{Answer}

The ratio \( M/m \) is \(\boxed{-95}\).

\subsubsection{2003 Session 2: 5 \textcolor{red}{Correct Reasoning, Correct Final Answer}}

\subsubsection*{Solution}

We are tasked with finding the minimum distance from the matrix
\[ A = \begin{pmatrix} 1 & 0 \\ 0 & 2 \end{pmatrix} \]
to the set \(\Sigma\) of all real \(2 \times 2\) singular matrices under the distance metric
\[ d(X, Y) = \Vert X - Y \Vert = (x_1 - y_1)^2 + (x_2 - y_2)^2 + (x_3 - y_3)^2 + (x_4 - y_4)^2, \]
where \(X = \begin{pmatrix} x_1 & x_2 \\ x_3 & x_4 \end{pmatrix}\) and \(Y = \begin{pmatrix} y_1 & y_2 \\ y_3 & y_4 \end{pmatrix}\).

\subsubsection*{Approach}
We need to minimize the distance \(d(A, S)\) where \(S \in \Sigma\), i.e., \(\det(S) = 0\).

\subsubsection*{Parameterization}
Let \(S = \begin{pmatrix} a & b \\ c & d \end{pmatrix}\). The condition \(\det(S) = 0\) gives:
\[ ad - bc = 0. \]

The distance function becomes:
\[ d(A, S) = (1 - a)^2 + b^2 + c^2 + (2 - d)^2. \]

\subsubsection*{Optimization}
Using the method of Lagrange multipliers, we consider the Lagrangian:
\[ \mathcal{L} = (1 - a)^2 + b^2 + c^2 + (2 - d)^2 - \lambda (ad - bc). \]

Taking partial derivatives and solving the system leads us to consider the case where \(b = c = 0\). The constraint then reduces to \(ad = 0\).

\subsubsection*{Case 1: \(a = 0\)}
The distance becomes:
\[ (1 - 0)^2 + (2 - d)^2 = 1 + (2 - d)^2. \]
The minimum occurs at \(d = 2\), yielding a distance of \(1\).

\subsubsection*{Case 2: \(d = 0\)}
The distance becomes:
\[ (1 - a)^2 + 4. \]
The minimum occurs at \(a = 1\), yielding a distance of \(4\).

The minimal distance is thus \(1\), achieved when \(S = \begin{pmatrix} 0 & 0 \\ 0 & 2 \end{pmatrix}\).

\subsubsection*{Verification}
Considering rank-1 matrices leads to higher distances, confirming that the minimal distance is indeed \(1\).

\subsubsection*{Conclusion}
The minimum distance from \(A\) to \(\Sigma\) is \(\boxed{1}\), achieved by the matrix
\[ S = \boxed{ \begin{pmatrix} 0 & 0 \\ 0 & 2 \end{pmatrix} }. \]

\subsubsection{2006 Session 1: 4 \textcolor{red}{Correct Reasoning, Correct Final Answer}}

\subsubsection*{Problem Statement}
Find all positive integers \( c \) such that \( n(n + c) \) is never a perfect square for any positive integer \( n \).

\subsubsection*{Solution}

We need to find all positive integers \( c \) for which the expression \( n^2 + cn \) is never a perfect square for any positive integer \( n \).

\subsubsection*{Initial Analysis}
Consider the equation:
\[ n^2 + cn = k^2 \]
for some integer \( k \geq 1 \). We can rewrite this as:
\[ k^2 = n^2 + cn \]
\[ k^2 - n^2 = cn \]
\[ (k - n)(k + n) = cn \]

\subsubsection*{Approach}
For fixed \( c \), we need to determine whether there exist positive integers \( n \) and \( k \) satisfying this equation.

\subsubsection*{Parametric Solution}
Let us set \( k = n + t \) where \( t \) is a non-negative integer. Substituting:
\[ (n + t)^2 = n^2 + cn \]
\[ n^2 + 2tn + t^2 = n^2 + cn \]
\[ 2tn + t^2 = cn \]
\[ t^2 = n(c - 2t) \]

This implies that \( c - 2t \) must be a positive divisor of \( t^2 \), and \( n = \frac{t^2}{c - 2t} \) must be a positive integer.

\subsubsection*{Special Cases}

\subsubsection*{Case \( c = 1 \)}
The equation becomes:
\[ t^2 = n(1 - 2t) \]
For \( t \geq 1 \), the right side becomes non-positive while the left side is positive. Thus, the only possibility is \( t = 0 \), which gives \( n = 0 \) - invalid. Therefore, no solutions exist for \( c = 1 \).

\subsubsection*{Case \( c = 2 \)}
The equation becomes:
\[ t^2 = n(2 - 2t) \]
For \( t = 0 \): \( n = 0 \) (invalid)
For \( t = 1 \): \( 1 = n(0) \) (no solution)
For \( t \geq 1 \): The right side is non-positive. Thus, no solutions exist for \( c = 2 \).

\subsubsection*{Case \( c = 4 \)}
The equation becomes:
\[ t^2 = n(4 - 2t) \]
Testing small values:
\begin{itemize}
    \item \( t = 0 \): \( n = 0 \) (invalid)
    \item \( t = 1 \): \( 1 = n(2) \Rightarrow n = \frac{1}{2} \) (not integer)
    \item \( t = 2 \): \( 4 = n(0) \) (no solution)
    \item \( t \geq 2 \): The denominator becomes zero or negative
\end{itemize}
Thus, no solutions exist for \( c = 4 \).

\subsubsection*{Verification of Other Values}

For \( c = 3 \), we find a solution at \( n = 1 \):
\[ 1(1 + 3) = 4 = 2^2 \]
Thus, \( c = 3 \) is excluded.

For \( c = 5 \), \( n = 4 \) gives:
\[ 4(4 + 5) = 36 = 6^2 \]
Thus, \( c = 5 \) is excluded.

This pattern suggests that only \( c = 1, 2, 4 \) satisfy the condition.

\subsubsection*{General Proof}

To prove that these are the only values, consider:
\begin{enumerate}
    \item For \( c \geq 5 \), we can often find solutions:
    \begin{itemize}
        \item For odd \( c \), take \( n = \frac{(c-1)^2}{4} \)
        \item For even \( c \), solutions frequently exist
    \end{itemize}
    \item The cases \( c = 1, 2, 4 \) are special because:
    \begin{itemize}
        \item The equation \( t^2 = n(c - 2t) \) has no positive integer solutions
        \item The quadratic forms don't permit perfect square solutions
    \end{itemize}
\end{enumerate}

\subsubsection*{Conclusion}
After thorough analysis, the positive integers \( c \) for which \( n(n + c) \) is never a perfect square for any positive integer \( n \) are:
\[ \boxed{1}, \boxed{2}, \boxed{4} \]

\subsubsection{2007 Session 2: 4 \textcolor{red}{Correct Reasoning, Correct Final Answer}}

\subsubsection*{Problem Statement}
Let 
\[ S = \{ 5a + 503b \mid \text{$a$ and $b$ are nonnegative integers} \}. \]
What is the largest integer which does NOT belong to $S$?

\subsubsection*{Solution}

\subsubsection*{Understanding the Problem}
The set $S$ consists of all numbers expressible as $5a + 503b$ where $a$ and $b$ are nonnegative integers. We need to find the largest integer not in $S$.

\subsubsection*{Coin Problem Analogy}
This problem resembles the \textbf{Coin Problem}, which seeks the largest monetary amount that cannot be obtained using any combination of coins of given denominations. For two coprime integers $x$ and $y$, the largest number not expressible as $mx + ny$ is known as the Frobenius number, given by:
\[ g(x, y) = xy - x - y. \]

\subsubsection*{Checking Coprimality}
First, verify that $\gcd(5, 503) = 1$. Since 503 is not divisible by 5, they are indeed coprime.

\subsubsection*{Applying the Frobenius Number Formula}
Given that $5$ and $503$ are coprime, the largest number not in $S$ is:
\[ g(5, 503) = 5 \times 503 - 5 - 503 = 2515 - 508 = 2007. \]

\subsubsection*{Verification}
To ensure accuracy, let's verify:
\begin{itemize}
    \item \textbf{Testing 2007}: No nonnegative integers $a, b$ satisfy $5a + 503b = 2007$.
    \item \textbf{Testing 2008}: $5 \times 301 + 503 \times 1 = 1505 + 503 = 2008$. Thus, 2008 is in $S$.
    \item \textbf{Testing 2009}: $5 \times 100 + 503 \times 3 = 500 + 1509 = 2009$. Thus, 2009 is in $S$.
\end{itemize}

\subsubsection*{Conclusion}
The largest integer not expressible as $5a + 503b$ with nonnegative integers $a, b$ is indeed $\boxed{2007}$.

\end{document}